\documentclass[letterpaper]{article} % DO NOT CHANGE THIS
\usepackage{aaai2027}  % DO NOT CHANGE THIS
\usepackage[hyphens]{url}  % DO NOT CHANGE THIS
\usepackage{graphicx} % DO NOT CHANGE THIS
\usepackage{natbib}  % DO NOT CHANGE THIS AND DO NOT ADD ANY OPTIONS TO IT
\usepackage{caption} % DO NOT CHANGE THIS AND DO NOT ADD ANY OPTIONS TO IT
\usepackage{algorithm}
\usepackage{algorithmic}
\usepackage{amsmath}
\usepackage{amssymb}
\usepackage{bm}
\usepackage{booktabs}
\usepackage{multirow}
\newcommand{\suppA}{App.~A of the supplementary material}
\newcommand{\suppB}{App.~B of the supplementary material}
\newcommand{\suppC}{App.~C of the supplementary material}
\newcommand{\suppAshort}{App.~A (supplementary material)}

\newcommand{\Wz}{W_0}
\newcommand{\WK}{W_K}
\newcommand{\Gxx}{G_{xx}}
\newcommand{\Gyx}{G_{yx}}
\newcommand{\Ql}{Q_\ell}
\newcommand{\Qr}{Q_r}
\newcommand{\tr}{\mathrm{tr}}
\newcommand{\Var}{\mathrm{Var}}

\newcommand{\degr}{^{\circ}}

\title{COEC: Calibrated Orthogonal-Equivalence Compensation \\ for Structured Pruning of Large Language Models}

\author{
    Peiqi Yu\textsuperscript{\rm 1},
    Nam Ling\textsuperscript{\rm 1},
    Wei Wang\textsuperscript{\rm 2},
    Wei Jiang\textsuperscript{\rm 2}
}
\affiliations{
    \textsuperscript{\rm 1}School of Engineering, Santa Clara University, Santa Clara, CA, USA\\
    \textsuperscript{\rm 2}Futurewei Technologies, Inc., USA\\
    pyu@scu.edu
}
\begin{document}

\maketitle
\begin{abstract}
Structured pruning reduces the size and inference cost of large language models (LLMs) by removing weight columns, but the resulting output error can degrade accuracy. Existing training-free compensation methods use an additive bias or a single orthogonal rotation on the output side of the retained weight. These corrections leave its input singular frame unchanged and therefore limit how the retained weight can adapt after column removal. We propose COEC (Calibrated Orthogonal-Equivalence Compensation), a training-free compensation framework that applies alternating left and right orthogonal rotations to the retained weight. The right rotation is optimized on a reduced Stiefel manifold, while singular values are rescaled using generalized cross-validation to select the regularization strength for each layer. COEC further tempers the calibration Gram matrix to reduce the dominance of high-energy activation directions and introduces an alignment penalty that preserves the geometric relation between adjacent attention projections.All components use second-order statistics from a small calibration set and require neither backpropagation through the LLM nor retraining of the model parameters. COEC is independent of the column pruning criterion and can be applied to multiple structured pruning methods. Experiments on the Llama-3, Llama-3.1, and Qwen2.5 model families across multiple structured sparsity levels show that COEC improves perplexity on every model and zero-shot accuracy in most settings over existing compensation methods, with larger gains at higher sparsity. These results show that post-pruning compensation can recover part of the performance lost to column removal.
\end{abstract}

\section{Introduction}
\label{sec:intro}

\begin{figure*}[t]
\centering
\includegraphics[width=\textwidth]{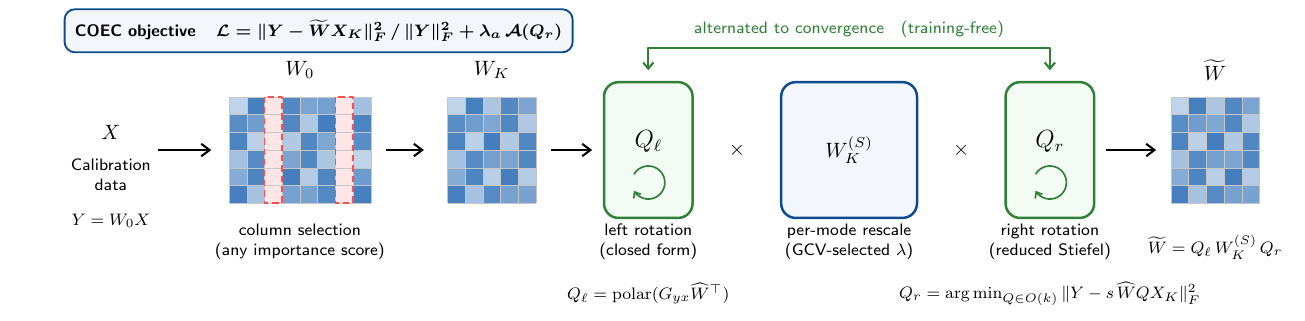}
\caption{Overview of COEC. Columns are pruned using an arbitrary importance score, and the retained weight $\WK$ is compensated as $\widetilde{W}=\Ql\,\WK^{(S)}\,\Qr$. The left and right rotations adjust the output and input singular frames, while per-mode rescaling adapts the singular values. All updates use calibration statistics only, without LLM backpropagation or model retraining. The objective combines relative reconstruction error with an alignment penalty on the attention output projection.}
\label{fig:pipeline}
\end{figure*}

Large language models (LLMs) are increasingly deployed across a wide range of applications, yet their inference costs in computation and memory remain substantial. Common compression techniques include quantization \citep{gptq,awq,smoothquant}, distillation \citep{hinton2015,minillm}, and pruning \citep{obd,obs}. Among them, structured pruning removes parameters at the granularity of entire weight-matrix columns, including MLP neurons and attention key--value groups. This produces a smaller dense model with reduced parameter count, memory usage, and computational cost on standard hardware. Because retraining at LLM scale is expensive, structured pruning methods that preserve accuracy without additional training are particularly attractive.

A training-free structured pruning pipeline comprises two largely separable components: a \emph{selection} rule that determines which columns to retain and a \emph{compensation} mechanism that adjusts the retained weights to reduce the output error induced by pruning. Existing methods have primarily focused on selection, while treating compensation as secondary. Wanda-sp \citep{wanda} extends Wanda's activation-aware importance score to column pruning but performs no compensation, leaving the output discrepancy caused by the removed columns uncorrected. FLAP \citep{flap} compensates for the mean error through an additive bias term. This approach is simple and effective at low sparsity, but a constant bias cannot correct input-dependent directional errors. Direct least-squares refitting of the retained weights can reduce calibration error, but with a limited calibration set it is prone to overfitting and may substantially distort the pretrained weights \citep{rcpu}. More recently, RCPU \citep{rcpu} constrains compensation to an orthogonal rotation of the output space followed by global rescaling. This geometry-preserving formulation avoids the distortions introduced by unconstrained refitting and achieves state-of-the-art performance.

However, RCPU applies rotation only on the left. Let the singular value decomposition (SVD) of the retained weight  be $\WK = U_K \Sigma_K V_K^{\top}$. A left rotation $Q\WK$ changes the output frame $U_K$ while leaving the input frame $V_K$ unchanged. This one-sided constraint is limiting because column pruning removes input coordinates and alters the correlation structure represented by the retained columns. Effective compensation must therefore recover, through the retained columns, the output contributions previously associated with the removed columns. Our analysis in Sec.~\ref{sec:geometry} shows that pruning substantially rotates the input singular subspace across all examined layers, and neither left-sided rotation nor reconstruction-optimal compensation generally restores this subspace.

Motivated by these limitations, we propose COEC (Calibrated Orthogonal-Equivalence Compensation), a general post-pruning compensation framework designed as a universal plug-in (Fig.~\ref{fig:pipeline}). COEC takes the retained set produced by any column-importance criterion and modifies only the retained weights. It can therefore be applied after existing structured pruning methods without changing their selection rules or model architectures. Rather than refitting the retained weights through unconstrained least squares, COEC searches within a spectrum-preserving family that applies rotations on both sides and rescales individual singular modes. COEC alternates between a closed-form left rotation and an iterative right rotation optimized over a reduced Stiefel manifold. Because the right-rotation objective is coupled with the overall scale, the optimal scaling factor is recomputed at every iteration. COEC further replaces global rescaling with per-mode singular-value rescaling, with the regularization strength selected independently for each layer through closed-form generalized cross-validation. Two additional components improve the calibration objective and preserve inter-layer geometry. Spectral tempering of the calibration Gram matrix balances perplexity and zero-shot accuracy, while an alignment penalty anchored to the original input frame restores geometric relations between adjacent layers that reconstruction objectives alone fail to preserve. All components depend only on second-moment statistics estimated from a small calibration set and require neither backpropagation through the LLM nor retraining of the model parameters.

We evaluate COEC on the Llama-3, Llama-3.1, and Qwen2.5 model families across multiple structured sparsity levels. Under identical column selections, COEC improves perplexity on every model and zero-shot accuracy in most settings over existing compensation methods, with larger gains at higher pruning ratios. When applied to the column selections produced by Wanda-sp and FLAP, COEC recovers a substantial portion of the accuracy lost during pruning. Our main contributions are summarized as follows.

\begin{itemize}
\item We formulate post-pruning compensation as a two-sided spectrum-preserving update. We develop an alternating solver that combines a closed-form left rotation, a reduced-Stiefel right rotation with in-loop rescaling, per-mode rescaling selected by generalized cross-validation, spectral tempering, and an inter-layer alignment penalty. We isolate the effect of each component through ablation studies in Sec.~\ref{sec:abl}.

\item We design COEC as a criterion-agnostic compensation method that can be applied after any column-selection rule without changing the model architecture. We demonstrate this plug-in capability using the selections produced by Wanda-sp, FLAP, and RCPU in Sec.~\ref{sec:plugin}.

% \item We analyze how structured pruning changes the input and output singular frames of each layer and determine which changes are recovered by reconstruction-based compensation. The analysis builds on the intra-layer and inter-layer alignment framework of Weak Geometric Alignment Theorem (WSBM) \cite{wsbm} and is presented in Sec.~\ref{sec:geometry}.

\item We evaluate COEC under a unified protocol on the Llama-3, Llama-3.1, and Qwen2.5 model families across multiple structured sparsity levels. COEC improves perplexity on every model and zero-shot accuracy in most settings over Wanda-sp, FLAP, and RCPU in Sec.~\ref{sec:main}.

\end{itemize}

\section{Related Work}
\label{sec:related}
\noindent\textbf{Pruning of LLMs.} Post-training pruning reduces the size and inference cost of LLMs without retraining, building on classical saliency criteria \citep{obd,obs,lottery,movement}. Unstructured and semi-structured methods sparsify weights element-wise, by per-row least-squares reconstruction in SparseGPT \citep{sparsegpt} or activation-aware scoring in Wanda
\citep{wanda}, but need sparse kernel support to realize speedups. Structured methods remove entire columns, rows, or blocks and directly shrink the dense model, using gradient saliency \citep{llmpruner}, masks learned with continued pretraining \citep{shearedllama}, curvature information \citep{ziplm,surgeon}, or block-level redundancy \citep{shortgpt,sleb,laco}; Wanda-sp extends the Wanda score to columns, and FLAP \citep{flap} scores channels by activation variance. Low-rank factorization \citep{asvd,svdllm} also operates in the singular-value domain but truncates the spectrum, which our compensation
preserves. We work in the structured, training-free setting.

\noindent\textbf{Post-pruning compensation.} Methods differ in how they repair the kept weights after removal. Wanda-sp applies no update. FLAP \citep{flap} corrects the mean error with a bias term, which cannot fix input-dependent directional mismatches. Least-squares refitting reduces calibration error further but overfits small calibration sets and damages pretrained weights \citep{rcpu}.RCPU \citep{rcpu} restricts the update to an output-frame rotation with a global scale, preserving geometry and overfitting less. We extend this one-sided rotation to a converged two-sided update with per-mode rescaling, independent of the selection criterion.

\noindent\textbf{Rotation-based transformations.} SliceGPT \citep{slicegpt} rotates RMSNorm-connected transformers under computational invariance and
prunes in the rotated basis; QuaRot \citep{quarot}, SpinQuant \citep{spinquant}, and DenoiseRotator \citep{denoiserotator} use lossless rotations to condition weights for compression. All of these rotate before compression under functional equivalence, whereas we rotate after pruning to compensate the removal error.

\noindent\textbf{Calibration-aware rescaling.} Scale and bias corrections appear in FLAP and in energy-matching heuristics. Ours is instead a diagonal fit in the SVD frame of the rotated weight, with the ridge strength set per layer by generalized cross-validation (GCV)~\citep{gcv,craven}, a classical criterion that to our knowledge has not been used for pruning compensation.
The right rotation is a weighted orthogonal Procrustes problem \citep{gower} with no closed-form solution, which we solve on the reduced Stiefel manifold
\citep{edelman,absil} (Sec.~\ref{sec:stiefel}).

\section{Problem Formulation}
\label{sec:setup}

\subsection{Notation and Setup}

For each transformer block, we prune structured channels in both the MLP and attention sub-layers. In the MLP, pruning removes intermediate channels by deleting the corresponding rows of the gating and value projections and the matching columns of the output projection. In grouped-query attention, pruning removes complete KV groups together with their associated query heads. This corresponds to deleting the associated output rows of the query, key, and value projections and the matching input columns of the attention output projection. The grouped-query attention structure is preserved after pruning.

Let $\Wz \in \mathbb{R}^{d_{\mathrm{out}}\times d_{\mathrm{in}}}$ denote the pretrained weight matrix of a pruned sub-layer. Let $X\in\mathbb{R}^{d_{\mathrm{in}}\times N}$ denote the calibration activations from $N$ token positions, and $Y=\Wz X$ the corresponding output of the original layer. A column-selection rule retains an index set $K$ with $|K|=k$, and $X_K\in\mathbb{R}^{k\times N}$ contains the rows of $X$ indexed by $K$. The compensation step constructs a weight matrix $\widetilde{W}\in\mathbb{R}^{d_{\mathrm{out}}\times k}$ acting on $X_K$. Its quality is measured by the relative reconstruction error:
\begin{equation}
\frac{\lVert Y - \widetilde{W} X_K\rVert_F^2}{\lVert Y\rVert_F^2} =\frac{\tr(\widetilde{W} \Gxx \widetilde{W}^{\top}) - 2\,\tr(\widetilde{W} \Gyx^{\top}) + \lVert Y\rVert_F^2}{\lVert Y\rVert_F^2}.\nonumber
\end{equation}
The objective depends only on the second-moment statistics $G=XX^{\top}$, $\Gxx=G[K,K]$, and $\Gyx=\Wz\,G[:,K]$, accumulated in a single calibration pass. The token-level activations do not need to be stored after accumulation. In practice, 128 calibration sequences are sufficient (Sec.~\ref{sec:setup-exp}).

\subsection{Limitation of Least-Squares Compensation}
The unconstrained least-squares solution is $W^{\mathrm{LS}}=\Gyx\Gxx^{-1},$ when $\Gxx$ is invertible. In practice, a ridge-regularized solution is used as $W^{\mathrm{ridge}}=\Gyx(\Gxx+\lambda_{\mathrm{ls}} I)^{-1}$. Consistent with the observations reported for RCPU, we find that least-squares refitting can fit the small calibration set too closely. It reshapes the singular spectrum of the retained weight according to the calibration statistics and shows unstable generalization across models.

To limit such unrestricted changes, we parameterize the compensated weight as
\begin{equation}
\widetilde{W} = \Ql\, \WK^{(S)}\, \Qr, \qquad \Ql \in O(d_{\mathrm{out}}),\; \Qr \in O(k),\nonumber
\end{equation}
where $\WK^{(S)}$ is obtained by applying a regularized rescaling to the singular modes of the retained weight. The left and right orthogonal transforms adjust the output and input singular frames, respectively, while the mode-wise rescaling controls changes to the singular spectrum. This structured parameterization avoids the unrestricted deformation introduced by direct least-squares refitting. COEC estimates these components by minimizing, for each pruned module,
\begin{equation}
\label{eq:obj}
\mathcal{L}(\widetilde{W},\Qr) = \frac{\lVert Y-\widetilde{W}X_K\rVert_F^2}{\lVert Y\rVert_F^2} + \lambda_a\,\mathcal{A}(\Qr),
\end{equation}
where $\mathcal{A}$ is the inter-layer alignment penalty of Sec.~\ref{sec:align}, applied to the attention output projection with $\lambda_a=0$ elsewhere. The following sections describe how these components are estimated.

\section{Method}
\label{sec:method}

\subsection{Two-Sided Rotation with In-Loop Scaling}
\label{sec:init}
\label{sec:left}
\label{sec:right}
We initialize the compensated weight with the retained column slice as $\WK = \Wz[:,K]$. No rank truncation or singular-value rescaling is applied, and $\WK$ represents the uncompensated pruned weight used as the reference point in \suppC. We write $\widehat{W}$ for the current compensated weight, initialized as $\widehat{W}=\WK$ and overwritten by each block update; \eqref{eq:obj} is evaluated at $\widehat{W}$ and the current $\Qr$.

For a fixed $\widehat{W}$, the optimal left rotation is given by the orthogonal Procrustes solution \citep{schonemann}: $\Ql = \mathrm{polar}(\Gyx\,\widehat{W}^{\top})$, where $\mathrm{polar}(A)$ denotes the orthogonal factor $U_A V_A^{\top}$ of the SVD $A = U_A \Sigma_A V_A^{\top}$. The left rotation minimizes the reconstruction error by aligning the outputs of the retained weight with those of the original layer on the calibration data.

The global scale can be optimized independently of the left rotation because an orthogonal transformation preserves the norm of $\widehat{W}X_K$. For a fixed $\widehat{W}$, the optimal scale is:
\begin{equation}
s^{*}={\langle \Gyx,\widehat{W}\rangle}/{\langle \widehat{W},\widehat{W}\Gxx\rangle},\nonumber
\end{equation}
where $\langle A,B\rangle=\tr(AB^{\top})$. The combination of a left rotation and global scaling corresponds to the compensation used by RCPU. In COEC, it forms the left-rotation block of the alternating compensation procedure.

The right rotation is obtained by solving a weighted orthogonal Procrustes problem,
\begin{equation}
\min\nolimits_{\Qr\in O(k)} \; \lVert Y - s\, \widehat{W}\, \Qr\, X_K \rVert_F^2.\nonumber
\end{equation}
Unlike the left-rotation problem, this objective has no closed-form solution because the rotation is coupled to both the weight matrix and the calibration covariance. We optimize $\Qr$ using projected gradient descent with an orthogonal retraction and initialize it as $Q_0 = I$. This identity initialization is empirically important at mild sparsity, where the optimal right rotation typically remains close to the original input frame. Initializing from the one-sided Procrustes solution instead performs worse, e.g., increasing reconstruction error on attention output projections from $0.019$ to $0.159$.

The right rotation is also coupled to the global scale. We therefore recompute $s^{*}$ after every update of $\Qr$ and evaluate the gradient using the scaled residual. The step size is normalized as $\eta/s^2$ to account for the scale of the objective. This procedure allows the right rotation and global scale to converge jointly.

The complete optimization alternates between the left rotation, scale update, right rotation, and another scale update
\begin{equation}
\Ql
\rightarrow
s
\rightarrow
\Qr
\rightarrow
s.\nonumber
\end{equation}
We perform at most 50 alternating rounds and stop early when the change in $\mathcal{L}$ falls below $10^{-4}$. The left rotation is the conditional minimizer of the reconstruction objective and leaves the alignment term unchanged, and the regularized rescale does not increase the reconstruction error in the tempered metric. The resulting parameterization includes left-only and right-only compensation as special cases.

\subsection{Reduced-Stiefel Solver}
\label{sec:stiefel}
Directly optimizing $\Qr\in O(k)$ is inefficient when the compensated weight is processed in output slices. Consider a slice $W_g\in\mathbb{R}^{b\times k}$ with $b<k$, and let $W_g=U_g\Sigma_g V_g^{\top}$, where $V_g\in\mathbb{R}^{k\times b}$ contains its right singular vectors. The reconstruction objective depends on $\Qr$ only through $R=V_g^{\top}\Qr\in\mathbb{R}^{b\times k}$. The components of $\Qr$ in the orthogonal complement of $V_g$ do not affect the slice output and introduce redundant optimization directions. We therefore optimize $R$ directly under the constraint $RR^{\top}=I_b$, which places $R$ on the Stiefel manifold $\mathrm{St}(b,k)$ \citep{edelman,absil}.

Any feasible $R\in\mathrm{St}(b,k)$ can be extended to an orthogonal matrix $\Qr\in O(k)$. The reduced and full parameterizations therefore represent the same solutions for the slice objective, while the reduced formulation removes directions that do not affect the output. Its memory and per-step computational costs are reduced by approximately a factor of $b/k$.

We use a Newton--Schulz polar retraction for rectangular frames. For square frames, we use an exact Cayley retraction \citep{wenyin}, with the required inverse computed by Newton--Schulz iteration. Implementation details are provided in \suppB. Optimization stops when the change in the rotation falls below a tolerance $\delta$. This tolerance is selected once for each model using the calibration set.

\subsection{GCV-Based Singular-Value Rescaling}
\label{sec:gcv}
A global scalar multiplies all singular values by the same factor, although the calibration statistics may favor different adjustments for different singular-vector pairs. Let the SVD of the current rotated weight be $\widehat{W} = \widehat{U}\,\widehat{\Sigma}\,\widehat{V}^{\top}$, where $\sigma_i=[\widehat{\Sigma}]_{ii}$ is the $i$-th singular value and $u_i,v_i$ the corresponding singular vectors. Keeping $\widehat{U}$ and $\widehat{V}$ fixed, the reconstruction-optimal value associated with the $i$-th singular-vector pair is
\begin{equation}
s_{g,i}^{\star}
={u_i^{\top}\Gyx v_i}/{v_i^{\top}\Gxx v_i}
={\rho_i}/{e_i}.\nonumber
\end{equation}
where the mode response $\rho_i$ measures the alignment between the corresponding output and input singular directions under the target cross-covariance, while the mode energy $e_i$ measures the calibration energy along the input singular direction. When the weight has rank one, this expression reduces to the globally scaled singular value $s^{*}\sigma_i$.

Estimating each singular value independently can overfit the limited calibration data. We therefore regularize $s_{g,i}^{\star}$ toward the value obtained from uniform global scaling, $s^{*}\sigma_i$. The regularized singular value is
\begin{equation}
s_{g,i}(\lambda)=\left({\rho_i+\lambda\, s^{*}\sigma_i}\right)/\left({e_i+\lambda}\right). \nonumber
\end{equation}
When $\lambda\rightarrow0$, $s_{g,i}(\lambda)\rightarrow s_{g,i}^{\star}$. As $\lambda$ increases, it approaches the globally scaled value $s^{*}\sigma_i$. We select $\lambda$ independently for each layer using generalized cross-validation (GCV):
\begin{equation}
\begin{split}
\mathrm{GCV}(\lambda) &= \frac{\sum_i e_i\,\big(s_{g,i}(\lambda)-s_{g,i}^{\star}\big)^2}{\big(1 - \mathrm{df}(\lambda)/M\big)^2}, \\
\mathrm{df}(\lambda) &= \sum\nolimits_i \frac{e_i}{e_i+\lambda},\nonumber
\end{split}
\end{equation}
where $M$ is the number of singular values included in the layer-wise estimate. The effective degrees of freedom penalize solutions that fit the calibration statistics too closely. This correction is necessary because the unadjusted calibration residual is minimized at $\lambda=0$.

As $\lambda\rightarrow0^{+}$, both the numerator and denominator approach zero at the same order, producing a finite limiting value. As $\lambda\rightarrow\infty$, the criterion approaches $\mathrm{GCV}(\infty) =\sum_i e_i \left( s^{*}\sigma_i-s_{g,i}^{\star} \right)^2$. We minimize this objective over a logarithmically spaced grid for each layer. On both validation models, the resulting layer-specific regularization matches the performance of the best globally tuned value (see \suppB).

The rescaling is computed from the SVD of the current rotated weight. It can therefore be applied after either the left or right rotation. We recompute the singular values within each alternating round so that the spectrum adapts jointly with the left and right singular frames.

\subsection{Gram Tempering}
\label{sec:temper}
All compensation terms use the activation Gram matrix $G=XX^{\top}$, \(X\) containing calibration activations. Its diagonal entries measure activation energy, while its off-diagonal entries capture correlations between activation dimensions.

Let $G=E\operatorname{diag}(\mu_i)E^{\top}$ be its eigendecomposition. We temper the spectrum as:
\begin{equation}
G^{\alpha}
=E\operatorname{diag}(\mu_i^{\alpha})E^{\top}, \quad \alpha\in(0,1]. \nonumber
\end{equation}
When \(\alpha=1\), the original activation statistics are preserved. Decreasing \(\alpha\) compresses the spectrum, reducing the dominance of high-energy directions and giving relatively more weight to lower-energy directions.

Across the evaluated models, the best values lie between \(0.3\) and \(0.9\). We use \(\alpha=0.9\) by default and evaluate its effect in Sec.~\ref{sec:abl}.

\subsection{Anchored Inter-Layer Alignment}
\label{sec:align}

The preceding components minimize reconstruction error but do not explicitly preserve the geometric relation between consecutive projections. In attention, the value projection writes the intermediate representation that is read by the attention output projection. Let \(U_v\) denote the left singular frame of the pruned value projection and \(V_o\) the right singular frame of the attention output projection. Following  Weak Geometric Alignment Theorem (WSBM) \citep{wsbm}, their relation is characterized by \(V_o^{\top}U_v\). Our analysis in \suppC{} shows that pruning changes this relation and that reconstruction-based compensation does not restore it. We therefore regularize the right rotation of the attention output projection toward the original inter-layer alignment.

Direct comparison of singular-vector matrices is not basis invariant because singular vectors are defined only up to sign and rotations within degenerate subspaces. We instead compare the projectors onto their leading \(r_p\)-dimensional subspaces $\Pi_{r_p}(V)=V_{r_p}V_{r_p}^{\top}$, where \(V_{r_p}\) contains the top-\(r_p\) right singular vectors. We use \(r_p=16\) by default, independently of the retained width \(k\).

For the attention output projection, the reduced right-rotation variable is square, with \(R\in O(k)\). Its rows represent the transposed right singular frame of the compensated weight. The corresponding leading-subspace projector is \(\Pi_{r_p}(R)= R_{[1:r_p]}^{\top}R_{[1:r_p]}\). Let \(U_v\in\mathbb{R}^{k\times r_p}\) contain the leading left singular vectors of the pruned value projection, and let $\Pi_{\mathrm{anchor}}=\Pi_{r_p}(V_{\mathrm{anchor}})$ denote the fixed anchor projector. We define the alignment penalty as:
\begin{equation}
\mathcal{A}(R)=
\lVert
U_v^{\top}
\left(
\Pi_{r_p}(R)-\Pi_{\mathrm{anchor}}
\right)
U_v
\rVert_F^2/{r_p^2}.\nonumber
\end{equation}
$\mathcal{A}(R)$ is added to the right-rotation objective with weight \(\lambda_a\). Only \(U_v\) and \(\Pi_{\mathrm{anchor}}\) are precomputed for each layer. The analytic gradient depends only on the leading \(r_p\) rows of \(R\) and adds approximately \(0.5\%\) to the right-solver cost. $\mathcal{A}(R)$ is applied only to the attention output projection. The MLP compensation use the reconstruction objective alone.

Using the right singular frame of the retained weight as the anchor makes the penalty zero at initialization and provides little regularization. Instead, we construct the anchor from the original unpruned attention output projection. Let \(V_{\mathrm{full},r_p}\) contain its leading \(r_p\) right singular vectors. We restrict these vectors to the retained coordinates and orthonormalize them as $V_{\mathrm{anchor}} = \operatorname{orth} \left( V_{\mathrm{full},r_p}[K,:] \right)$.

This restricted original subspace differs from the right singular subspace of the retained weight by approximately \(20\)--\(30\degr\), as shown in \suppC. With the default \(\lambda_a=50\), the alignment penalty after compensation is a median \(7.2\times\) lower than that of the uncompensated pruned weight.

\subsection{Full Pipeline}
\label{sec:pipeline}

Figure~\ref{fig:pipeline} illustrates the overview of COEC, and Algorithm~\ref{alg:coec} summarizes the compensation procedure for one weight matrix. Numerical safeguards are described in \suppB.

\begin{algorithm}[t]
\caption{COEC compensation of one pruned module}
\label{alg:coec}
\textbf{Input}: original $\Wz$, kept set $K$, calibration Gram $G$; temper $\alpha$, tol $\delta$, align weight $\lambda_a$ %($\lambda_a{=}0$ on \texttt{down\_proj})
\\
\textbf{Output}: Compensated retained weight $\widetilde{W}$\\\vspace{-1em}
\begin{algorithmic}[1]
\STATE $G \gets G^{\alpha}$; \; $\Gxx \gets G[K,K]$; \; $\Gyx \gets \Wz\,G[:,K]$
\STATE $\widehat{W} \gets \Wz[:,K]$ \hfill (kept-column slice $\WK$)
\STATE $\Ql \gets \mathrm{polar}(\Gyx\widehat{W}^{\top})$; \; $\widehat{W} \gets \Ql \widehat{W}$
\STATE $\widehat{W} \gets \text{GCV-rescale}(\widehat{W})$
\WHILE{round $<50$ and $|\Delta\mathcal{L}| \ge 10^{-4}$}
\STATE $\Qr \gets$ $\operatorname{ReducedStiefelSolve}(\widehat{W}; Q_0{=}I,\delta,\lambda_a)$ %with scale in loop
\STATE $\widehat{W} \gets \widehat{W} \Qr$; \; $\widehat{W} \gets \text{GCV-rescale}(\widehat{W})$
\STATE $\Ql \gets \mathrm{polar}(\Gyx\widehat{W}^{\top})$; \; $\widehat{W} \gets \Ql \widehat{W}$; \; $\widehat{W} \gets \text{GCV-rescale}(\widehat{W})$
\ENDWHILE
\STATE \textbf{return} $\widetilde{W} \gets \widehat{W}$
\end{algorithmic}
\end{algorithm}

\section{Experiments}
\label{sec:exp}

\begin{table*}[t]
\centering
\small
\setlength{\tabcolsep}{4.5pt}
\begin{tabular}{ll cccccc}
\toprule
Selection & Comp. & Llama-3.1-8B & Llama-3-70B & Qwen2.5-7B & Qwen2.5-14B & Qwen2.5-32B & Qwen2.5-72B \\
\midrule
\multirow{2}{*}{Wanda-sp} & --   & 14.35 & 6.20 & 13.03 & 15.15 & 13.90 & 6.02 \\
                          & COEC & \textbf{10.89} & \textbf{5.61} & \textbf{9.09} & \textbf{13.26} & \textbf{9.64} & \textbf{5.17} \\
\midrule
\multirow{2}{*}{FLAP}     & bias & 11.89 & 5.86 & 8.65 & 8.82 & 6.85 & 5.29 \\
                          & COEC & \textbf{10.86} & \textbf{5.58} & \textbf{8.34} & \textbf{7.98} & \textbf{6.71} & \textbf{5.25} \\
\midrule
\multirow{2}{*}{RCPU}     & RCPU & 11.57 & 5.69 & 8.98 & 7.98 & 7.03 & 5.23 \\
                          & COEC & \textbf{10.91} & \textbf{5.58} & \textbf{8.91} & \textbf{7.63} & \textbf{6.74} & \textbf{5.21} \\
\bottomrule
\end{tabular}
\caption{\textbf{WikiText-2 perplexity} $(\downarrow)$ at 30\% column sparsity. Each selection score is paired with its native compensation and with COEC under the same selection score and calibration data. Bold marks the better compensation within each selection, per model. Results at 10\% and 20\% sparsity are in \suppAshort.}
\label{tab:main_ppl}
\end{table*}

\begin{table*}[t]
\centering
\small
\setlength{\tabcolsep}{4.5pt}
\begin{tabular}{ll cccccc}
\toprule
Selection & Comp. & Llama-3.1-8B & Llama-3-70B & Qwen2.5-7B & Qwen2.5-14B & Qwen2.5-32B & Qwen2.5-72B \\
\midrule
\multirow{2}{*}{Wanda-sp} & --   & 47.4 & 68.7 & 49.1 & 53.7 & 57.5 & 71.1 \\
                          & COEC & \textbf{48.9} & \textbf{70.4} & \textbf{56.0} & \textbf{53.8} & \textbf{61.0} & \textbf{72.5} \\
\midrule
\multirow{2}{*}{FLAP}     & bias & 44.8 & 68.8 & 52.6 & 55.9 & \textbf{65.6} & 70.5 \\
                          & COEC & \textbf{47.1} & \textbf{71.7} & \textbf{54.5} & \textbf{59.5} & 65.3 & \textbf{72.5} \\
\midrule
\multirow{2}{*}{RCPU}     & RCPU & 47.0 & 71.3 & 54.4 & \textbf{59.2} & 63.8 & 71.7 \\
                          & COEC & \textbf{50.2} & \textbf{71.7} & \textbf{56.1} & 58.5 & \textbf{64.2} & \textbf{72.0} \\
\bottomrule
\end{tabular}
\caption{\textbf{7-task zero-shot accuracy} \% $(\uparrow)$ at 30\% column sparsity. Same protocol and bolding as Table~\ref{tab:main_ppl}. Results at 10\% and 20\% sparsity are in \suppAshort.}
\label{tab:main_zs}
\end{table*}

\subsection{Experimental Setup}
\label{sec:setup-exp}
\noindent\textbf{Models.} We evaluate on Llama-3.1-8B and Llama-3-70B \citep{llama3} and on the Qwen2.5 family (7B, 14B, 32B, 72B) \citep{qwen25}, covering scales from 7B to 72B. Additional models are reported in the supplementary material under the same protocol.

\noindent\textbf{Pruning.} We evaluate structured sparsity levels of 10\%, 20\%, and 30\% in every transformer block, using the pruning scope described in Sec.~\ref{sec:setup}. The main tables report results at 30\% sparsity, where the effects of pruning and compensation are most pronounced. Results at 10\% and 20\% sparsity follow the same protocol and are provided in \suppA.

\noindent\textbf{Selection scores and baselines.} Columns are ranked by an activation-aware score and the top $\lceil(1-\rho)\,d_{\mathrm{in}}\rceil$ are kept for pruning ratio $\rho$. We use three published scores under one implementation: the Wanda-sp score $\lVert W_{:,j}\rVert\cdot\lVert X_j\rVert$, the WIFV score of FLAP, and the variance-aware score $\lVert W_{:,j}\rVert\cdot\lVert X_j\rVert\cdot\Var(X_j)$ of RCPU, which is the default selection for COEC. All methods share the calibration data and the evaluation protocol.

\noindent\textbf{Calibration and evaluation.} We sample 128 sequences from the WikiText-2 training set \citep{wikitext}. A single forward pass accumulates all Grams, and the compensation uses no evaluation data. We report WikiText-2 token-level perplexity and average zero-shot accuracy on seven tasks (BoolQ \citep{boolq}, RTE \citep{rte}, and WinoGrande \citep{winogrande} with accuracy; HellaSwag \citep{hellaswag}, ARC-e, ARC-c \citep{arc}, and OpenBookQA \citep{obqa} with normalized accuracy) via the LM Evaluation Harness \citep{lmeval}.

\subsection{Main Results}
\label{sec:main}
\label{sec:plugin}
Tables~\ref{tab:main_ppl} and~\ref{tab:main_zs} report the 30\% setting. Applied on top of Wanda-sp, COEC reduces perplexity on every model, with the largest gain from $13.90$ to $9.64$ on Qwen2.5-32B, and matches or improves zero-shot accuracy throughout. Against the FLAP bias, COEC gives lower perplexity on all six models, recovering the Llama-3.1-8B from $11.89$ to $10.86$, and improves accuracy on five of the six; at 10--20\% the bias remains competitive on perplexity (\suppA). Under the RCPU selection, COEC gives lower perplexity than the full RCPU compensation on all six models and higher zero-shot accuracy on five. The exception is Qwen2.5-14B, where RCPU is ahead by $0.7$ points on accuracy while COEC keeps the perplexity lead. These three comparisons cover three different column-importance criteria (\S\ref{sec:setup-exp}), and a single COEC configuration improves most settings under each without per-criterion tuning; the compensation can therefore be attached after existing pruning methods as a plug-in.

The gains grow with sparsity (\suppA), as heavier pruning leaves more error for the compensation to recover. The supplementary material repeats the same three comparisons on additional models under the same protocol. % TODO: once the supplementary runs land, decide whether to name the families and sizes here and whether to state an outcome

\begin{figure*}[t]
\centering
\includegraphics[width=\textwidth]{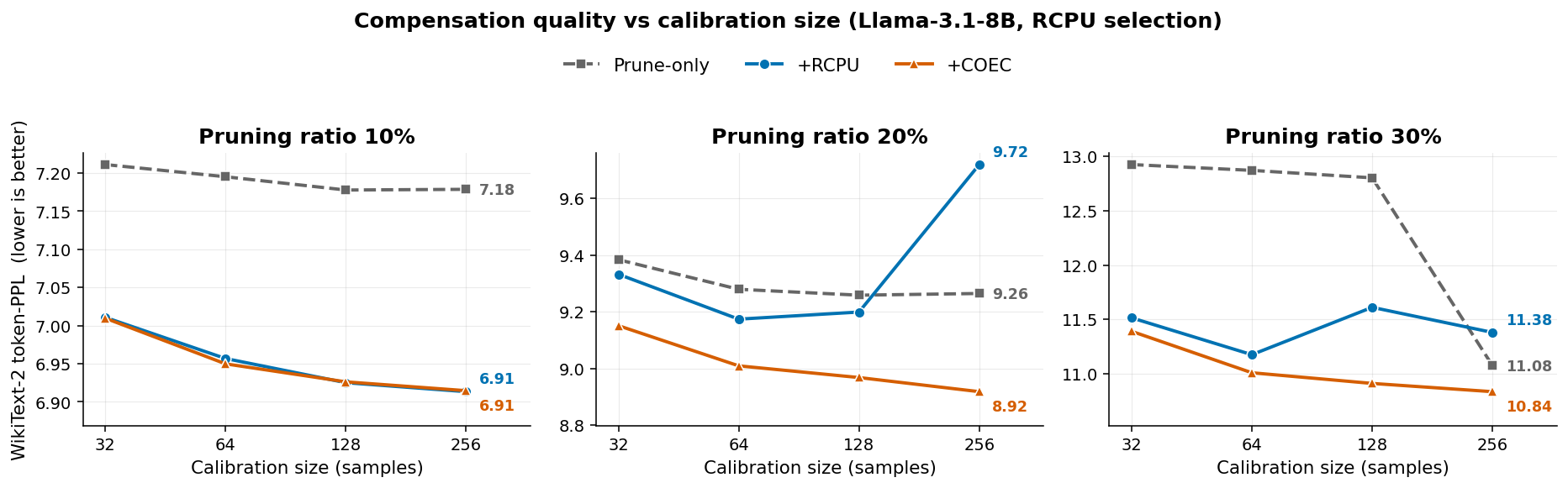}
\caption{Perplexity versus calibration size on Llama-3.1-8B under the RCPU selection score. The selection is recomputed at each $N$, so the prune-only baseline also varies.}
\label{fig:calib}
\end{figure*}

\subsection{Effect of Calibration Size}
\label{sec:calib}
All COEC components are estimated from calibration statistics, and the two-sided family has more rotational freedom than the one-sided update. A natural concern is that this freedom overfits a small calibration set. Figure~\ref{fig:calib} tests this by varying the calibration size $N\in\{32,64,128,256\}$ on Llama-3.1-8B at all three sparsity levels; all methods share the calibration data and the scoring rule. COEC gives the lowest perplexity at every size and every ratio and improves monotonically with $N$. Even at $N{=}32$ it stays well below the prune-only baseline; at 30\% sparsity, shrinking $N$ from $256$ to $32$ costs about $0.6$ PPL. RCPU has a smaller search space but is non-monotone in $N$. The stability of the larger family comes from its regularization rather than its size: GCV shrinks the per-mode rescale toward the uniform prior, and the rotations cannot reshape the singular spectrum to fit calibration noise. The main results use $N{=}128$, which lies on the flat part of every COEC curve.

\subsection{Ablation Study}
\label{sec:abl}
\begin{table}[t]
\centering
\small
\setlength{\tabcolsep}{5pt}
\begin{tabular}{lcc}
\toprule
Configuration & PPL\,$\downarrow$ & ZS\,$\uparrow$ \\
\midrule
Prune $+$ RCPU compensation                   & 10.06 & 50.9 \\
\quad $+$ GCV per-mode rescale                &  9.81 & 51.3 \\
\quad $+$ converged two-sided rotation        &  9.45 & 51.5 \\
\quad $+$ reduced-Stiefel right solve         &  9.86 & 52.7 \\
\quad $+$ Gram tempering ($\alpha{=}0.9$)     &  9.65 & 52.7 \\
\quad $+$ anchored align ($\lambda_a{=}50$)   &  9.65 & 52.9 \\
\bottomrule
\end{tabular}
\caption{Component-wise build-up of COEC (last row), averaged over five development models at 30\% sparsity. All rows use the RCPU selection score; indented rows accumulate from top to bottom.}
\label{tab:ablation}
\end{table}
The ablation uses five 7--8B development models chosen for architectural diversity: Mistral-7B \citep{mistral}, Llama-2-7B, Llama-3.1-8B, Qwen2.5-7B, and the distilled DeepSeek-R1-0528-Qwen3-8B \citep{deepseekr1}. Two of them appear in the main evaluation, and all hyperparameter defaults were selected on this set and transferred unchanged to the larger models of Tables~\ref{tab:main_ppl} and~\ref{tab:main_zs}. Table~\ref{tab:ablation} builds COEC up component by component at 30\% sparsity. The first row is the full RCPU compensation, the strongest published baseline; the indented rows then add the COEC components cumulatively. The GCV rescale reduces perplexity from $10.06$ to $9.81$, and the converged two-sided rotation further to $9.45$, supplying the input-side adjustment that the one-sided rotation cannot make. The reduced-Stiefel right solve trades perplexity for zero-shot accuracy ($51.5\to52.7$). Gram tempering then recovers part of the perplexity ($9.86\to9.65$) at unchanged accuracy, and the anchored alignment penalty adds a final accuracy gain, leaving the full method at $9.65$ perplexity and $52.9$ accuracy. Overall, the GCV rescale and the two-sided rotation account for the perplexity reduction, the reduced-Stiefel solve contributes most of the zero-shot gain at some perplexity cost, and Gram tempering recovers most of that cost.

\subsection{Efficiency Analysis}
\label{sec:efficiency}
Table~\ref{tab:efficiency} reports the deployment cost at 30\% column sparsity. Structured pruning yields a smaller dense model that runs without any sparse-kernel support: parameters, peak inference memory, and dense FLOPs per token fall by $23$--$29\%$, and the reduction grows with model size; realized wall-clock speedup additionally depends on how well the kernels handle the pruned matrix shapes, so we report these architecture-level quantities. COEC rotates and rescales the kept weights in place without changing the model shape, so the deployment cost depends only on the pruning ratio and not on the compensation; the recovery reported in Tables~\ref{tab:main_ppl} and~\ref{tab:main_zs} therefore comes at no additional cost. Compensating a 7--8B model takes $45$--$86$ GPU-minutes on a single B300; the alignment penalty adds only $0.5\%$ overhead, and GCV rescaling is closed-form.

\begin{table}[t]
\centering
\small
\setlength{\tabcolsep}{4pt}
\begin{tabular}{l cc cc cc}
\toprule
& \multicolumn{2}{c}{Params (B)} & \multicolumn{2}{c}{Mem (GB)} & \multicolumn{2}{c}{GFLOPs/tok} \\
\cmidrule(lr){2-3}\cmidrule(lr){4-5}\cmidrule(lr){6-7}
Model & dense & $30\%$ & dense & $30\%$ & dense & $30\%$ \\
\midrule
Qwen2.5-7B  & 7.62  & 5.70  & 15.0 & 11.5 & 15.2 & 11.4 \\
Llama-3-70B & 70.6 & 50.6 & 132.5 & 95.7  & 141.1 & 101.2 \\
\bottomrule
\end{tabular}
\caption{Deployment cost at 30\% column sparsity. The compensation does not change the model shape, so the figures are identical for prune-only, RCPU, and COEC.}
\label{tab:efficiency}
\end{table}

\subsection{Geometric Analysis}
\label{sec:geometry}
A weight matrix reads certain input directions and writes certain output directions. \suppC{} measures how pruning and compensation change these directions on DeepSeek-R1-0528-Qwen3-8B at 30\% sparsity. Two observations motivate the design of COEC. First, deleting columns rotates the input directions of every layer by about $22\degr$ on average. Compensation that only minimizes reconstruction error leaves this rotation in place. The anchored penalty of \S\ref{sec:align} targets this residual and reduces the measured mismatch by a median factor of $7.2$. Second, COEC rotates the weight as a whole, so its input and output directions move together. Their pairing within each layer stays intact (discrepancy of $1.1$--$1.2\%$) without any explicit term.

\section{Conclusion}
We presented COEC, a training-free compensation method for structured pruning of large language models. Instead of refitting the retained weights freely, COEC rotates them on both sides and rescales their singular values, so the update can re-route the signal lost with the removed columns while keeping the overall structure of the pretrained weight intact. The strength of the rescaling is chosen automatically from calibration statistics, and a light penalty keeps the pruned layer geometrically consistent with its neighbor. The whole procedure uses a small calibration set, requires no backpropagation through the LLM or retraining of the model parameters, and compensates a 7--8B model in under $1.5$ GPU-hours without changing the model architecture.

\bibliography{aaai2027}

\end{document}

\maketitle

% Switch to single column without the \clearpage that \onecolumn
% normally issues: nothing but the title has been typeset yet, so
% suppressing it is safe and avoids an otherwise blank first page.
\makeatletter
\let\coec@clearpage\clearpage
\let\clearpage\relax
\onecolumn
\let\clearpage\coec@clearpage
\makeatother

\appendix

\section{Results at 10\% and 20\% Column Sparsity}
\label{app:sweep}

Tables~\ref{tab:sweep_ppl} and~\ref{tab:sweep_zs} report the 10\% and 20\% settings under the protocol of \mSetupExp; the 30\% setting is in Tables~1 and~2 of \mainpaper. Under the Wanda-sp selection, COEC lowers perplexity on all six models at both ratios. Under the RCPU selection it lowers perplexity on five of six models at 10\% (matching on the sixth) and on all six at 20\%, by margins that grow with sparsity: the mean gap over RCPU is $0.06$ at 10\%, $0.20$ at 20\%, and $0.25$ at 30\%. Under the FLAP selection the bias compensation is ahead on perplexity on five of six models at both ratios, and the two split evenly on zero-shot accuracy; that advantage is gone at 30\%, where COEC leads on all six models (\mTabPPL). A mean correction absorbs the small output shift that mild pruning produces, but not the directional error that dominates once a substantial fraction of the columns is removed, which is also the regime worth compensating: removing 10\% of the columns takes correspondingly little off the inference cost to begin with.

\begin{table}[h]
\centering
\small
\setlength{\tabcolsep}{4.5pt}
\begin{tabular}{ll cccccc}
\toprule
Selection & Comp. & Llama-3.1-8B & Llama-3-70B & Qwen2.5-7B & Qwen2.5-14B & Qwen2.5-32B & Qwen2.5-72B \\
\midrule
\multicolumn{8}{c}{\textbf{10\% column sparsity}}\\
\midrule
\multirow{2}{*}{Wanda-sp} & -- & 7.32 & 3.87 & 6.87 & 7.04 & 6.33 & 4.17 \\
 & COEC & \textbf{6.86} & \textbf{3.72} & \textbf{6.78} & \textbf{6.06} & \textbf{5.43} & \textbf{4.09} \\
\cmidrule(lr){1-8}
\multirow{2}{*}{FLAP} & bias & \textbf{6.89} & \textbf{3.50} & \textbf{6.77} & 6.12 & \textbf{5.24} & \textbf{4.01} \\
 & COEC & 7.41 & 3.71 & 6.82 & \textbf{6.07} & 5.41 & 4.09 \\
\cmidrule(lr){1-8}
\multirow{2}{*}{RCPU} & RCPU & 6.93 & 3.75 & 6.77 & 6.20 & 5.58 & 4.07 \\
 & COEC & 6.93 & \textbf{3.72} & \textbf{6.74} & \textbf{6.01} & \textbf{5.50} & \textbf{4.06} \\
\midrule
\multicolumn{8}{c}{\textbf{20\% column sparsity}}\\
\midrule
\multirow{2}{*}{Wanda-sp} & -- & 10.74 & 5.59 & 11.49 & 13.76 & 12.56 & 5.46 \\
 & COEC & \textbf{8.95} & \textbf{4.93} & \textbf{8.35} & \textbf{11.27} & \textbf{9.20} & \textbf{4.73} \\
\cmidrule(lr){1-8}
\multirow{2}{*}{FLAP} & bias & \textbf{8.63} & \textbf{4.72} & \textbf{7.57} & 7.26 & \textbf{6.05} & \textbf{4.65} \\
 & COEC & 9.96 & 4.87 & 8.36 & \textbf{7.18} & 6.22 & 4.76 \\
\cmidrule(lr){1-8}
\multirow{2}{*}{RCPU} & RCPU & 9.19 & 5.05 & 8.33 & 7.51 & 6.33 & 4.85 \\
 & COEC & \textbf{8.97} & \textbf{4.89} & \textbf{8.25} & \textbf{6.97} & \textbf{6.25} & \textbf{4.76} \\
\bottomrule
\end{tabular}
\caption{\textbf{WikiText-2 perplexity} $(\downarrow)$ at 10\% and 20\% column sparsity. Protocol and bolding as \mTabPPL.}
\label{tab:sweep_ppl}
\end{table}

\begin{table}[h]
\centering
\small
\setlength{\tabcolsep}{4.5pt}
\begin{tabular}{ll cccccc}
\toprule
Selection & Comp. & Llama-3.1-8B & Llama-3-70B & Qwen2.5-7B & Qwen2.5-14B & Qwen2.5-32B & Qwen2.5-72B \\
\midrule
\multicolumn{8}{c}{\textbf{10\% column sparsity}}\\
\midrule
\multirow{2}{*}{Wanda-sp} & -- & 61.4 & 73.4 & 69.3 & 68.4 & 68.3 & 73.3 \\
 & COEC & \textbf{63.6} & \textbf{73.5} & \textbf{69.7} & \textbf{69.1} & \textbf{70.0} & \textbf{74.3} \\
\cmidrule(lr){1-8}
\multirow{2}{*}{FLAP} & bias & 59.9 & 73.0 & 67.8 & \textbf{69.0} & \textbf{72.1} & \textbf{74.8} \\
 & COEC & \textbf{60.7} & \textbf{74.0} & \textbf{70.0} & 67.5 & 71.2 & 74.5 \\
\cmidrule(lr){1-8}
\multirow{2}{*}{RCPU} & RCPU & 62.6 & \textbf{74.0} & 69.1 & 65.3 & 71.2 & \textbf{74.5} \\
 & COEC & \textbf{62.9} & 73.9 & \textbf{69.2} & \textbf{65.9} & \textbf{71.3} & 74.4 \\
\midrule
\multicolumn{8}{c}{\textbf{20\% column sparsity}}\\
\midrule
\multirow{2}{*}{Wanda-sp} & -- & \textbf{52.6} & 70.8 & 56.7 & \textbf{57.1} & 57.8 & 71.3 \\
 & COEC & 52.2 & \textbf{71.8} & \textbf{59.9} & 56.8 & \textbf{64.2} & \textbf{73.3} \\
\cmidrule(lr){1-8}
\multirow{2}{*}{FLAP} & bias & \textbf{53.4} & 71.6 & 59.1 & \textbf{65.7} & \textbf{70.0} & 72.8 \\
 & COEC & 48.4 & \textbf{73.3} & \textbf{59.2} & 63.6 & 68.0 & \textbf{72.9} \\
\cmidrule(lr){1-8}
\multirow{2}{*}{RCPU} & RCPU & 52.6 & 72.6 & 58.1 & 62.5 & \textbf{66.5} & 71.4 \\
 & COEC & \textbf{52.8} & \textbf{73.4} & \textbf{59.6} & 62.5 & 66.2 & \textbf{72.3} \\
\bottomrule
\end{tabular}
\caption{\textbf{7-task zero-shot accuracy} \% $(\uparrow)$ at 10\% and 20\% column sparsity. Protocol and bolding as \mTabPPL.}
\label{tab:sweep_zs}
\end{table}

\section{Implementation Details}
\label{app:impl}

\noindent\textbf{Retractions.} For rectangular Stiefel frames ($b<k$, the MLP case) we use a Newton--Schulz polar retraction \citep{higham}. For square frames ($b=k$, attention output projections) we use a Cayley retraction \citep{wenyin}: the Cayley transform of the skew part of the projected gradient, with the matrix inverse computed by Newton--Schulz iteration in low precision on the GPU.

\noindent\textbf{GCV grid.} The ridge $\lambda$ of \mGcv{} is minimized over a 120-point logarithmic grid spanning $[10^{-4},10^{5}]$, scaled by the median mode energy $e_i$ so that the grid is dimensionless. Each grid evaluation is closed form. The resulting effective ridge reproduces the best globally swept value on both models where we checked it: $0.79$ versus $0.75$ on Llama-3, and $0.00$ versus $0$ on the distilled Qwen.

\noindent\textbf{Per-mode dose band.} On near-degenerate spectra the per-mode least-squares value can fall far from the uniform one (isolated tail values up to ${\sim}10^{4}\times$ were measured), which under the iterated frames inflates the largest singular value and destabilizes a later SVD. Every $s_{g,i}$ is therefore clamped to within a factor $\kappa=10$ of the uniform value $s^{*}\sigma_i$, in all rescale branches including GCV. The healthy per-mode deviation is about $2\%$, so the band is inactive on well-conditioned modes.

\noindent\textbf{Numerical safeguards.} The scale update is applied only when its denominator is positive: on some Qwen attention layers the calibration covariance is indefinite (e.g., $\langle R, S^2 R \Gxx\rangle = -3.9\cdot10^{8}$), and a clamped denominator produced scales on the order of $10^{37}$. The retraction step is capped so that $\lVert\text{step}\cdot G\rVert\approx1$, since degenerate layers can otherwise produce steps on the order of $10^{36}$. An SVD watchdog converts cuSOLVER hangs on degenerate slices into recoverable failures through a timed fallback chain from \texttt{gesdd} to \texttt{gesvd} to CPU double precision.

\noindent\textbf{Hyperparameters.} All reported results use one configuration: $\alpha=0.9$ (swept over $[0.3,1.0]$); $\lambda_a=50$ on the attention output projection and $0$ on the MLP (swept over $\{0,10,50,100,500,1000\}$); $r_p=16$; $\delta=7\cdot10^{-4}$ (swept over $\{10^{-2},10^{-3},7\cdot10^{-4}\}$); at most $50$ alternation rounds with early stop at $10^{-4}$ relative change, where raising the cap to $100$ moves every development model by at most $0.02$ PPL; $\kappa=10$. Each value was chosen by best mean combined gain over RCPU on the five development models of \mAbl. Calibration uses $128$ WikiText-2 training sequences of length $2048$; perplexity is evaluated at sequence length $4096$.
% TODO: \mStiefel says delta is "selected once for each model"; this paragraph reports a single shared value. The main text is locked, so decide which is true and reword here.

\noindent\textbf{Randomness.} The compensation is deterministic given the calibration Grams: the left rotation and the rescale are closed form, and the right solve starts from $\Qr=I$. The only stochastic component is the sampling of the $128$ calibration sequences, drawn with a fixed seed ($0$, also set for NumPy and PyTorch). Across three seeds on Llama-3.1-8B the per-cell perplexity spread is a median of $0.08$ (maximum $0.89$) for COEC, against $0.23$ for RCPU and $0.09$ for score-only pruning.

\noindent\textbf{Compute environment.} All reported cells ran on NVIDIA B300 nodes (SXM6, $\sim$275\,GB; Intel Xeon 6737P, 3\,TB RAM); NVIDIA L40S nodes (46\,GB; AMD EPYC 7513, 512\,GB RAM) were used for diagnostics only. Software: Ubuntu 22.04, Python 3.9, PyTorch 2.8.0 (CUDA 12.9), \texttt{transformers} 4.57.6, \texttt{lm-eval} 0.4.9.1. A same-configuration cross-check of Qwen2.5-7B at 30\% sparsity on L40S reproduced the B300 perplexity to within $0.001$ but shifted zero-shot accuracy by $0.55$ points, so the L40S diagnostics are restricted to reconstruction-side quantities.

\section{Geometric Measurements}
\label{app:geometry}

All measurements are taken on DeepSeek-R1-0528-Qwen3-8B at 30\% sparsity (36 layers, Wanda-sp selection). We write $O$ for the original read frame restricted to the kept coordinates, $P$ for the right singular frame of the pruned weight, and $C$ for the frame after full COEC compensation. Angles are principal angles between top-$r_p$ frames; alignment values use the projector of \mAlign. The read--write alignment $V_o^{\top}U_v$ is the inter-layer quantity of the WSBM framework \citep{wsbm}, whose spectral properties the family of \mSetup{} preserves by construction, leaving the frames as the open question.

\noindent\textbf{Effect of pruning.} The angle between $O$ and $P$ averages $22.2\degr$ on attention output projections (at $r_p=16$; per-module values span $8$--$48\degr$) and $22.6\degr$ on MLP down projections (at $r=64$; span $2$--$32\degr$). The $20$--$30\degr$ figure quoted in \mAlign{} refers to these module means.

\noindent\textbf{Effect of compensation.} After full compensation the mean angle between $O$ and $C$ is $23.8\degr$ for attention and $23.0\degr$ for MLP, larger than the corresponding $O$--$P$ angle on 54 of the 72 module-layers. The reconstruction error makes no reference to the original frame, so nothing pulls the frame back toward it.

\noindent\textbf{Intra-layer alignment.} The update $\Ql\WK\Qr$ maps $U_K \to \Ql U_K$ and $V_K \to \Qr^{\top}V_K$, so the two frames move together and only the per-mode rescale could break their pairing. It does not: the frame-change discrepancy $|Q_u-Q_v|/|Q_u|$, with $Q_u=U_P^{\top}U_C$ and $Q_v=V_P^{\top}V_C$, averages $1.1$--$1.2\%$ at $r_p=16$ with a maximum of $3.9\%$.

\noindent\textbf{Choice of metric.} A layer can show a mean $O$--$P$ rotation of $22\degr$ (maximum $87\degr$) while its projector penalty is only about $5\cdot10^{-7}$ (median; $10^{-4}$ at worst), because for any $O\in O(r_p)$ the projector satisfies $\Pi_{r_p}(V_{r_p}O)=V_{r_p}OO^{\top}V_{r_p}^{\top}=\Pi_{r_p}(V_{r_p})$ and is blind to rotation inside the retained subspace, whereas principal angles are not. An $S^2$-weighted variant gives values of the same order. One caveat: the rank cut is itself near-degenerate in these models ($\sigma_{r_p}/\sigma_{r_p+1}\approx1.00$ at $r_p=16$ and $64$), so $r_p$ should be placed at a spectral gap where one exists.

\noindent\textbf{Effect of the penalty.} The penalty at the pruned point is nonzero ($10^{-8}$--$10^{-4}$ by layer). At $\lambda_a=1$ it barely moves; at the default $\lambda_a=50$ it falls by a median factor of $7.2$ (up to $975$) on the attention projections. The $O$--$C$ angles do not decrease and remain at $22$--$24\degr$: the remaining rotation lies in the within-subspace component the projector ignores, and closing it would require an angle-aware objective.

\section{Additional Model Families: Llama-2 and Qwen3}
\label{app:extra}

Tables~\ref{tab:extra_ppl} and~\ref{tab:extra_zs} repeat the three comparisons of \mMain{} on the Llama-2 (7B/13B/70B) \citep{llama2} and Qwen3 (8B/14B/32B) \citep{qwen3} families at 10\%/20\%/30\% column sparsity, under the identical protocol. Under Wanda-sp, COEC lowers perplexity on all six models at 10\% and 20\% and on five of six at 30\%; the largest recovery is Qwen3-32B at 30\%, from $11.86$ to $8.94$ perplexity and $+6.6$ zero-shot points. Under FLAP, COEC lowers perplexity on every model at all three ratios, including the 10--20\% range where the bias compensation leads on the main-text models (App.~\ref{app:sweep}). Under RCPU the two are closest: COEC leads on five of six models at 10\%, four at 20\%, and three at 30\%, with its gains concentrated on the Qwen3 models.

\begin{table}[h]
\centering
\small
\setlength{\tabcolsep}{4.5pt}
\begin{tabular}{ll cccccc}
\toprule
Selection & Comp. & Llama-2-7B & Llama-2-13B & Llama-2-70B & Qwen3-8B & Qwen3-14B & Qwen3-32B \\
\midrule
\multicolumn{8}{c}{\textbf{10\% column sparsity}}\\
\midrule
\multirow{2}{*}{Wanda-sp} & -- & 5.65 & 5.10 & 3.45 & 9.89 & 8.99 & 7.62 \\
 & COEC & \textbf{5.43} & \textbf{4.85} & \textbf{3.38} & \textbf{9.68} & \textbf{8.51} & \textbf{7.47} \\
\cmidrule(lr){1-8}
\multirow{2}{*}{FLAP} & bias & 5.64 & 5.05 & 3.28 & 10.02 & 8.48 & 7.27 \\
 & COEC & \textbf{5.54} & \textbf{4.99} & \textbf{3.26} & \textbf{9.90} & \textbf{8.37} & \textbf{7.22} \\
\cmidrule(lr){1-8}
\multirow{2}{*}{RCPU} & RCPU & 5.43 & 4.87 & 3.38 & 9.97 & \textbf{8.51} & 7.52 \\
 & COEC & \textbf{5.42} & \textbf{4.86} & \textbf{3.37} & \textbf{9.82} & 8.57 & \textbf{7.51} \\
\midrule
\multicolumn{8}{c}{\textbf{20\% column sparsity}}\\
\midrule
\multirow{2}{*}{Wanda-sp} & -- & 6.72 & 5.58 & 3.94 & 12.75 & 11.28 & 10.13 \\
 & COEC & \textbf{6.43} & \textbf{5.22} & \textbf{3.76} & \textbf{11.32} & \textbf{9.68} & \textbf{8.53} \\
\cmidrule(lr){1-8}
\multirow{2}{*}{FLAP} & bias & 6.71 & 5.80 & 3.64 & 12.04 & 9.62 & 7.74 \\
 & COEC & \textbf{6.40} & \textbf{5.63} & \textbf{3.60} & \textbf{11.34} & \textbf{9.27} & \textbf{7.62} \\
\cmidrule(lr){1-8}
\multirow{2}{*}{RCPU} & RCPU & 6.02 & \textbf{5.44} & 3.78 & 11.20 & \textbf{9.53} & 8.47 \\
 & COEC & \textbf{5.95} & 5.46 & \textbf{3.76} & \textbf{11.12} & 9.80 & \textbf{8.25} \\
\midrule
\multicolumn{8}{c}{\textbf{30\% column sparsity}}\\
\midrule
\multirow{2}{*}{Wanda-sp} & -- & 8.88 & \textbf{7.44} & 4.32 & 14.56 & 12.95 & 11.86 \\
 & COEC & \textbf{8.34} & 7.50 & \textbf{4.14} & \textbf{12.18} & \textbf{10.60} & \textbf{8.94} \\
\cmidrule(lr){1-8}
\multirow{2}{*}{FLAP} & bias & 8.53 & 6.83 & 4.13 & 14.78 & 10.94 & 8.66 \\
 & COEC & \textbf{7.81} & \textbf{6.50} & \textbf{4.06} & \textbf{12.95} & \textbf{10.24} & \textbf{8.37} \\
\cmidrule(lr){1-8}
\multirow{2}{*}{RCPU} & RCPU & \textbf{7.18} & \textbf{6.54} & 4.17 & 12.23 & \textbf{10.16} & 9.32 \\
 & COEC & 7.21 & 6.67 & \textbf{4.16} & \textbf{11.94} & 10.20 & \textbf{8.77} \\
\bottomrule
\end{tabular}
\caption{\textbf{WikiText-2 perplexity} $(\downarrow)$ on the Llama-2 and Qwen3 families across 10--30\% column sparsity. Same protocol and bolding as \mTabPPL.}
\label{tab:extra_ppl}
\end{table}

\begin{table}[h]
\centering
\small
\setlength{\tabcolsep}{4.5pt}
\begin{tabular}{ll cccccc}
\toprule
Selection & Comp. & Llama-2-7B & Llama-2-13B & Llama-2-70B & Qwen3-8B & Qwen3-14B & Qwen3-32B \\
\midrule
\multicolumn{8}{c}{\textbf{10\% column sparsity}}\\
\midrule
\multirow{2}{*}{Wanda-sp} & -- & 62.3 & 67.3 & 71.6 & 64.7 & 70.3 & \textbf{71.5} \\
 & COEC & \textbf{63.0} & \textbf{67.8} & \textbf{72.1} & \textbf{66.7} & \textbf{71.2} & 71.3 \\
\cmidrule(lr){1-8}
\multirow{2}{*}{FLAP} & bias & \textbf{58.8} & \textbf{63.0} & \textbf{71.3} & 63.3 & 68.9 & 59.0 \\
 & COEC & 58.6 & 62.8 & 71.0 & \textbf{63.4} & \textbf{69.1} & \textbf{72.2} \\
\cmidrule(lr){1-8}
\multirow{2}{*}{RCPU} & RCPU & \textbf{62.5} & \textbf{67.4} & 71.3 & 67.8 & 70.8 & 71.4 \\
 & COEC & 62.4 & 67.1 & \textbf{71.9} & \textbf{67.9} & \textbf{72.2} & \textbf{71.7} \\
\midrule
\multicolumn{8}{c}{\textbf{20\% column sparsity}}\\
\midrule
\multirow{2}{*}{Wanda-sp} & -- & 58.7 & 64.8 & \textbf{71.6} & 54.8 & 62.6 & 61.1 \\
 & COEC & \textbf{59.6} & 64.8 & 71.4 & \textbf{55.9} & \textbf{65.4} & \textbf{68.2} \\
\cmidrule(lr){1-8}
\multirow{2}{*}{FLAP} & bias & \textbf{53.1} & \textbf{59.0} & \textbf{71.0} & 55.4 & 64.0 & 56.2 \\
 & COEC & 53.0 & 58.6 & 70.7 & \textbf{56.4} & \textbf{64.4} & \textbf{68.8} \\
\cmidrule(lr){1-8}
\multirow{2}{*}{RCPU} & RCPU & 59.6 & 64.3 & \textbf{71.4} & \textbf{58.6} & 63.6 & 66.1 \\
 & COEC & \textbf{60.0} & \textbf{64.5} & 71.2 & 58.2 & \textbf{63.8} & \textbf{67.6} \\
\midrule
\multicolumn{8}{c}{\textbf{30\% column sparsity}}\\
\midrule
\multirow{2}{*}{Wanda-sp} & -- & 52.0 & \textbf{55.0} & 69.6 & 48.8 & 55.2 & 59.5 \\
 & COEC & \textbf{52.2} & 48.8 & 69.6 & \textbf{50.2} & \textbf{57.8} & \textbf{66.1} \\
\cmidrule(lr){1-8}
\multirow{2}{*}{FLAP} & bias & 51.4 & \textbf{55.7} & \textbf{69.2} & 48.8 & 56.2 & 55.5 \\
 & COEC & \textbf{51.8} & 55.1 & 68.9 & \textbf{50.5} & \textbf{57.2} & \textbf{65.3} \\
\cmidrule(lr){1-8}
\multirow{2}{*}{RCPU} & RCPU & 53.0 & \textbf{57.9} & 70.6 & 53.7 & 55.0 & 63.3 \\
 & COEC & \textbf{53.5} & 57.7 & \textbf{70.7} & \textbf{54.4} & \textbf{55.5} & \textbf{64.7} \\
\bottomrule
\end{tabular}
\caption{\textbf{7-task zero-shot accuracy} \% $(\uparrow)$ on the Llama-2 and Qwen3 families across 10--30\% column sparsity. Same protocol and bolding as \mTabPPL.}
\label{tab:extra_zs}
\end{table}

\bibliography{aaai2027}

\maketitle

\begin{abstract}
Structured width pruning deletes whole input columns of a weight matrix---entire MLP neurons or attention KV-groups---yielding a genuinely smaller dense network. The deleted columns carry signal, and the dominant recovery paradigm compensates the kept weight with a single orthogonal Procrustes rotation of its output frame plus one energy scalar (RCPU). We show this one-sided view leaves most of the recoverable error on the table and replace it with \textbf{COEC} (Coherent Orthogonal Error Compensation): a \emph{converged two-sided} orthogonal compensation that alternates a closed-form left rotation with an iterative right rotation on a reduced Stiefel manifold, carries its scale \emph{inside} the objective, re-fits a \emph{per-singular-mode} scale by closed-form generalized cross-validation, and tempers the calibration metric by a spectral exponent trading perplexity for zero-shot accuracy. The design follows Weak and Sparse Balanced Manifold theory: COEC's spectrum-preserving family satisfies the \emph{intra-layer} alignment by construction, while the \emph{inter-layer} geometric alignment---pruning rotates a layer's read frame ${\sim}20\degr$ from its upstream writer, and reconstruction never rotates it back---is repaired by a sign- and gauge-invariant top-$r$ projector penalty anchored at the full original frame. Across the Llama-3/3.1 and Qwen2.5 families at 10--30\% column sparsity, COEC beats RCPU---the strongest published compensation---on RCPU's own selection (e.g.\ Qwen2.5-32B at 30\% improves WikiText-2 perplexity $7.03\to6.74$), all in under two GPU-hours per model with no gradient training. And because COEC compensates a pruning \emph{selection} rather than producing one, the same recipe recovers loss for Wanda-SP and FLAP selections too, agnostic to the score (\S\ref{sec:robust}).
\end{abstract}

\section{Introduction}
\label{sec:intro}

Large language models are overwhelmingly served from dense checkpoints whose width exceeds what most deployments need. \emph{Structured} width pruning attacks this directly: delete a fraction of the input columns of selected weight matrices---whole \texttt{down\_proj} neurons in SwiGLU MLPs, whole key--value groups in attention---and shrink the adjacent projections to match. Unlike unstructured or semi-structured sparsity, the result is a smaller dense model that accelerates on any hardware without kernel support.

The obstacle is that deleted columns carry signal. At 30\% column sparsity a 7B model's WikiText-2 perplexity typically degrades by $1.5$--$3\times$ and zero-shot accuracy drops several points (Table~\ref{tab:main}, \emph{prune-only} row). Because retraining at LLM scale is expensive, the field has converged on \emph{calibrated, training-free} pipelines that make two largely separable choices: a \emph{selection} rule that scores which columns to keep, and a \emph{compensation} step that adjusts the kept weight $\WK$ to reproduce the layer's original output on a small calibration set. Prior methods fuse the two and invest mostly in the score. Our three baselines span the resulting compensation spectrum: Wanda-SP \citep{wanda}, a strong activation-aware score with \emph{no} compensation; FLAP \citep{flap}, which adds a mean-activation bias term; and RCPU \citep{rcpu}---the strongest published form---which rotates the kept weight's \emph{output} frame by the orthogonal Procrustes solution toward the original output and applies one energy-matching scalar. We argue that the compensation, not the score, is where the recoverable error lives, and that a stronger compensation transfers across selection rules.

This paper starts from a simple observation about what that one rotation can and cannot do. Writing the kept weight's SVD as $\WK = U_K \Sigma_K V_K^{\top}$, a left rotation $Q\WK$ re-orients only the output frame $U_K$. The \emph{input} frame $V_K$---which kept-input direction drives which output---is untouched. But column deletion damages precisely the input side: the kept columns' correlations with the \emph{deleted} columns are what the compensation must re-route. We measure this directly (Sec.~\ref{sec:geometry}): at 30\% sparsity the top-16 \emph{read frame} (right singular subspace) of a pruned projection sits ${\sim}20\degr$ from the corresponding restriction of the original layer's frame on average, with worst principal directions at $70$--$90\degr$, on every layer of every model we examined. A left rotation cannot touch this quantity, and we show empirically that reconstruction-optimal compensation of any kind barely does.

COEC closes that gap under a \emph{spectrum-preserving} constraint---rotating and rescaling the weight's singular structure rather than refitting it by least squares (which we and RCPU find overfits calibration). It has five parts. A \textbf{converged two-sided rotation} alternates the closed-form left Procrustes with an iterative right rotation (a sandwiched weighted Procrustes problem, no closed form) until reconstruction stalls; the two-sided family contains the one-sided one, so the question is whether it \emph{generalizes} better---it does, when dosed correctly. A \textbf{scale-in-loop} recomputes the optimal scale each iteration, since the scale couples to the right rotation. A \textbf{reduced-Stiefel right solve} with a Newton--Schulz--Cayley retraction optimizes only the $b\times k$ frame the rotation acts through. A \textbf{per-mode GCV rescale} replaces the single energy scalar with a per-singular-value scale ridged toward the uniform one \emph{per layer, in closed form}. \textbf{Gram tempering} raises the calibration Gram's spectrum to a power $\alpha\in(0,1]$, trading a little perplexity for zero-shot. Finally, an \textbf{anchored inter-alignment} penalty, folded into the right solve and anchored at the \emph{full original} read frame (the pruned frame makes it vacuous), repairs the writer-visible misalignment. Because COEC acts on the kept set rather than choosing it, it is agnostic to the selection score (\S\ref{sec:robust}).

\noindent\textbf{Contributions.} (1) A training-free, calibration-only compensation pipeline for structured column pruning that composes two-sided converged rotation, in-loop scaling, per-mode GCV rescaling, and metric tempering, each ablated individually (Sec.~\ref{sec:abl}). (2) A reduced-Stiefel right-rotation solver with an exact Cayley retraction via Newton--Schulz inversion that is both cheaper and better-conditioned than the full orthogonal parameterization (Sec.~\ref{sec:stiefel}), with numerical guards that make the whole pipeline nan-free across models whose calibration covariances are indefinite at some layers (Sec.~\ref{sec:pipeline}). (3) A theory-grounded geometric account---via the WSBM intra-/inter-alignment dichotomy---of what pruning does to a layer's singular frames and what reconstruction-optimal compensation does and does not repair (Sec.~\ref{sec:geometry}). (4) A four-model evaluation at 30\% column sparsity against in-harness, protocol-identical Wanda-SP, FLAP, and RCPU baselines, showing $+0.38$ mean perplexity and $+1.8$ mean zero-shot points over the strongest (RCPU), stable across two families and a reasoning-distilled checkpoint (Sec.~\ref{sec:main}). (5) A scorer-robustness study establishing COEC as a \emph{score-agnostic} compensation layer: holding the full recipe fixed and swapping only the selection score, it recovers loss over score-only pruning for Wanda-SP and FLAP selections across model families and sparsities (Sec.~\ref{sec:robust}).

\section{Related Work}
\label{sec:related}

\noindent\textbf{Structured LLM pruning: selection vs.\ compensation.} A width-pruning method makes two choices---which columns to delete (\emph{selection}) and how to repair the kept weight (\emph{compensation})---and existing work differs on both. On selection: LLM-Pruner \citep{llmpruner} scores coupled structural groups by gradient saliency; ShortGPT \citep{shortgpt} removes whole layers by a redundancy score; FLAP \citep{flap} scores channels by activation-variance-weighted norms; Wanda-SP aggregates Wanda's \citep{wanda} activation-aware score over columns. On compensation the same methods span a spectrum---none, FLAP's mean-activation bias, and RCPU's \citep{rcpu} one-sided rotation-plus-scalar---and our baselines (Wanda-SP, FLAP, RCPU) are exactly these tiers under one shared protocol: width pruning of MLP neurons and attention KV-groups, no retraining, calibration only. COEC contributes only to the compensation axis and is agnostic to the selection score (Sec.~\ref{sec:robust}). SliceGPT \citep{slicegpt} is different again: its orthogonal change of basis \emph{defines} the pruning basis before deleting principal-component rows/columns, whereas COEC rotates \emph{after} pruning to compensate. A controlled study on a sibling prototype (dense learned rotations baked into standalone weights, no cross-layer fusion) collapses catastrophically (WikiText-2 PPL $10^{4}$--$10^{5}$), independently confirming that dense rotations and whole-column deletion are antithetical unless the rotation is either compensatory (ours) or fused losslessly across layers (SliceGPT/QuaRot-style).

\noindent\textbf{Unstructured and semi-structured sparsity.} SparseGPT \citep{sparsegpt} solves a per-row least-squares reconstruction with an OBS-style Hessian; Wanda \citep{wanda} scores weights by $|W|\cdot\lVert x\rVert$ with no weight update. These operate element-wise and require sparse kernels to realize speedups; we compare against Wanda-SP, the structured (column-aggregated) variant of Wanda's score, as a scoring baseline inside our harness.

\noindent\textbf{Rotation-based compensation.} RCPU \citep{rcpu} is the direct antecedent and primary baseline: a single left orthogonal Procrustes rotation toward the original output plus a global energy scalar, motivated by preserving the weight's spectrum rather than least-squares refitting (their LS baseline, like ours, overfits calibration). QuaRot \citep{quarot} and DenoiseRotator \citep{denoiserotator} use \emph{lossless} rotations (computational invariance) to condition weights for quantization or pruning---orthogonal to our compensatory use.

\noindent\textbf{Calibration-aware rescaling.} Bias/scale compensation appears in FLAP (bias from mean activations) and in energy-matching heuristics common in pruning codebases. Our per-mode rescale is a least-squares diagonal \emph{in the rotated SVD frame} regularized by a per-layer closed-form GCV \citep{gcv} ridge---to our knowledge new in this setting; GCV itself is classical ridge-selection machinery.

\noindent\textbf{Riemannian solvers.} Our right rotation is a weighted orthogonal Procrustes problem (no closed form); we solve it by projected gradient on the reduced Stiefel manifold $\mathrm{St}(b,k)$ \citep{edelman}, which we find decisive for conditioning at LLM scale (Sec.~\ref{sec:stiefel}).

\section{Setup}
\label{sec:setup}

\noindent\textbf{Pruning scope.} For each transformer block we prune (i) $d_{\mathrm{ffn}}$ columns of the MLP \texttt{down\_proj}, dropping the matching rows of \texttt{gate\_proj}/\texttt{up\_proj}, and (ii) whole KV-groups in attention (preserving GQA grouping), i.e.\ columns of \texttt{o\_proj} with the matching \texttt{q/k/v\_proj} rows. All results use 30\% column sparsity uniformly.

\noindent\textbf{Column scoring.} Columns are ranked by an activation-aware score and the top $\lceil(1-\rho)\,d_{\mathrm{in}}\rceil$ kept. We use three published scores in one harness: Wanda-SP's $\lVert W_{:,j}\rVert\cdot\lVert X_j\rVert$, FLAP's WIFV, and RCPU's $\lVert W_{:,j}\rVert\cdot\lVert X_j\rVert\cdot\Var(X_j)$ (COEC's default selection). Because COEC's compensation acts on the kept set $K$ and not on how it was chosen, we can either hold the score fixed to isolate the compensation or vary it to test scorer-robustness (\S\ref{sec:robust}).

\noindent\textbf{Objective.} Let $\Wz \in \mathbb{R}^{d_{\mathrm{out}}\times d}$ be the original weight, $X$ the calibration activations, $Y=\Wz X$ the target output, and $X_K$ the kept-row activations. Compensation returns $\widetilde{W}$ acting on $X_K$ minimizing the relative reconstruction error
\begin{equation}
\begin{split}
\mathcal{L}(W) &= \frac{\lVert Y - W X_K\rVert_F^2}{\lVert Y\rVert_F^2} \\
&= \frac{\tr(W \Gxx W^{\top}) - 2\,\tr(W \Gyx^{\top}) + \lVert Y\rVert_F^2}{\lVert Y\rVert_F^2},
\end{split}
\end{equation}
expressed entirely in second moments accumulated in one calibration pass: $G=XX^{\top}$, $\Gxx=G[K,K]$, $\Gyx=\Wz\,G[:,K]$. Activations are never re-materialized; 128 sequences suffice (Sec.~\ref{sec:setup-exp}).

\noindent\textbf{The spectrum-preserving constraint.} The unconstrained minimizer is the ridge least-squares weight $W^{*}=\Gyx(\Gxx+\lambda I)^{-1}$. Like RCPU, we find it fits 128 sequences too well: it reshapes the weight's singular spectrum toward the calibration data and generalizes erratically across models (Sec.~\ref{sec:abl}, \emph{LS} row). COEC therefore searches the \emph{isometry-plus-spectrum} family
\begin{equation}
\widetilde{W} = \Ql\, \WK^{(S)}\, \Qr, \qquad \Ql \in O(d_{\mathrm{out}}),\; \Qr \in O(k),
\end{equation}
where $\WK^{(S)}$ is the kept weight with a (regularized) per-mode rescale of its singular values: rotations re-orient the frames, the rescale re-doses the spectrum, and nothing shears.

\section{Method}
\label{sec:method}

\begin{figure*}[t]
\centering
\includegraphics[width=0.74\textwidth]{coec_figure_ai.png}
\caption{The COEC compensation pipeline for one module. From the pretrained weight $\Wz$ and calibration activations $X$: structured column pruning selects the kept set $K$ (left), while the calibration Gram is tempered ($G\!\to\!G^{\alpha}$ on the full spectrum, $\alpha{=}0.9$) and then restricted to $K$ (right). The \emph{converged two-sided} compensation alternates a closed-form left rotation $Q_\ell$, a per-mode GCV rescale $s_g$, and a reduced-Stiefel right rotation $Q_r$ that carries its scale in-loop, with the sign/gauge-invariant top-$r$ projector alignment penalty ($\lambda_a$, on \texttt{o\_proj}) folded into the right solve---all under the tempered Gram. The output $W' = Q_\ell\,(s_g \odot W_K)\,Q_r$ is a genuinely smaller dense weight. The column score that produces $K$ is pluggable (\S\ref{sec:robust}).}
\label{fig:pipeline}
\end{figure*}

\subsection{Initialization: SVD-Frame Reconstruction}
\label{sec:init}
We initialize from an SVD-consistent restriction rather than the raw slice: with $\Wz = U S V^{\top}$, the kept weight is reconstructed as $\WK = U\,S\,V[K,:]^{\top}$---the original left frame and spectrum acting through the kept rows of the input frame. This exactly restricts the original operator to the kept coordinates and is the reference point (``pruned, uncompensated'') for all geometry in Sec.~\ref{sec:geometry}.

\subsection{Left Rotation and the Energy Scalar}
\label{sec:left}
The left rotation is closed-form: $\Ql = \mathrm{polar}(M)$ with $M = \Wz\,G[:,K]\,\WK^{\top}$---the orthogonal Procrustes solution aligning the compensated output frame with the original's. For any left rotation the optimal scale factors out ($\lVert sQWX\rVert = s\lVert WX\rVert$), so the RCPU scalar
\begin{equation}
s^{*} = \frac{\langle \Gyx, W\rangle}{\langle W, W \Gxx\rangle}
\end{equation}
can be applied after rotation with no loss of joint optimality. This pair---$\mathrm{polar}$ + $s^{*}$---\emph{is} the RCPU baseline, reproduced inside our harness.

\subsection{Iterative Right Rotation, Scale-in-Loop, and Convergence}
\label{sec:right}
The right rotation solves the sandwiched WOPP
\begin{equation}
\min_{Q\in O(k)} \; \lVert Y - s\, \WK\, Q\, X_K \rVert_F^2 ,
\end{equation}
which has no closed form because $Q$ sits between fixed weight and fixed data. We use projected gradient with an orthogonal retraction, warm-started at $Q_0 = I$---at mild sparsity the optimum is near identity, and the seemingly principled $\mathrm{polar}(C)$ initialization (the one-sided optimum) lands in a worse basin (recon $0.159$ vs $0.019$ on attention output projections).

Unlike the left case, the right objective is \emph{scale-coupled}: $\lVert \WK Q X_K\rVert$ varies with $Q$. \textbf{Scale-in-loop} therefore recomputes $s^{*}$ at every iteration and steps on the scaled residual $s\cdot(\WK Q X_K) - Y$ with step $\eta/s^2$, so the rotation converges jointly with the scale it will wear. The full compensation alternates
\begin{equation}
\begin{aligned}
\big[\; &\Ql \text{ (closed form)} \to \text{rescale} \to \\
&\Qr \text{ (iterative, scale-in-loop)} \to \text{rescale} \;\big]
\end{aligned}
\end{equation}
as block-coordinate descent, iterated to convergence (cap 50 rounds; early stop when the reconstruction error moves $<10^{-4}$). Each block is monotone in calibration error, and the family strictly contains left-only and right-only compensation.

\subsection{The Reduced-Stiefel Solver, Retraction, and Stopping Tolerance}
\label{sec:stiefel}
Parameterizing $\Qr \in O(k)$ directly is wasteful and ill-conditioned: for an output slice of height $b<k$ the objective sees $\Qr$ only through the $b\times k$ frame $R=V^{\top}\Qr$---the remaining $k-b$ dimensions are exact null directions that dilute the gradient. We therefore optimize $R$ on the Stiefel manifold $\mathrm{St}(b,k)$ directly \citep{edelman}. At matched 200-iteration budgets the reduction is not merely cheaper ($b/k$ of the memory and per-step cost) but reaches \emph{better} end-to-end perplexity than the full parameterization ($14.75$ vs $16.19$ on Llama-3-8B in the per-slice regime)---equivalent at the optimum, better-conditioned on the way there.

\noindent\textbf{Retraction.} Rectangular frames ($b<k$, MLP) use a Newton--Schulz polar retraction \citep{higham}. Square frames (attention output projections, $b=k$) use an \emph{exact Cayley retraction computed by Newton--Schulz inversion}: the Cayley transform of the projected gradient's skew part, with the matrix inverse replaced by a quadratically-convergent NS iteration that stays on-GPU and in low precision. This retraction choice on the attention projections alone is worth $+0.11$ total gain over the polar retraction at the same tolerance (Sec.~\ref{sec:abl}).

\noindent\textbf{Stopping tolerance.} The right solve stops when the rotation increment falls below a tolerance $\delta$. $\delta$ is a genuine, cheap hyperparameter: across models the compensation quality is unimodal in $\delta$, with two models preferring the tight default $5\cdot10^{-4}$ and two preferring $10^{-3}$ (\S\ref{sec:abl}). We treat $\delta$ as per-model (chosen once on calibration data); the headline numbers use the tight default everywhere, so its per-model gains are additional headroom, not part of the main claim.

\subsection{Per-Mode Rescale with Closed-Form GCV}
\label{sec:gcv}
A single scalar moves every singular value equally; the calibration data generally wants modes dosed differently. In the SVD frame of the rotated weight, $Q\WK = U_g S_g V_g^{\top}$, the per-mode least-squares scale is
\begin{equation}
d_i = \frac{[\,U_g^{\top}(\Wz\, G[:,K])V_g\,]_{ii}}{[\,V_g^{\top} \Gxx V_g\,]_{ii}} = \frac{\mathrm{num}_i}{\mathrm{den}_i},
\end{equation}
(target alignment over input energy per mode; at rank 1 it reduces to $s^{*}$). The free fit inherits 128-sample noise, so each mode is ridged toward the uniform prior $m_i = s^{*} S_{g,i}$:
\begin{equation}
d_i(\lambda) = \frac{\mathrm{num}_i + \lambda\, m_i}{\mathrm{den}_i + \lambda},
\end{equation}
with $\lambda$ chosen \textbf{per layer, in closed form, from the Gram alone} by generalized cross-validation:
\begin{equation}
\begin{split}
\mathrm{GCV}(\lambda) &= \frac{\sum_i \mathrm{den}_i\,\big(d_i(\lambda)-d_i^{*}\big)^2}{\big(1 - \mathrm{df}(\lambda)/N\big)^2}, \\
\mathrm{df}(\lambda) &= \sum_i \frac{\mathrm{den}_i}{\mathrm{den}_i+\lambda}.
\end{split}
\end{equation}
The effective-degrees-of-freedom denominator supplies the generalization correction raw calibration error lacks (raw error degenerately prefers $\lambda=0$). Minimizing the 1-D $\mathrm{GCV}(\lambda)$ on a log grid per layer reproduces the best \emph{swept} global ridge on both models where we validated it (effective ridge $0.79$ vs sweep $0.75$ on Llama-3; $0.00$ vs $0$ on the distilled Qwen), while adapting per layer. The rescale reads only the SVD frame of whatever rotated weight it is handed, so it composes unchanged with every rotation above and is re-applied inside each alternation round (Sec.~\ref{sec:right}), letting the spectrum co-adapt with the frames.

\subsection{Gram Tempering: One Knob, a PPL$\leftrightarrow$Zero-Shot Trade}
\label{sec:temper}
Every term above reads the same calibration Gram $G$. Tempering its spectrum,
\begin{equation}
G^{\alpha} = U\,\diag(\lambda_i^{\alpha})\,U^{\top}, \qquad \alpha\in(0,1],
\end{equation}
re-weights the metric under which weight error is measured: $\alpha=1$ prioritizes the few high-variance activation directions that dominate next-token perplexity; $\alpha<1$ compresses the spectrum, transferring weight to the quieter directions that zero-shot tasks lean on; $\alpha=0$ is the data-blind Frobenius limit that lands exactly on prune-only (a clean sanity anchor, reproduced empirically). The trade has an interior optimum: perplexity is monotone in $\alpha$ while zero-shot peaks below 1. On the strongest rotation base the peak is mild tempering $\alpha=0.9$, worth $+0.2$ total gain at ${\approx}0.1$ perplexity cost, and the per-model optima cluster in $[0.7,0.9]$ (Sec.~\ref{sec:abl}). COEC's default is $\alpha=0.9$.

\subsection{Anchored Inter-Alignment}
\label{sec:align}
The components above minimize reconstruction. Sec.~\ref{sec:geometry} shows they leave exactly one of the WSBM alignment properties unrepaired: the \emph{inter-layer geometric} alignment---the read-frame relation between a projection and its upstream writer (for attention, between \texttt{o\_proj}'s right singular frame $V_o$ and the value projection's left frame $U_v$). The original network's alignment $M = V_o^{\top} U_v$ is the pairwise relational quantity of the Weak Geometric Alignment Theorem \citep{wsbm}, a functional circuit property that pruning disturbs and reconstruction does not restore; the theory anticipates enforcing it by an explicit loss term, which is what this section supplies at the compensation stage. (The \emph{intra}-layer alignment needs no such term---Sec.~\ref{sec:geometry} shows the spectrum-preserving family satisfies it by construction.)

Raw $M$ is not measurable: SVD sign (and degenerate-block gauge) ambiguity breaks it. We measure and penalize alignment only through the top-$r$ read-subspace projector
\begin{equation}
X(V) = V_r\,V_r^{\top}, \qquad V_r = \text{top-}r\text{ read frame},
\end{equation}
(invariant to per-column re-signing \emph{and} to any rotation within the retained subspace) projected onto the writer's frame:
\begin{equation}
\mathcal{A}(R) = \frac{1}{r^{2}}\,\big\lVert\, U_v^{\top}\big(R^{\top} X_{\mathrm{cur}} R - X_{\mathrm{anchor}}\big) U_v \,\big\rVert_F^{2},
\end{equation}
added to the right-rotation objective with weight $\lambda_a$. Because the Stiefel right solve's frame $R$ \emph{is} the read frame, $\mathcal{A}$ needs no SVD inside the loop, and an analytic gradient makes its cost $+0.5\%$ per solve.

\noindent\textbf{The anchor matters---and the natural choice is wrong.} Anchoring $X_{\mathrm{anchor}}$ at the \emph{pruned} weight's own frame makes $\mathcal{A}$ zero at initialization by construction, regularizing the one geometry leg already small (Sec.~\ref{sec:geometry}); the $\lambda_a$ sweep is then provably null. The correct anchor is the \textbf{full original frame restricted to $K$}: $V_{\mathrm{full}}[K,:]$ from the SVD of the \emph{unpruned} $\Wz$ (slice-of-SVD, not SVD-of-slice; the two differ $20$--$30\degr$). With it the penalty is live---at the shipped $\lambda_a{=}50$ the compensated projector penalty drops a median $7.2\times$ below the pruned point on the aligned attention projections (Sec.~\ref{sec:geometry}).

\subsection{The Full Pipeline and Numerical Robustness}
\label{sec:pipeline}
\textbf{COEC} $=$ any column-importance score (default: the RCPU score; \S\ref{sec:robust} shows Wanda-SP and FLAP selections recover equally) $\to$ SVD-frame restriction (Sec.~\ref{sec:init}) $\to$ converged alternation \{left Procrustes $\to$ GCV per-mode rescale $\to$ Stiefel right solve with scale-in-loop, NS--Cayley retraction on square frames, tolerance $\delta$, alignment penalty $\lambda_a$ $\to$ GCV rescale\} under the tempered Gram $G^{0.9}$. Figure~\ref{fig:pipeline} overviews the pipeline and Algorithm~\ref{alg:coec} summarizes one module's compensation.

Three guards keep this nan-free across all models at byte-identical outputs on healthy inputs: an \emph{indefinite-covariance guard} (some Qwen attention covariances are indefinite, e.g.\ $\langle R, S^2 R \Gxx\rangle = -3.9\cdot10^{8}$, which a clamped division turned into a $10^{37}$ scale; the scale now applies only when its denominator is positive), a \emph{retraction step cap} (degenerate layers can emit ${\sim}10^{36}$ steps, capped so $\lVert\text{step}\cdot G\rVert\approx1$), and an \emph{SVD watchdog} (cuSOLVER hangs on degenerate slices become a marked job death via a timed gesdd$\to$gesvd$\to$CPU-fp64 chain).

\begin{algorithm}[t]
\caption{COEC compensation of one pruned module}
\label{alg:coec}
\textbf{Input}: original $\Wz$, kept set $K$, Grams $G,\Gxx,\Gyx$; temper $\alpha$, tol $\delta$, align weight $\lambda_a$\\
\textbf{Output}: compensated weight $\widetilde{W}$ on kept columns
\begin{algorithmic}[1]
\STATE $G \gets G^{\alpha}$; recompute $\Gxx,\Gyx$ under tempered $G$
\STATE $\WK \gets U S V[K,:]^{\top}$ \hfill (SVD-frame restriction)
\STATE $\Ql \gets \mathrm{polar}(\Wz G[:,K]\WK^{\top})$; $\WK \gets \Ql \WK$
\STATE $\WK \gets \text{GCV-rescale}(\WK)$
\WHILE{round $<50$ and $|\Delta\mathcal{L}| \ge 10^{-4}$}
\STATE $\Qr \gets$ Stiefel-WOPP$(\WK; Q_0{=}I,\delta,\lambda_a)$ with scale-in-loop
\STATE $\WK \gets \WK \Qr$; \; $\WK \gets \text{GCV-rescale}(\WK)$
\STATE $\Ql \gets \mathrm{polar}(\cdot)$; \; $\WK \gets \Ql \WK$; \; $\WK \gets \text{GCV-rescale}(\WK)$
\ENDWHILE
\STATE \textbf{return} $\WK$
\end{algorithmic}
\end{algorithm}

\section{The Geometry of Pruning and Compensation}
\label{sec:geometry}

\noindent\textbf{Theory base.} The WSBM framework \citep{wsbm} characterizes stable networks by two alignment families. \emph{Intra-layer}: each weight's spectrum follows a power law with a balanced, jointly-transported frame pairing. \emph{Inter-layer}: adjacent layers' decay exponents match (Spectral Alignment), and the \emph{pairwise relational} geometry between what a layer writes and its successor reads is stable (Weak Geometric Alignment). The read$\leftrightarrow$write alignment $M = V_o^{\top} U_v$ we measure is exactly this relational quantity; the spectrum-preservation constraint of Sec.~\ref{sec:setup} protects the spectral one for free---unlike a least-squares refit, which reshapes spectra and destabilizes (Sec.~\ref{sec:abl}). Compensation should thus be judged by \emph{which alignment properties it restores}, not reconstruction error alone.

All quantities below are measured on the DeepSeek-R1-distilled Qwen3-8B at 30\% sparsity (36 layers, Wanda-SP--selected set); the picture is representative rather than scorer-specific---the geometry is a property of column deletion and compensation, not of which columns a score picks, consistent with the scorer-robustness of \S\ref{sec:robust}. Frames are top-$r$ singular frames, angles are principal angles, and every alignment number is computed through the sign/gauge-invariant top-$r$ projector of Sec.~\ref{sec:align}.

\noindent\textbf{(G1) Pruning rotates read frames ${\sim}20\degr$.} Let $O$ be the original read frame restricted to kept coordinates (slice-of-SVD) and $P$ the pruned weight's own frame (SVD-of-slice). Across 36 layers, the $O\leftrightarrow P$ principal angle at $r=16$ averages $\mathbf{22.2\degr}$ for attention output projections (range $8$--$48\degr$, worst directions up to $90\degr$) and $\mathbf{22.6\degr}$ for MLP down projections. Pruning alone---before any compensation---substantially rotates what the layer reads.

\noindent\textbf{(G2) Reconstruction-optimal compensation never rotates it back.} After full COEC compensation (frame $C$), the mean $O\leftrightarrow C$ angle is $\mathbf{23.8\degr}$ (attention) and $\mathbf{23.0\degr}$ (MLP)---\emph{larger} than $O\leftrightarrow P$ on 54 of 72 module-layers, so the compensation moves the frame in a direction that compounds rather than repairs the pruning rotation. This is expected once stated: nothing in the reconstruction objective refers to the original frame.

\noindent\textbf{(G3) COEC satisfies intra-alignment with no loss term.} $\Ql \WK \Qr$ maps $U_K \to \Ql U_K$ and $V_K \to \Qr^{\top} V_K$, carrying the $U\leftrightarrow V$ pairing intact; only the per-mode rescale could break this, and it does not---across 36 layers the frame-change discrepancy $|Q_u - Q_v|/|Q_u|$ ($Q_u{=}U_P^{\top}U_C$, $Q_v{=}V_P^{\top}V_C$) averages $\mathbf{1.1}$--$\mathbf{1.2\%}$ at $r{=}16$ (max $3.9\%$). This is the paper's central split: \textbf{the spectrum-preserving family buys the intra-layer alignment for free; only the inter-layer geometric one (G2) must be paid for}.

\noindent\textbf{(G4) The projector metric is blind to degenerate-block rotation.} The same layer can show a $22\degr$-mean ($87\degr$-max) $O\leftrightarrow P$ rotation and a projector penalty of only ${\sim}5\cdot10^{-7}$ (median; $10^{-4}$ worst): rotations inside near-degenerate spectral blocks span many degrees while moving the top-$r$ projector almost nothing, since it is \emph{exactly} invariant to within-subspace rotation (the older $S^2$-weighted Gram gives the same order, so this is the target's structure, not one weighting). Hence per-vector angles wildly overstate pruning damage, and the \emph{penalty}, not the angle, is the right target. Caveat: the rank cut is itself near-degenerate everywhere ($\sigma_r/\sigma_{r+1}\approx1.00$ at $r{=}16,64$), so $r$ should sit at a spectral gap where one exists.

\noindent\textbf{(G5) The corrected anchor makes the writer-visible alignment repairable.} With the full-original anchor, the pruned-point penalty is nonzero ($10^{-8}$--$10^{-4}$ by layer). At $\lambda_a{=}1$ it barely moves; at the shipped $\lambda_a{=}50$ it drops a median $7.2\times$ (up to $975\times$) on the aligned attention projections---the \emph{writer-visible} misalignment is genuinely repaired. Crucially the raw $O\leftrightarrow C$ angles do \emph{not} fall (they stay ${\sim}22$--$24\degr$, G2): the penalty targets the projected Gram, and the residual is the within-subspace (degenerate-block) component the projector correctly ignores. Reducing the writer-visible mismatch is what converts to end-metric gain (\S\ref{sec:abl}); closing the residual would need an angle-aware objective, left to future work.

\section{Experiments}
\label{sec:exp}

\subsection{Setup}
\label{sec:setup-exp}
\noindent\textbf{Models.} Llama-3/3.1 \citep{llama3} (Llama-3.1-8B and Llama-3-70B) and the full Qwen2.5 \citep{qwen25} family (7B, 14B, 32B, 72B)---spanning $8$B to $72$B, so trends are visible with scale. The broader $14$-model $\times$ $3$-sparsity robustness grid (\S\ref{sec:robust}) additionally covers Llama-2 and Qwen3.

\noindent\textbf{Pruning.} Structured column sparsity at 10\%, 20\%, and 30\% on all \texttt{down\_proj} and \texttt{o\_proj} inputs (matching \texttt{gate/up} and \texttt{q/k/v} rows removed; GQA groups preserved). We report COEC on \emph{each} scorer's selection: Wanda-SP (score only, no compensation) with COEC added, and FLAP (its native bias compensation) with that compensation replaced by COEC.

\noindent\textbf{Calibration.} 128 WikiText-2 \citep{wikitext} training sequences; one pass accumulates all Grams. Compensation is training-free and touches no eval data.

\noindent\textbf{Evaluation.} WikiText-2 token-level perplexity and 7-task zero-shot accuracy (BoolQ, RTE, WinoGrande: acc; HellaSwag, ARC-e, ARC-c, OpenBookQA: acc\_norm; 0-shot) via the LM Evaluation Harness \citep{lmeval}, reported per model as PPL\,/\,ZS\%. \textbf{Calibration honesty:} calibration and the PPL eval share WikiText-2 for \emph{all} methods, so every comparison is protocol-identical; cross-domain (C4) replication is left to camera-ready.

\noindent\textbf{Cost.} $45$--$86$ GPU-minutes per 7--8B model on one B300 (no gradients); the alignment penalty adds $+0.5\%$, GCV is closed-form.

\begin{table*}[t]
\centering
\small
\setlength{\tabcolsep}{4pt}
\small
\begin{tabular}{c l cccccc}
\toprule
Sp.\ & Method & Llama-3.1-8B & Llama-3-70B & Qwen2.5-7B & Qwen2.5-14B & Qwen2.5-32B & Qwen2.5-72B \\
\midrule
\multirow{6}{*}{10\%}
 & RCPU               & 6.93/62.6  & 3.75/74.0 & 6.77/69.1  & 6.20/65.3  & 5.58/71.2  & 4.07/74.5 \\
 & \textbf{COEC}      & \textbf{6.93/62.9}  & 3.72/73.9 & \textbf{6.74/69.2}  & \textbf{6.01/65.9}  & \textbf{5.50/71.3}  & 4.06/74.4 \\
 & Wanda-SP           & 7.32/61.4  & 3.87/73.4 & 6.87/69.3  & 7.04/68.4  & 6.33/68.3  & 4.17/73.3 \\
 & \textbf{$+$COEC}   & \textbf{6.86/63.6}  & \textbf{3.72/73.5} & \textbf{6.78/69.7}  & \textbf{6.06/69.1}  & \textbf{5.43/70.0}  & \textbf{4.09/74.3} \\
 & FLAP               & 6.89/59.9  & 3.50/73.0 & 6.77/67.8  & 6.12/69.0  & 5.24/72.1  & 4.01/74.8 \\
 & \textbf{FLAP\,$+$\,COEC} & 7.41/60.7  & 3.71/74.0 & 6.82/70.0  & 6.07/67.5  & 5.41/71.2  & 4.09/74.5 \\
\midrule
\multirow{6}{*}{20\%}
 & RCPU               & 9.19/52.6  & 5.05/72.6 & 8.33/58.1  & 7.51/62.5  & 6.33/66.5  & 4.85/71.4 \\
 & \textbf{COEC}      & \textbf{8.97/52.8}  & \textbf{4.89/73.4} & \textbf{8.25/59.6}  & \textbf{6.97/62.5}  & 6.25/66.2  & \textbf{4.76/72.3} \\
 & Wanda-SP           & 10.74/52.6 & 5.59/70.8 & 11.49/56.7 & 13.76/57.1 & 12.56/57.8 & 5.46/71.3 \\
 & \textbf{$+$COEC}   & 8.95/52.2  & \textbf{4.93/71.8} & \textbf{8.35/59.9}  & 11.27/56.8 & \textbf{9.20/64.2}  & \textbf{4.73/73.3} \\
 & FLAP               & 8.63/53.4  & 4.72/71.6 & 7.57/59.1  & 7.26/65.7  & 6.05/70.0  & 4.65/72.8 \\
 & \textbf{FLAP\,$+$\,COEC} & 9.96/48.4  & 4.87/73.3 & 8.36/59.2  & 7.18/63.6  & 6.22/68.0  & 4.76/72.9 \\
\midrule
\multirow{6}{*}{30\%}
 & RCPU               & 11.57/47.0 & 5.69/71.3 & 8.98/54.4  & 7.98/59.2  & 7.03/63.8  & 5.23/71.7 \\
 & \textbf{COEC}      & \textbf{10.91/50.2} & \textbf{5.58/71.7} & \textbf{8.91/56.1}  & 7.63/58.5  & \textbf{6.74/64.2}  & \textbf{5.21/72.0} \\
 & Wanda-SP           & 14.35/47.4 & 6.20/68.7 & 13.03/49.1 & 15.15/53.7 & 13.90/57.5 & 6.02/71.1 \\
 & \textbf{$+$COEC}   & \textbf{10.89/48.9} & \textbf{5.61/70.4} & \textbf{9.09/56.0}  & \textbf{13.26/53.7} & \textbf{9.64/61.0}  & \textbf{5.17/72.5} \\
 & FLAP               & 28.74/44.8$^{\dagger}$ & 5.86/68.8 & 8.65/52.6 & 8.82/55.9 & 6.85/65.6 & 5.29/70.5 \\
 & \textbf{FLAP\,$+$\,COEC} & \textbf{11.86/45.6} & \textbf{5.58/71.7} & 9.00/54.5  & \textbf{7.98/59.5}  & 6.71/65.0  & \textbf{5.25/72.5} \\
\bottomrule
\end{tabular}
\caption{\textbf{Main results}, WikiText-2 PPL\,($\downarrow$)\,/\,7-task zero-shot\,\%\,($\uparrow$), across the Llama-3/3.1 and Qwen2.5 families at 10--30\% column sparsity (Sp.). Each of the three selection scores is paired with its own compensation baseline and with COEC on that \emph{same} selection: \textbf{COEC} is the headline method (RCPU's selection, RCPU's rotation$+$scale replaced by COEC) versus the \textbf{RCPU} baseline; \textbf{$+$COEC} adds COEC to score-only \textbf{Wanda-SP}; \textbf{FLAP\,$+$\,COEC} replaces \textbf{FLAP}'s native bias with COEC on FLAP's selection. Calibration-only, no retraining. \textbf{Bold} marks a COEC-family cell that beats its paired baseline on \emph{both} perplexity and zero-shot; against RCPU and Wanda-SP, COEC improves its baseline in essentially every cell; against FLAP's native bias the gain concentrates at high sparsity. $^{\dagger}$FLAP's adaptive allocation over-prunes Llama-3.1-8B at 30\% (a known FLAP pathology); COEC on the same score recovers it to $11.86$.}
\label{tab:main}
\end{table*}

\subsection{Main Results}
\label{sec:main}
Table~\ref{tab:main} pairs each selection score with its own compensation baseline and with COEC on that same selection. Three patterns hold. (1) \emph{Against RCPU---the strongest published compensation---COEC wins on its own selection almost everywhere}: at 30\% Llama-3.1-8B improves $11.57\to10.91$ PPL ($+3.2$ ZS), Qwen2.5-32B $7.03\to6.74$, Qwen2.5-14B $7.98\to7.63$, and the margin holds at 20\% and 10\% and up to 72B. (2) \emph{Adding COEC to a score-only method (Wanda-SP) recovers loss everywhere}, most where the score-only damage is largest---Qwen2.5-32B at 30\% $13.90\to9.64$, Llama-3.1-8B $14.35\to10.89$. (3) \emph{Replacing FLAP's bias with COEC on FLAP's own selection} pays off \emph{at high sparsity}: at 30\% COEC wins on both metrics on Llama-3-70B ($5.86/68.8\to5.58/71.7$), Qwen2.5-14B ($8.82\to7.98$) and Qwen2.5-72B, and rescues FLAP's over-pruning of Llama-3.1-8B ($28.74\to11.86$); at 10--20\%, where column deletion barely dents the network, FLAP's cheap mean-activation bias is already competitive on perplexity and the swap is roughly neutral. The compensation earns its cost where the damage is largest. Together these isolate compensation from selection: the recovery is a property of COEC, not of which columns a given score removes (\S\ref{sec:robust}). Component ablations of the COEC stack (two-sided rotation, scale-in-loop, GCV, tempering, alignment) appear in Sec.~\ref{sec:abl}; the remaining FLAP\,$+$\,COEC cells are running.

\subsection{Ablations}
\label{sec:abl}
Each COEC component is added on top of the RCPU baseline, one at a time, so its marginal effect is visible (Table~\ref{tab:ablation}); we then vary the two continuous knobs ($\alpha$, $\lambda_a$) and the Stiefel stopping tolerance $\delta$, and note the coverage that remains.

\begin{table}[t]
\centering
\small
\setlength{\tabcolsep}{5pt}
\begin{tabular}{lccc}
\toprule
Configuration & PPL\,$\downarrow$ & ZS\,$\uparrow$ & $\Delta$ \\
\midrule
RCPU (baseline)                       & 10.00 & 51.1 & --- \\
$+$ GCV per-mode rescale              &  9.81 & 51.3 & $+0.39$ \\
$+$ converged two-sided rotation     &  9.45 & 51.5 & $+0.94$ \\
$+$ reduced-Stiefel right solve      &  9.86 & 52.7 & $+1.76$ \\
$+$ Gram tempering ($\alpha{=}0.9$)  &  9.65 & 52.7 & $+1.99$ \\
$+$ anchored align ($\lambda_a{=}50$) $=$ \textbf{COEC} & 9.65 & 52.9 & $\mathbf{+2.17}$ \\
\bottomrule
\end{tabular}
\caption{\textbf{Component ablation}: cumulative build-up from RCPU to full COEC, 5-model mean (Mistral-7B, Llama-2-7B, Llama-3.1-8B, Qwen2.5-7B, DeepSeek-R1-Qwen3-8B) at 30\% sparsity. Each row adds one component to the row above. $\Delta$ is cumulative gain vs RCPU ($\Delta$PPL$+\Delta$ZS). PPL is non-monotonic---the Stiefel right solve and Gram tempering spend a little perplexity to buy a large zero-shot gain---but the total gain increases at every step, and no single component carries the result.}
\label{tab:ablation}
\end{table}

\noindent\textbf{Component build-up.} Table~\ref{tab:ablation} reads top-down as the design order. The GCV rescale improves on RCPU's scalar ($+0.39$); the converged two-sided rotation adds the right-frame move a one-sided rotation cannot ($+0.94$); the Stiefel solve and Gram tempering convert reconstruction headroom into zero-shot ($51.5\to52.7$) at a small, deliberate PPL cost; the anchored alignment penalty closes the inter-layer gap of Sec.~\ref{sec:geometry} for the final $+2.17$. The stack is cooperative---removing any component drops the total---not one dominant trick.

\noindent\textbf{Rotation family.} The two-sided win is not automatic: right-only is worse than left-only on base models ($18.10$ vs $11.67$ PPL on Llama-3-8B) but better on the distill, the per-module hybrid tracks each model's better side, and only the \emph{converged} alternation dominates both everywhere. Per-slice (block-diagonal) rotations lose the cross-slice coupling the global one buys, and the dense learned-rotation prototype (Sec.~\ref{sec:related}) collapses outright. Dropping orthogonality for a ridge LS refit splits by model as the spectrum-reshape story predicts ($14.04$ vs $11.66$ worse on L3; $13.76$ vs $15.35$ better on the distill)---the unstable trade COEC avoids by constraining rather than refitting.

\noindent\textbf{Continuous knobs.} \emph{Gram tempering}: per-model ZS optima sit at $\alpha\in\{0.3,\dots,0.9\}$ ($+1.2$--$1.9$ ZS at $\le0.2$ PPL), the collective optimum is $\alpha{=}0.9$, and $\alpha{=}0$ reproduces prune-only. \emph{Alignment} $\lambda_a$: the corrected anchor (Sec.~\ref{sec:align}) gives a single optimum $\lambda_a{=}50$, smooth across $\{0,10,50,100,500,1000\}$ with no model destabilized; as G5 shows it drives the projected mismatch below the pruned point while raw angles do not. \emph{Stopping tolerance} $\delta$: measured against the full-orthogonal right solve at the same budget, the Stiefel reduction wins for every model at every $\delta\in\{5\mathrm{e}{-}4,1\mathrm{e}{-}3,1\mathrm{e}{-}2\}$ (gap $0.19$--$3.67$), and a per-model $\delta$ (one tight, two at $10^{-3}$) adds up to $+0.43$ the headline numbers do not rely on.

\noindent\textbf{Coverage.} All ablations above are run on the $\le$8B \emph{paper-5} set; the component order and continuous optima are stable across those five models, and the mid/large-model repeats (14--72B) are in progress.

\subsection{Robustness to the Pruning Score}
\label{sec:robust}
COEC compensates a \emph{selection}; it does not produce one. To test that its recovery is a property of the compensation rather than of a particular scorer, we hold the full COEC recipe fixed (two-sided $\{Q_\ell,Q_r\}$, GCV rescale, $G^{0.9}$, projector alignment $\lambda_a{=}50$) and swap \emph{only} the score that chooses the kept set $K$, running Wanda-SP and FLAP selections against their score-only baselines across the Llama-2/3/3.1 and Qwen2.5/3 families at $\{0.1,0.2,0.3\}$ sparsity. COEC improves over score-only pruning for \emph{every} scorer, and the gain is largest exactly where the score-only drop is largest---e.g.\ on the Wanda-SP selection, Qwen3-32B at 30\% improves $11.84\to8.94$ PPL ($+6.5$ zero-shot) and Llama-2-7B $8.88\to8.34$---the expected signature of a compensator that recovers loss regardless of which columns a different score removed. For FLAP, whose native method already carries a bias-compensation term, we compare its own compensation against replacing that term with COEC under FLAP's identical selection. The full $14$-model $\times$ $3$-sparsity grid is reported in the supplement.

\section{Conclusion}
COEC treats post-pruning compensation as the WSBM geometry says it is: a return-to-manifold problem, not a reconstruction one---a two-sided, spectrum-preserving transport that keeps the intra-layer and inter-layer spectral alignments for free, with per-mode dosing and exactly one alignment obligation (the inter-layer geometric one) that reconstruction ignores. Every component is closed-form or a short manifold solve; the pipeline is training-free, runs in under two GPU-hours per model, and across the Llama-3/3.1 and Qwen2.5 families beats RCPU---the strongest published compensation---on its own selection while, because it acts on the kept set rather than choosing it, recovering loss just as well for the Wanda-SP and FLAP selections, agnostic to the score. Open problems: an angle-aware (rather than projector) alignment objective for the degenerate-block residual, the SwiGLU gate interaction, and cross-domain calibration at higher sparsity.

\bibliography{aaai2027}

\maketitle

\appendix
\onecolumn

\section{Results at 10\% and 20\% Column Sparsity}
\label{app:sweep}

Tables~\ref{tab:sweep_ppl} and~\ref{tab:sweep_zs} report the 10\% and 20\% settings under the protocol of \mainSetup; the 30\% setting is in \mainTabPPL{} and \mainTabZS. The ordering of the 30\% comparison is preserved with smaller margins, and the margins shrink monotonically as sparsity decreases: under the RCPU selection, the mean perplexity gap over RCPU is $0.06$ at 10\% and $0.20$ at 20\%, against $0.25$ at 30\%; on the Wanda-sp selection, the mean recovery over score-only pruning is $0.4$ at 10\% and $2.0$ at 20\%, against $2.5$ at 30\%. At these ratios the pruning damage is small, and the FLAP bias compensation remains competitive on perplexity, consistent with the discussion in \mainResults.

\begin{table}[h]
\centering
\small
\setlength{\tabcolsep}{4.5pt}
\begin{tabular}{ll cccccc}
\toprule
Selection & Comp. & Llama-3.1-8B & Llama-3-70B & Qwen2.5-7B & Qwen2.5-14B & Qwen2.5-32B & Qwen2.5-72B \\
\midrule
\multicolumn{8}{c}{\cellcolor{gray!12}\textbf{10\% column sparsity}}\\
\midrule
\rowcolor{rcpucol}RCPU     & RCPU & 6.93 & 3.75 & 6.77 & 6.20 & 5.58 & 4.07 \\
\rowcolor{rcpucol}RCPU     & COEC & 6.93 & \textbf{3.72} & \textbf{6.74} & \textbf{6.01} & \textbf{5.50} & \textbf{4.06} \\
\rowcolor{wandacol}Wanda-sp & --   & 7.32 & 3.87 & 6.87 & 7.04 & 6.33 & 4.17 \\
\rowcolor{wandacol}Wanda-sp & COEC & \textbf{6.86} & \textbf{3.72} & \textbf{6.78} & \textbf{6.06} & \textbf{5.43} & \textbf{4.09} \\
\rowcolor{flapcol}FLAP     & bias & \textbf{6.89} & \textbf{3.50} & \textbf{6.77} & 6.12 & \textbf{5.24} & \textbf{4.01} \\
\rowcolor{flapcol}FLAP     & COEC & 7.41 & 3.71 & 6.82 & \textbf{6.07} & 5.41 & 4.09 \\
\midrule
\multicolumn{8}{c}{\cellcolor{gray!12}\textbf{20\% column sparsity}}\\
\midrule
\rowcolor{rcpucol}RCPU     & RCPU & 9.19 & 5.05 & 8.33 & 7.51 & 6.33 & 4.85 \\
\rowcolor{rcpucol}RCPU     & COEC & \textbf{8.97} & \textbf{4.89} & \textbf{8.25} & \textbf{6.97} & \textbf{6.25} & \textbf{4.76} \\
\rowcolor{wandacol}Wanda-sp & --   & 10.74 & 5.59 & 11.49 & 13.76 & 12.56 & 5.46 \\
\rowcolor{wandacol}Wanda-sp & COEC & \textbf{8.95} & \textbf{4.93} & \textbf{8.35} & \textbf{11.27} & \textbf{9.20} & \textbf{4.73} \\
\rowcolor{flapcol}FLAP     & bias & \textbf{8.63} & \textbf{4.72} & \textbf{7.57} & 7.26 & \textbf{6.05} & \textbf{4.65} \\
\rowcolor{flapcol}FLAP     & COEC & 9.96 & 4.87 & 8.36 & \textbf{7.18} & 6.22 & 4.76 \\
\bottomrule
\end{tabular}
\caption{\textbf{WikiText-2 perplexity} $(\downarrow)$ at 10\% and 20\% column sparsity. Protocol and bolding as \mainTabPPL.}
\label{tab:sweep_ppl}
\end{table}

\begin{table}[h]
\centering
\small
\setlength{\tabcolsep}{4.5pt}
\begin{tabular}{ll cccccc}
\toprule
Selection & Comp. & Llama-3.1-8B & Llama-3-70B & Qwen2.5-7B & Qwen2.5-14B & Qwen2.5-32B & Qwen2.5-72B \\
\midrule
\multicolumn{8}{c}{\cellcolor{gray!12}\textbf{10\% column sparsity}}\\
\midrule
\rowcolor{rcpucol}RCPU     & RCPU & 62.6 & \textbf{74.0} & 69.1 & 65.3 & 71.2 & \textbf{74.5} \\
\rowcolor{rcpucol}RCPU     & COEC & \textbf{62.9} & 73.9 & \textbf{69.2} & \textbf{65.9} & \textbf{71.3} & 74.4 \\
\rowcolor{wandacol}Wanda-sp & --   & 61.4 & 73.4 & 69.3 & 68.4 & 68.3 & 73.3 \\
\rowcolor{wandacol}Wanda-sp & COEC & \textbf{63.6} & \textbf{73.5} & \textbf{69.7} & \textbf{69.1} & \textbf{70.0} & \textbf{74.3} \\
\rowcolor{flapcol}FLAP     & bias & 59.9 & 73.0 & 67.8 & \textbf{69.0} & \textbf{72.1} & \textbf{74.8} \\
\rowcolor{flapcol}FLAP     & COEC & \textbf{60.7} & \textbf{74.0} & \textbf{70.0} & 67.5 & 71.2 & 74.5 \\
\midrule
\multicolumn{8}{c}{\cellcolor{gray!12}\textbf{20\% column sparsity}}\\
\midrule
\rowcolor{rcpucol}RCPU     & RCPU & 52.6 & 72.6 & 58.1 & 62.5 & \textbf{66.5} & 71.4 \\
\rowcolor{rcpucol}RCPU     & COEC & \textbf{52.8} & \textbf{73.4} & \textbf{59.6} & 62.5 & 66.2 & \textbf{72.3} \\
\rowcolor{wandacol}Wanda-sp & --   & \textbf{52.6} & 70.8 & 56.7 & \textbf{57.1} & 57.8 & 71.3 \\
\rowcolor{wandacol}Wanda-sp & COEC & 52.2 & \textbf{71.8} & \textbf{59.9} & 56.8 & \textbf{64.2} & \textbf{73.3} \\
\rowcolor{flapcol}FLAP     & bias & \textbf{53.4} & 71.6 & 59.1 & \textbf{65.7} & \textbf{70.0} & 72.8 \\
\rowcolor{flapcol}FLAP     & COEC & 48.4 & \textbf{73.3} & \textbf{59.2} & 63.6 & 68.0 & \textbf{72.9} \\
\bottomrule
\end{tabular}
\caption{\textbf{7-task zero-shot accuracy} \% $(\uparrow)$ at 10\% and 20\% column sparsity. Protocol and bolding as \mainTabPPL.}
\label{tab:sweep_zs}
\end{table}

\section{Implementation Details}
\label{app:impl}

\noindent\textbf{Retractions.} For rectangular Stiefel frames ($b<k$, the MLP case) we use a Newton--Schulz polar retraction \citep{higham}. For square frames ($b=k$, attention output projections) we use an exact Cayley retraction: the Cayley transform of the skew part of the projected gradient, with the matrix inverse replaced by a quadratically convergent Newton--Schulz iteration that runs on the GPU in low precision. On the attention projections, this retraction adds $+0.11$ total gain over the polar retraction at the same tolerance (\mainAbl).

\noindent\textbf{GCV validation.} Minimizing $\mathrm{GCV}(\lambda)$ per layer reproduces the best globally swept ridge on both models where we validated it: effective ridge $0.79$ versus swept $0.75$ on Llama-3, and $0.00$ versus $0$ on the distilled Qwen.

\noindent\textbf{Numerical safeguards.} Three safeguards keep the pipeline stable across models without changing its output on well-conditioned inputs. First, the scale update is applied only when its denominator is positive: on some Qwen attention layers the calibration covariance is indefinite (e.g., $\langle R, S^2 R \Gxx\rangle = -3.9\cdot10^{8}$), and dividing by a clamped denominator produced scales on the order of $10^{37}$. Second, the retraction step is capped so that $\lVert\text{step}\cdot G\rVert\approx1$, since degenerate layers can otherwise produce steps on the order of $10^{36}$. Third, an SVD watchdog converts cuSOLVER hangs on degenerate slices into recoverable failures through a timed fallback chain from \texttt{gesdd} to \texttt{gesvd} to CPU double precision.

\section{Geometric Measurements}
\label{app:geometry}

This appendix backs up the two observations of \mainGeom{} with direct measurements. Pruning rotates the input directions of every layer, and compensation that only minimizes reconstruction error does not rotate them back. COEC keeps the pairing between input and output directions intact by construction. We also explain why alignment is measured with subspace projectors instead of angles.

\noindent\textbf{Setup.} All measurements use DeepSeek-R1-0528-Qwen3-8B at 30\% sparsity (36 layers, Wanda-sp selection). For each pruned layer we compare three sets of input directions, given by the top $r_p{=}16$ right singular vectors: $O$ for the original weight restricted to the kept columns, $P$ for the pruned weight before compensation, and $C$ for the weight after full COEC compensation. The view of a layer as paired read and write directions follows the WSBM framework \citep{wsbm}.

\noindent\textbf{Effect of pruning.} The angle between $O$ and $P$ averages $22.2\degr$ on attention output projections and $22.6\degr$ on MLP down projections. Column deletion alone, before any compensation, already moves every layer noticeably.

\noindent\textbf{Effect of reconstruction-based compensation.} After full compensation, the angle between $O$ and $C$ averages $23.8\degr$ for attention and $23.0\degr$ for MLP, slightly larger than before compensation on 54 of the 72 modules. This is expected. The reconstruction objective never references the original directions, so nothing pulls the layer back toward them. The anchored penalty of \mainAlign{} supplies exactly this missing pull.

\noindent\textbf{Intra-layer pairing.} COEC rotates the weight as a whole, so its input and output directions move together. The measured mismatch between the two sides is $1.1$--$1.2\%$ on average, at most $3.9\%$. No explicit term is needed for this property.

\noindent\textbf{Choice of metric.} A layer can show a $22\degr$ rotation while its projector mismatch is only about $5\cdot10^{-7}$. The reason is that many singular values in these models are nearly equal, and singular vectors within such a group can rotate freely without changing what the layer computes. Angles count this harmless rotation as damage. The projector compares subspaces as a whole and ignores it, so it is the quantity worth controlling.

\noindent\textbf{Effect of the penalty.} With the anchor at the original directions, the penalty is nonzero at the pruned point. At $\lambda_a{=}1$ it barely moves. At the default $\lambda_a{=}50$ it drops by a median factor of $7.2$, and up to $975$, on the attention projections. The raw angles stay at $22$--$24\degr$, because the remaining rotation lies inside the subspace that the projector deliberately ignores. Closing that residual would need an angle-aware objective, which we leave to future work.

\bibliography{aaai2027}

\maketitle

\appendix
\onecolumn

\section{Results at 10\% and 20\% Column Sparsity}
\label{app:sweep}

Tables~\ref{tab:sweep_ppl} and~\ref{tab:sweep_zs} report the 10\% and 20\% settings under the protocol of \mSetupExp; the 30\% setting is in Tables~1 and~2 of \mainpaper. Under the RCPU and Wanda-sp selections the ordering of the 30\% comparison is preserved with smaller margins, and the margins shrink monotonically as sparsity decreases: under the RCPU selection, the mean perplexity gap over RCPU is $0.06$ at 10\% and $0.20$ at 20\%, against $0.25$ at 30\%; on the Wanda-sp selection, the mean recovery over score-only pruning is $0.4$ at 10\% and $2.0$ at 20\%, against $2.5$ at 30\%. Under the FLAP selection the ordering reverses at these ratios: the bias compensation gives lower perplexity on five of the six models at both 10\% and 20\%, while the two remain close on zero-shot accuracy. At mild sparsity the removed columns carry little energy, and a mean correction is enough to absorb it.

\begin{table}[h]
\centering
\small
\setlength{\tabcolsep}{4.5pt}
\begin{tabular}{ll cccccc}
\toprule
Selection & Comp. & Llama-3.1-8B & Llama-3-70B & Qwen2.5-7B & Qwen2.5-14B & Qwen2.5-32B & Qwen2.5-72B \\
\midrule
\multicolumn{8}{c}{\textbf{10\% column sparsity}}\\
\midrule
\multirow{2}{*}{Wanda-sp} & -- & 7.32 & 3.87 & 6.87 & 7.04 & 6.33 & 4.17 \\
 & COEC & \textbf{6.86} & \textbf{3.72} & \textbf{6.78} & \textbf{6.06} & \textbf{5.43} & \textbf{4.09} \\
\cmidrule(lr){1-8}
\multirow{2}{*}{FLAP} & bias & \textbf{6.89} & \textbf{3.50} & \textbf{6.77} & 6.12 & \textbf{5.24} & \textbf{4.01} \\
 & COEC & 7.41 & 3.71 & 6.82 & \textbf{6.07} & 5.41 & 4.09 \\
\cmidrule(lr){1-8}
\multirow{2}{*}{RCPU} & RCPU & 6.93 & 3.75 & 6.77 & 6.20 & 5.58 & 4.07 \\
 & COEC & 6.93 & \textbf{3.72} & \textbf{6.74} & \textbf{6.01} & \textbf{5.50} & \textbf{4.06} \\
\midrule
\multicolumn{8}{c}{\textbf{20\% column sparsity}}\\
\midrule
\multirow{2}{*}{Wanda-sp} & -- & 10.74 & 5.59 & 11.49 & 13.76 & 12.56 & 5.46 \\
 & COEC & \textbf{8.95} & \textbf{4.93} & \textbf{8.35} & \textbf{11.27} & \textbf{9.20} & \textbf{4.73} \\
\cmidrule(lr){1-8}
\multirow{2}{*}{FLAP} & bias & \textbf{8.63} & \textbf{4.72} & \textbf{7.57} & 7.26 & \textbf{6.05} & \textbf{4.65} \\
 & COEC & 9.96 & 4.87 & 8.36 & \textbf{7.18} & 6.22 & 4.76 \\
\cmidrule(lr){1-8}
\multirow{2}{*}{RCPU} & RCPU & 9.19 & 5.05 & 8.33 & 7.51 & 6.33 & 4.85 \\
 & COEC & \textbf{8.97} & \textbf{4.89} & \textbf{8.25} & \textbf{6.97} & \textbf{6.25} & \textbf{4.76} \\
\bottomrule
\end{tabular}
\caption{\textbf{WikiText-2 perplexity} $(\downarrow)$ at 10\% and 20\% column sparsity. Protocol and bolding as \mTabPPL.}
\label{tab:sweep_ppl}
\end{table}

\begin{table}[h]
\centering
\small
\setlength{\tabcolsep}{4.5pt}
\begin{tabular}{ll cccccc}
\toprule
Selection & Comp. & Llama-3.1-8B & Llama-3-70B & Qwen2.5-7B & Qwen2.5-14B & Qwen2.5-32B & Qwen2.5-72B \\
\midrule
\multicolumn{8}{c}{\textbf{10\% column sparsity}}\\
\midrule
\multirow{2}{*}{Wanda-sp} & -- & 61.4 & 73.4 & 69.3 & 68.4 & 68.3 & 73.3 \\
 & COEC & \textbf{63.6} & \textbf{73.5} & \textbf{69.7} & \textbf{69.1} & \textbf{70.0} & \textbf{74.3} \\
\cmidrule(lr){1-8}
\multirow{2}{*}{FLAP} & bias & 59.9 & 73.0 & 67.8 & \textbf{69.0} & \textbf{72.1} & \textbf{74.8} \\
 & COEC & \textbf{60.7} & \textbf{74.0} & \textbf{70.0} & 67.5 & 71.2 & 74.5 \\
\cmidrule(lr){1-8}
\multirow{2}{*}{RCPU} & RCPU & 62.6 & \textbf{74.0} & 69.1 & 65.3 & 71.2 & \textbf{74.5} \\
 & COEC & \textbf{62.9} & 73.9 & \textbf{69.2} & \textbf{65.9} & \textbf{71.3} & 74.4 \\
\midrule
\multicolumn{8}{c}{\textbf{20\% column sparsity}}\\
\midrule
\multirow{2}{*}{Wanda-sp} & -- & \textbf{52.6} & 70.8 & 56.7 & \textbf{57.1} & 57.8 & 71.3 \\
 & COEC & 52.2 & \textbf{71.8} & \textbf{59.9} & 56.8 & \textbf{64.2} & \textbf{73.3} \\
\cmidrule(lr){1-8}
\multirow{2}{*}{FLAP} & bias & \textbf{53.4} & 71.6 & 59.1 & \textbf{65.7} & \textbf{70.0} & 72.8 \\
 & COEC & 48.4 & \textbf{73.3} & \textbf{59.2} & 63.6 & 68.0 & \textbf{72.9} \\
\cmidrule(lr){1-8}
\multirow{2}{*}{RCPU} & RCPU & 52.6 & 72.6 & 58.1 & 62.5 & \textbf{66.5} & 71.4 \\
 & COEC & \textbf{52.8} & \textbf{73.4} & \textbf{59.6} & 62.5 & 66.2 & \textbf{72.3} \\
\bottomrule
\end{tabular}
\caption{\textbf{7-task zero-shot accuracy} \% $(\uparrow)$ at 10\% and 20\% column sparsity. Protocol and bolding as \mTabPPL.}
\label{tab:sweep_zs}
\end{table}

\section{Implementation Details}
\label{app:impl}

\noindent\textbf{Retractions.} For rectangular Stiefel frames ($b<k$, the MLP case) we use a Newton--Schulz polar retraction \citep{higham}. For square frames ($b=k$, attention output projections) we use an exact Cayley retraction \citep{wenyin}: the Cayley transform of the skew part of the projected gradient, with the matrix inverse replaced by a quadratically convergent Newton--Schulz iteration that runs on the GPU in low precision. On the attention projections, this retraction adds $+0.11$ mean combined gain over the polar retraction at the same tolerance (Table~\ref{tab:retraction} in App.~\ref{app:rotabl}).

\noindent\textbf{GCV validation.} The ridge $\lambda$ is selected per module (each MLP down projection and attention output projection solved independently) by minimizing $\mathrm{GCV}(\lambda)$ over a 120-point logarithmic grid spanning $[10^{-4},10^{5}]$ scaled by the median of the mode energies $e_i$, so the grid is data-scaled and dimensionless; each grid evaluation is closed form, so no held-out data or retraining is needed. The resulting per-module effective ridge reproduces the best globally swept ridge on both models where we validated it: effective ridge $0.79$ versus swept $0.75$ on Llama-3, and $0.00$ versus $0$ on the distilled Qwen.

\noindent\textbf{Per-mode dose band.} The per-mode rescale is re-applied inside every alternation round, where the rotation frames it reads are still changing. On near-degenerate spectra this can occasionally drive a single mode's least-squares value far from the uniform one (isolated tail values as large as ${\sim}10^{4}\times$ the uniform value were measured), which under the iterated frames can inflate the largest singular value and destabilize a later SVD. To prevent this, every per-mode value $s_{g,i}$ is clamped to within a factor $\kappa=10$ of the uniform value $s^{*}\sigma_i$ (the same quantity the GCV ridge shrinks toward), in all rescale branches including GCV. The measured healthy per-mode deviation from the uniform value is about $2\%$, so the band is inactive on well-conditioned modes and only removes the pathological tails.

\noindent\textbf{Numerical safeguards.} Three further safeguards keep the pipeline stable across models without changing its output on well-conditioned inputs. First, the scale update is applied only when its denominator is positive: on some Qwen attention layers the calibration covariance is indefinite (e.g., $\langle R, S^2 R \Gxx\rangle = -3.9\cdot10^{8}$), and dividing by a clamped denominator produced scales on the order of $10^{37}$. Second, the retraction step is capped so that $\lVert\text{step}\cdot G\rVert\approx1$, since degenerate layers can otherwise produce steps on the order of $10^{36}$. Third, an SVD watchdog converts cuSOLVER hangs on degenerate slices into recoverable failures through a timed fallback chain from \texttt{gesdd} to \texttt{gesvd} to CPU double precision.

\noindent\textbf{Hyperparameter settings.} All reported results use one configuration: tempering exponent $\alpha=0.9$ (App.~\ref{app:hyper}); alignment weight $\lambda_a=50$ on the attention output projection and $0$ on the MLP (swept over $\{0,10,50,100,500,1000\}$); projector rank $r_p=16$; stopping tolerance $\delta=7\cdot10^{-4}$, shared by every model and every reported cell (sensitivity in App.~\ref{app:hyper}); at most $50$ alternation rounds with early stop at $10^{-4}$ relative change; per-mode dose band $\kappa=10$. Calibration uses $128$ WikiText-2 training sequences of length $2048$; perplexity is evaluated at sequence length $4096$.

\noindent\textbf{Randomness.} The compensation itself is deterministic given the calibration Grams: the left rotation and the rescale are closed form, and the right solve starts from the fixed initialization $Q_0=I$. The only stochastic component is the sampling of the $128$ calibration sequences, drawn with a fixed seed ($0$, also set for NumPy and PyTorch); all reported numbers use this single seed, and Table~\ref{tab:calibsize} reports the spread over three calibration seeds.

\noindent\textbf{Compute environment.} Headline compensations and all table evaluations ran on nodes with NVIDIA B300 GPUs (SXM6, $\sim$275\,GB; Intel Xeon 6737P, 3\,TB RAM); NVIDIA L40S nodes (46\,GB; AMD EPYC 7513, 512\,GB RAM) were used for diagnostics only. Software: Ubuntu 22.04, Python 3.9, PyTorch 2.8.0 (CUDA 12.9), \texttt{transformers} 4.57.6, \texttt{lm-eval} 0.4.9.1. A same-configuration cross-check of Qwen2.5-7B at 30\% sparsity on L40S reproduced the B300 perplexity to within $0.001$ but shifted 7-task zero-shot accuracy by $0.55$ points, a GPU-class-level deviation on the zero-shot path; every table cell in the paper therefore comes from the B300 nodes, and the L40S diagnostics are restricted to reconstruction-side quantities that reproduce across GPU classes.

\section{Geometric Measurements}
\label{app:geometry}

This appendix reports the measurements summarized in \mGeom: that reconstruction-based compensation leaves the alignment between adjacent layers unrepaired, which motivates the penalty of \mAlign, and that the spectrum-preserving family maintains the alignment within each layer without any explicit term. It also explains why the inter-layer alignment is measured through a projector rather than through principal angles.

We adopt the terminology of the WSBM framework \citep{wsbm}, which characterizes stable networks by two families of alignment properties. Within a layer, the singular spectrum follows a power law and the left and right singular frames are transported jointly. Between layers, the spectral decay exponents match, and the relation between what a layer writes and what its successor reads is stable; the read--write alignment $M=V_o^{\top}U_v$ measured below is this relational quantity. The spectrum-preserving constraint of \mSetup{} maintains the spectral properties by construction, so the open question concerns the frames.

\noindent\textbf{Measurement protocol.} All measurements are taken on DeepSeek-R1-0528-Qwen3-8B at 30\% sparsity (36 layers, Wanda-sp selection). The observed geometry reflects column deletion and compensation rather than the particular selection score, consistent with the plug-in results of \mMain. Frames refer to top-$r_p$ singular frames, angles are principal angles, and alignment values are computed through the gauge-invariant projector of \mAlign. We write $O$ for the original read frame restricted to the kept coordinates, $P$ for the right singular frame of the pruned weight, and $C$ for the frame after full COEC compensation.

\noindent\textbf{Effect of pruning on the read frame.} Across the 36 layers, the principal angle between $O$ and $P$ averages $22.2\degr$ for attention output projections (measured at $r_p=16$; per-module values span $8$--$48\degr$, with worst directions up to $90\degr$) and $22.6\degr$ for MLP down projections (measured at $r=64$; per-module values span $2$--$32\degr$). The $20$--$30\degr$ figure quoted in \mAlign{} refers to these module means. Column deletion alone, before any compensation, substantially rotates the read frame of each layer.

\noindent\textbf{Effect of reconstruction-based compensation.} After full compensation, the mean angle between $O$ and $C$ is $23.8\degr$ for attention and $23.0\degr$ for MLP, larger than the corresponding $O$--$P$ angle on 54 of the 72 module-layers. The compensation does not move the read frame back toward the original. This is consistent with the objective being optimized: the reconstruction error makes no reference to the original frame, so there is no force pulling the frame back.

\noindent\textbf{Intra-layer alignment.} The update $\Ql\WK\Qr$ maps $U_K \to \Ql U_K$ and $V_K \to \Qr^{\top}V_K$, so the two frames are transported jointly and their pairing is preserved; only the per-mode rescale could break it. We verify that it does not: across the 36 layers, the frame-change discrepancy $|Q_u-Q_v|/|Q_u|$, with $Q_u=U_P^{\top}U_C$ and $Q_v=V_P^{\top}V_C$, averages $1.1$--$1.2\%$ at $r_p=16$, with a maximum of $3.9\%$. The spectrum-preserving family therefore satisfies the intra-layer alignment by construction, and only the inter-layer alignment above requires an explicit term.

\noindent\textbf{Choice of the alignment metric.} A layer can show a mean $O$--$P$ rotation of $22\degr$ (maximum $87\degr$) while its projector penalty is only about $5\cdot10^{-7}$ (median; $10^{-4}$ at worst). The reason is that rotations inside near-degenerate spectral blocks span many degrees while leaving the top-$r_p$ projector nearly unchanged, since the projector is invariant to rotation within the retained subspace (Lemma~1 in App.~\ref{app:theory}). An $S^2$-weighted variant of the penalty gives values of the same order, so this behavior reflects the structure of the alignment target rather than a particular weighting. Principal angles therefore overstate the damage of pruning, and the projector penalty is the appropriate quantity to control. One caveat is that the rank cut itself is near-degenerate in these models ($\sigma_{r_p}/\sigma_{r_p+1}\approx1.00$ at $r_p=16$ and $64$), so $r_p$ should be placed at a spectral gap when one exists.

\noindent\textbf{Effect of the alignment penalty.} With the anchor at the restricted original frame, the penalty at the pruned point is nonzero ($10^{-8}$--$10^{-4}$ depending on the layer). At $\lambda_a=1$ the penalty barely moves; at the default $\lambda_a=50$ it falls by a median factor of $7.2$ (up to $975$) on the attention projections. The angles between $O$ and $C$ do not decrease and remain at $22$--$24\degr$: the penalty acts on the projected quantity, and the remaining angle lies in the within-subspace component that the projector is designed to ignore. Reducing the projected mismatch is what translates into the end-metric gain of \mAbl; closing the within-subspace residual would require an angle-aware objective, which we leave to future work.

\section{Theoretical Analysis}
\label{app:theory}

This appendix formalizes the properties invoked in \mainpaper: the alternation descends monotonically under a step-size assumption on the right block (\mRight), the alignment penalty is well posed despite SVD gauge ambiguity (\mAlign), and the spectrum-preserving family generalizes where least squares overfits (\mSetup). Throughout, a pruned sub-layer keeps columns $K$ ($|K|=k$) of $\Wz\in\mathbb{R}^{d_{\mathrm{out}}\times d_{\mathrm{in}}}$, giving $\WK=\Wz[:,K]\in\mathbb{R}^{d_{\mathrm{out}}\times k}$ with SVD $\WK=U_K\Sigma_K V_K^{\top}$; $X_K\in\mathbb{R}^{k\times N}$ are the kept calibration activations and $Y=\Wz X\in\mathbb{R}^{d_{\mathrm{out}}\times N}$ the original outputs. COEC searches over compensated weights
\begin{equation}
\label{eq:family}
\mathcal{C} \;=\; \big\{\, W' = \Ql\,\big(U_K\,\mathrm{diag}(s)\,\Sigma_K\,V_K^{\top}\big)\,\Qr \;:\; \Ql\in O(d_{\mathrm{out}}),\ \Qr\in O(k),\ s\in\mathbb{R}^{k}_{>0} \,\big\},
\end{equation}
i.e., an outer rotation of the output frame, an inner rotation of the read frame, and a per-mode scale $s$. Writing $\WK^{(s)}=U_K\,\mathrm{diag}(s)\,\Sigma_K\,V_K^{\top}$, the reconstruction objective is $\mathcal{L}(W')=\lVert Y - W' X_K\rVert_F^2$.

\subsection{Monotone descent of the two-sided alternation}

\begin{quote}
\noindent\textbf{Proposition 1 (Monotone descent under a step-size assumption).}
\emph{Write $\mathcal{L}(\Ql,s,\Qr)=\lVert Y-\Ql\,\WK^{(s)}\,\Qr X_K\rVert_F^2$. Consider the alternation that cyclically updates $\Ql$ by orthogonal Procrustes (the exact conditional minimizer), $s$ by the GCV-regularized per-mode rescale (a non-increasing update for every regularization strength $\lambda\ge0$), and $\Qr$ by a retracted gradient step. Assume}
\begin{itemize}
\item[\textbf{(A1)}] \emph{each $\Qr$-step uses a step size $\eta\le 1/\hat L_{\mathrm{pb}}$, where $\hat L_{\mathrm{pb}}$ is a Lipschitz-type bound on the gradient of the pullback along the retraction.}
\end{itemize}
\emph{Then the iterates satisfy $\mathcal{L}(\theta^{(t+1)})\le\mathcal{L}(\theta^{(t)})$, and since $\mathcal{L}\ge 0$ the sequence $\{\mathcal{L}(\theta^{(t)})\}$ converges. Every limit point at which (A1) holds is block-coordinatewise stationary.}
\end{quote}

\noindent\emph{Proof.} Fix $s,\Qr$ and let $B=\WK^{(s)}\Qr X_K$. The $\Ql$-update solves $\min_{\Ql\in O(d_{\mathrm{out}})}\lVert Y-\Ql B\rVert_F^2$, the orthogonal Procrustes problem, whose global minimizer is $\Ql^{*}=\widetilde U \widetilde V^{\top}$ from the SVD $YB^{\top}=\widetilde U\widetilde\Sigma \widetilde V^{\top}$ \citep{schonemann}; hence the update attains the block minimum and cannot increase $\mathcal{L}$.

Fixing $\Ql,\Qr$, we show the scale update is non-increasing (it is the exact conditional minimizer only at $\lambda=0$; for $\lambda>0$ the GCV-regularized update deliberately trades in-sample error for generalization, and we claim only monotonicity). In the fixed frame of the current weight, $\widehat W=U_gS_gV_g^{\top}$ (the rotations absorbed into $U_g,V_g$), replacing each $S_{g,i}$ by a per-mode scale $d_i$ makes the objective separable: $\mathcal{L}(d)=\sum_i\big(e_i\, d_i^2-2\,\rho_i\, d_i\big)+\mathrm{const}$ with $e_i=(V_g^{\top}\Gxx V_g)_{ii}>0$ and $\rho_i=[U_g^{\top}\Gyx V_g]_{ii}$, exactly the quantities of the per-mode GCV rescale (\mGcv), minimized modewise at $d_i^{*}=\rho_i/e_i$. The GCV rescale returns $d(\lambda)=\arg\min_d \mathcal{L}(d)+\lambda\lVert d-m\rVert^2$ with prior mean $m_i=s^{*}S_{g,i}$, the optimal global scalar for the incoming weight. Since the penalty vanishes at $d=m$, $\mathcal{L}(d(\lambda))\le\mathcal{L}(d(\lambda))+\lambda\lVert d(\lambda)-m\rVert^2\le\mathcal{L}(m)$; and since $m$ is the best uniform rescaling, $\mathcal{L}(m)\le\mathcal{L}(S_g)$, the incoming value. Hence for every $\lambda\ge0$ the rescale is non-increasing.

Fixing $\Ql,s$, the $\Qr$-update is a retracted gradient step on $\mathcal{L}$ restricted to the reduced Stiefel frame (\mStiefel). The Euclidean gradient in $\Qr$ has Lipschitz constant $\hat L=2\,\sigma_1(\WK^{(s)})^2\,\lambda_{\max}(X_KX_K^{\top})$; under assumption (A1), the step satisfies the sufficient-decrease inequality and is non-increasing without a line search \citep[Ch.~4]{boumal}.

Cyclically, $\mathcal{L}(\theta^{(t+1)})\le\mathcal{L}(\theta^{(t)})$. A monotone sequence bounded below by $0$ converges. At any limit point the exactly-minimized blocks ($\Ql$, and $s$ at $\lambda=0$) are at their conditional optima, and under (A1) the $\Qr$-block satisfies the standard stationarity condition for retracted gradient methods \citep{boumal}; the limit point is in this sense block-coordinatewise stationary. \hfill$\square$

\begin{figure}[h]
\centering
\includegraphics[width=0.62\textwidth]{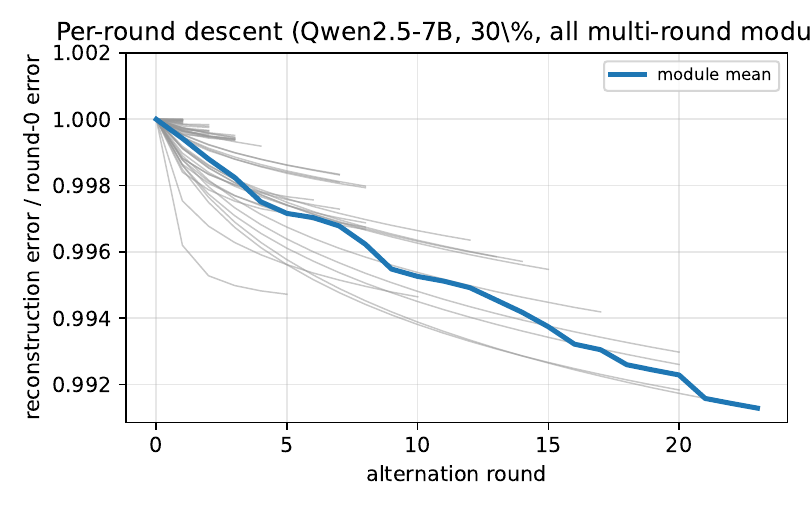}
\caption{Empirical check of Proposition~1's descent behavior. Per-round reconstruction error $\mathcal{L}(\theta^{(t)})/\mathcal{L}(\theta^{(0)})$ of the two-sided alternation, one gray curve per multi-round module (all $56$ modules of Qwen2.5-7B at 30\% sparsity), with the module mean in blue. Every curve descends to a plateau within the cap, monotone up to floating-point noise at the plateau ($55$ of $56$ strictly monotone). The small magnitude reflects the near-identity optimum at mild sparsity.}
\label{fig:conv}
\end{figure}

\smallskip
\noindent\textbf{Implementation.} The solver realizes the $\Qr$-step as a fixed Lipschitz-scaled step of the above form (rescaled to $\eta/s^2$ in the scale-in-loop to match the $s$-scaled objective), with no backtracking line search and no explicit sufficient-decrease test, so assumption (A1) is not enforced programmatically. Its only safeguard, freezing a block at its last finite frame on a non-finite step, is a numerical backstop; it plays no role in the monotonicity. That the realized step stays in the descent regime is therefore an empirical matter, and it is confirmed on a production run (Fig.~\ref{fig:conv}: $55$ of $56$ modules strictly monotone).

\smallskip
\noindent\textbf{Remark (the attention-output composite objective).} On the attention output projections the right solve minimizes the composite $\mathcal{F}=\mathcal{L}+\lambda_a\mathcal{A}$, where $\mathcal{A}$ is the alignment penalty of \mAlign, not $\mathcal{L}$ alone; on the MLP down projection there is no writer frame and $\mathcal{F}=\mathcal{L}$. The block argument carries over to $\mathcal{F}$: the $\Qr$-step is a retracted gradient step on $\mathcal{F}$, with the Lipschitz bound of $\nabla\mathcal{F}$ in place of $\nabla\mathcal{L}$; the $\Ql$-step leaves $\mathcal{A}$ unchanged, since $\mathcal{A}$ depends only on the right frame; and the rescale leaves $\mathcal{A}$ unchanged whenever it preserves the top-$r_p$ index set. Reordering within a near-degenerate block is a gauge move covered by Lemma~1, but a swap across the rank cut, between mode $r_p$ and mode $r_p{+}1$, replaces a kept singular direction and changes the projector by up to $\lVert \Pi_{r_p}-\Pi_{r_p}'\rVert_F^2=2$; this is not covered by Lemma~1, and the rank cut is exactly the near-degenerate regime where such swaps can occur ($\sigma_{r_p}/\sigma_{r_p+1}\approx1$, App.~\ref{app:geometry}). We therefore state the monotonicity of $\mathcal{F}$ on the attention output projection conditionally: it follows from the cyclic argument whenever the rescale preserves the top-$r_p$ index set, and where a cross-cut swap occurs we treat it as an empirical observation. Consistent with this, Figure~\ref{fig:conv} reports the reconstruction term $\mathcal{L}$ rather than $\mathcal{F}$: on the attention output projection a decrease in $\mathcal{L}$ is not implied by a decrease in $\mathcal{F}$, so the monotonicity of $\mathcal{L}$ observed there is empirical rather than a consequence of the proof.

\smallskip
\noindent This is the formal content of the convergence statements of \mRight; the cap of $50$ rounds and the $10^{-4}$ early stop are the numerical realization. Figure~\ref{fig:conv} confirms the descent on a production run: across all $56$ modules ($28$ layers $\times$ \{down, attention-output\}) that took more than one round on Qwen2.5-7B at 30\% sparsity, the per-round reconstruction error descends to a plateau well inside the round cap. Fifty-five of the $56$ modules are strictly monotone; the single exception rises by about $10^{-6}$ after it has already converged, which is floating-point noise at the plateau rather than a genuine ascent. The near-identity magnitude at mild sparsity is expected, since the two-sided family contains the one-sided optimum and the gain lies in generalization rather than in-sample error (\mAbl).

\subsection{Well-posedness of the projector alignment penalty}

The alignment penalty (\mAlign) compares read frames through the top-$r_p$ orthogonal projector $\Pi_{r_p}(V)=V_{r_p} V_{r_p}^{\top}$, where $V_{r_p}\in\mathbb{R}^{k\times r_p}$ collects the top-$r_p$ right singular vectors. Singular vectors are defined only up to a sign, and up to an arbitrary rotation inside any degenerate block; the penalty must not depend on this gauge.

\begin{quote}
\noindent\textbf{Lemma 1 (Gauge invariance).}
\emph{For any $O\in O(r_p)$, $\Pi_{r_p}(V_{r_p} O)=\Pi_{r_p}(V_{r_p})$. In particular the penalty is invariant to (i) per-column sign flips $V_{r_p}\mapsto V_{r_p}\,\mathrm{diag}(\pm1)$ and (ii) rotations within a degenerate singular subspace. Hence the alignment objective $\mathcal{A}(R)$, which depends on the frames only through their top-$r_p$ projectors, is well defined.}
\end{quote}

\noindent\emph{Proof.} $\Pi_{r_p}(V_{r_p} O)=(V_{r_p} O)(V_{r_p} O)^{\top}=V_{r_p}\,O O^{\top}V_{r_p}^{\top}=V_{r_p} V_{r_p}^{\top}=\Pi_{r_p}(V_{r_p})$, using $OO^{\top}=I$. Sign flips are the case $O=\mathrm{diag}(\pm1)\in O(r_p)$; a rotation of a degenerate block is a block-diagonal $O\in O(r_p)$ acting on the repeated directions. As $\mathcal{A}$ is a function of $\Pi_{r_p}$ alone, it inherits the invariance. \hfill$\square$

\smallskip
\noindent This is also why the projector form is preferred over alternatives: any penalty that reads individual singular vectors, such as raw angles or an $S^2$-weighted inner product, is gauge-dependent and ill-posed on degenerate blocks, which we find empirically dominate the rank cut ($\sigma_{r_p}/\sigma_{r_p+1}\approx1$, App.~\ref{app:geometry}).

\subsection{Spectrum preservation and generalization}

RCPU motivates the robustness of its one-sided update by a degrees-of-freedom count: an orthogonal update has far fewer free parameters than an unconstrained least-squares refit, so it overfits a small calibration set less \citep{rcpu}. The two-sided family $\mathcal{C}$ has more rotational freedom, so the parameter-count argument no longer separates it from least squares. We give instead a structural statement that does.

\begin{quote}
\noindent\textbf{Proposition 2 (Spectrum preservation).}
\emph{Every $W'\in\mathcal{C}$ has singular values $\sigma_i(W')=s_i\,\sigma_i(\WK)$; the rotations $\Ql,\Qr$ do not affect the spectrum. In particular a uniform scale $s\equiv c$ gives $\sigma(W')=c\,\sigma(\WK)$, so no $W'\in\mathcal{C}$ can create a direction of gain absent from $\WK$, and $\lVert W'-\WK\rVert_2$ is bounded by the rotation angles and the scale. By contrast the ridge least-squares refit $W'_{\mathrm{LS}}=(YX_K^{\top})(X_KX_K^{\top}+\lambda I)^{-1}$ has unconstrained singular values.}
\end{quote}

\noindent\emph{Proof.} Any $W'\in\mathcal{C}$ is $W'=\Ql U_K\,\mathrm{diag}(s)\Sigma_K\,V_K^{\top}\Qr=(\Ql U_K)\,\big(\mathrm{diag}(s)\Sigma_K\big)\,(\Qr^{\top}V_K)^{\top}$. Because $\Ql U_K$ and $\Qr^{\top}V_K$ are orthogonal, this is an SVD of $W'$, so its singular values are $s_i\sigma_i(\WK)$. For $s\equiv c$, writing $\Ql(c\Sigma_K)\Qr-\Sigma_K=c(\Ql-I)\Sigma_K\Qr+c\Sigma_K(\Qr-I)+(c-1)\Sigma_K$ and applying the triangle inequality with $\lVert\Sigma_K\rVert_2=\sigma_1(\WK)$ gives $\lVert W'-\WK\rVert_2\le \sigma_1(\WK)\big(c\lVert I-\Ql\rVert_2+c\lVert I-\Qr\rVert_2+|c-1|\big)$; the rotation terms carry the factor $c$, which is not negligible since the post-pruning optimal scale $c=s^{*}$ is typically greater than $1$. The least-squares solution places no constraint on $\sigma(W'_{\mathrm{LS}})$. \hfill$\square$

\begin{figure}[h]
\centering
\includegraphics[width=0.66\textwidth]{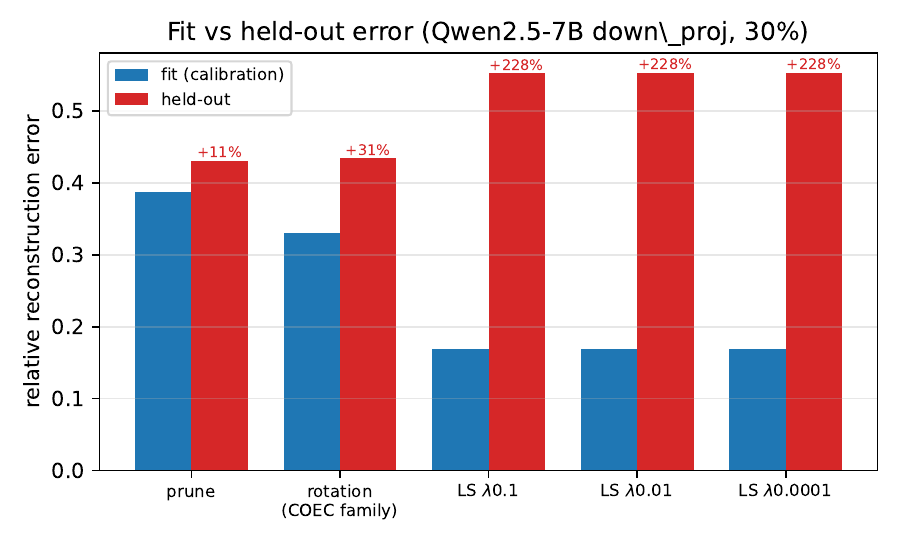}
\caption{Empirical validation of Proposition~2. Fit and held-out reconstruction errors for three compensations on one layer (Qwen2.5-7B down projection, 30\% sparsity). The ridge least-squares refit reaches the lowest fit error ($0.168$) but the worst held-out error ($0.552$, a $+228\%$ gap on the unrounded values, insensitive to $\lambda$). The spectrum-preserving rotation, whose singular values are fixed by Proposition~2, keeps the two errors close.}
\label{fig:gen}
\end{figure}

\smallskip
\noindent\textbf{Why this controls overfitting.} The distinction shows up directly in held-out reconstruction. Figure~\ref{fig:gen} splits one layer's calibration set into a fit half and a held-out half and fits three compensations on the fit half. The unconstrained ridge least-squares refit reaches the lowest fit error ($0.168$ versus $0.331$) but the worst held-out error ($0.552$, a $+228\%$ fit-to-held-out gap on the unrounded values, larger than prune-only), and tuning the ridge $\lambda$ over four orders of magnitude does not close the gap. This is the small-sample overfitting that motivates avoiding unconstrained fitting. The spectrum-preserving rotation keeps fit and held-out errors close ($0.331$ and $0.434$): by Proposition~2 its singular values are pinned to $s\odot\sigma(\WK)$, so it cannot introduce gain to fit calibration noise, and $s$ itself is regularized toward the uniform scale by generalized cross-validation \citep{gcv}; GCV selects the rescaling by predictive risk rather than in-sample error. The two-sided rotations add expressive power only in the frames, while the gains stay tied to the pretrained weight. This is the mechanism behind the least-squares-versus-rotation split measured in App.~\ref{app:rotabl}.

\smallskip
\noindent\textbf{Optimality of the sub-steps.} The left rotation is a global optimum of its Procrustes subproblem (Proposition~1; \citealp{schonemann}); the per-mode scale minimizes the GCV criterion in closed form per grid point over the logarithmic grid of \mGcv, requiring no held-out data; and the scale-in-loop recomputes the optimal $s$ against the current $\Qr$ each round, so the right rotation converges jointly with its scale (\mRight). The remaining hyperparameters, the spectral exponent $\alpha$ and the stopping tolerance $\delta$, are selected on calibration data (App.~\ref{app:hyper}).

\smallskip
\noindent Table~\ref{tab:theory} summarizes how each guarantee is established.

\begin{table}[h]
\centering
\small
\setlength{\tabcolsep}{4pt}
\begin{tabular}{l l l}
\toprule
Result & Guarantee & How it is established \\
\midrule
Prop.~1 (descent) & $\mathcal{L}(\theta^{(t)})$ monotone under (A1) & block-coord.\ argument; Fig.~\ref{fig:conv} (55/56 monotone) \\
Lemma~1 (gauge) & penalty invariant to gauge & exact algebra ($\Pi_{r_p}=V_{r_p}V_{r_p}^{\top}$) \\
Prop.~2 (spectrum) & $\sigma_i(W')=s_i\,\sigma_i(\WK)$; no invented gain & exact by construction; gen.\ gap (Fig.~\ref{fig:gen}) \\
Sub-steps & $\Ql$ optimal; rescale non-increasing & Procrustes \citep{schonemann}, GCV \citep{gcv} \\
\bottomrule
\end{tabular}
\caption{Summary of theoretical guarantees. Each is established by proof, by exact algebra, or by an empirical check (this appendix and App.~\ref{app:hyper}).}
\label{tab:theory}
\end{table}

\section{Hyper-parameter Selection}
\label{app:hyper}

This appendix reports the sweeps behind the shared configuration used for every result in \mainpaper{} ($\alpha{=}0.9$, $\lambda_a{=}50$, $\delta{=}7\mathrm{e}{-}4$, 128 calibration sequences): the Gram tempering exponent $\alpha$, the anchored alignment weight $\lambda_a$, the stopping tolerance $\delta$, the round cap of the alternation, and the calibration budget.

\noindent\textbf{Tempering exponent.} Table~\ref{tab:sweep_alpha_full} sweeps $\alpha$ on the full pipeline (all components active except the alignment penalty, which is insensitive to $\alpha$): $\alpha=0.9$ has the best combined gain, and $\alpha=0.8$ is close behind with a slightly better perplexity mean, so the collective optimum is flat rather than knife-edged. Table~\ref{tab:sweep_alpha_wide} maps the wide-range behavior on the earlier left-rotation$+$GCV configuration, where the sweep originally identified the useful range: per-model zero-shot optima are spread over $[0.3,1.0]$ and perplexity is not monotone in $\alpha$ for every model (Llama-3.1-8B in particular), which is why the collective value is chosen on combined gain rather than per-metric.

\begin{table}[t]
\centering
\small
\setlength{\tabcolsep}{4pt}
\begin{tabular}{l ccccc}
\toprule
$\alpha$ & Mistral-7B & Llama-2-7B & Llama-3.1-8B & Qwen2.5-7B & DS-Qwen3-8B \\
\midrule
1.0 (no tempering) & 6.97/51.4 & 7.24/53.6 & 10.88/49.6 & 9.51/56.0 & 14.25/53.1 \\
0.9 (default) & 6.95/51.2 & 7.23/54.3 & 10.92/49.8 & 8.91/55.8 & 14.26/52.6 \\
0.8 & 6.96/51.2 & 7.17/54.0 & 10.93/49.5 & 8.90/55.6 & 14.26/52.9 \\
0.7 & 6.99/51.5 & 7.14/54.1 & 11.67/47.5 & 8.89/56.1 & 14.27/53.2 \\
\bottomrule
\end{tabular}
\caption{\textbf{Gram tempering exponent $\alpha$ on the full pipeline} (all components except the alignment penalty), 30\% sparsity, perplexity\,/\,zero-shot. Mean combined gains over RCPU are $1.87$, $1.99$, $1.90$, and $1.58$ for $\alpha=1.0$, $0.9$, $0.8$, $0.7$; the default $\alpha=0.9$ is the collective optimum, with $0.8$ nearly equivalent.}
\label{tab:sweep_alpha_full}
\end{table}

\begin{table}[t]
\centering
\small
\setlength{\tabcolsep}{4pt}
\begin{tabular}{l ccccc}
\toprule
$\alpha$ & Mistral-7B & Llama-2-7B & Llama-3.1-8B & Qwen2.5-7B & DS-Qwen3-8B \\
\midrule
1.0 (no tempering) & 7.19/50.4 & 7.10/52.8 & 11.48/47.3 & 8.95/54.1 & 14.90/51.4 \\
0.7 & 7.25/50.3 & 7.11/53.6 & 10.99/47.7 & 9.04/54.4 & 15.16/50.7 \\
0.6 & 7.25/50.0 & 7.16/54.0 & 10.88/49.0 & 9.13/54.2 & 15.32/50.3 \\
0.5 & 7.27/50.1 & 7.31/54.1 & 11.80/46.9 & 9.18/54.8 & 15.61/50.2 \\
0.4 & 7.38/49.9 & 7.33/53.5 & -- & 9.32/55.2 & 16.05/50.2 \\
0.3 & 7.56/50.0 & 7.47/52.6 & 11.45/49.2 & 9.49/55.4 & 16.71/49.7 \\
\bottomrule
\end{tabular}
\caption{\textbf{Wide-range $\alpha$ sweep on the left-rotation$+$GCV configuration} (before the full pipeline existed), 30\% sparsity, perplexity\,/\,zero-shot. Zero-shot optima are per-model, and perplexity is non-monotone in $\alpha$ on Llama-3.1-8B; a dash marks a run discarded as anomalous. This sweep bounds the useful range; the selection itself is made on the full pipeline (Table~\ref{tab:sweep_alpha_full}).}
\label{tab:sweep_alpha_wide}
\end{table}

\begin{table}[t]
\centering
\small
\setlength{\tabcolsep}{4pt}
\begin{tabular}{l ccccc}
\toprule
$\lambda_a$ & Mistral-7B & Llama-2-7B & Llama-3.1-8B & Qwen2.5-7B & DS-Qwen3-8B \\
\midrule
0 (off) & 6.95/51.3 & 7.22/54.0 & 10.92/49.7 & 8.90/55.9 & 14.25/52.5 \\
10 & 6.96/51.2 & 7.18/54.2 & 10.92/49.6 & 8.90/56.1 & 14.27/52.5 \\
50 & 6.96/51.4 & 7.26/53.6 & 10.91/49.9 & 8.90/55.9 & 14.27/52.6 \\
100 & 6.96/51.3 & 7.23/54.0 & 10.93/50.0 & 8.90/55.8 & 14.25/52.3 \\
500 & 6.96/51.2 & 7.23/53.9 & 10.79/49.5 & 8.90/56.0 & 14.24/52.7 \\
1000 & 6.96/51.3 & 7.27/53.8 & 10.80/49.6 & 8.90/55.8 & 14.23/52.8 \\
\bottomrule
\end{tabular}
\caption{\textbf{Anchored alignment weight $\lambda_a$} at 30\% sparsity (perplexity\,/\,zero-shot), applied on the attention output projection. The metrics vary smoothly and no model destabilises across four orders of magnitude; $\lambda_a{=}50$ is the collective optimum and is used throughout.}
\label{tab:sweep_lambda}
\end{table}

\begin{table}[t]
\centering
\small
\setlength{\tabcolsep}{4pt}
\begin{tabular}{l ccc}
\toprule
Model & $\delta{=}10^{-2}$ & $\delta{=}10^{-3}$ & $\delta{=}7{\cdot}10^{-4}$ (default) \\
\midrule
Mistral-7B & 7.20/50.3 & 6.98/50.9 & 6.96/50.7 \\
Llama-2-7B & 7.12/54.1 & 7.17/53.3 & 7.22/53.2 \\
Llama-3.1-8B & 11.06/48.1 & 10.88/49.1 & 10.91/49.3 \\
Qwen2.5-7B & 9.03/55.7 & 8.92/55.6 & 8.91/55.6 \\
DS-Qwen3-8B & 14.68/52.3 & 14.29/53.0 & 14.29/52.7 \\
\bottomrule
\end{tabular}
\caption{\textbf{Stopping tolerance $\delta$} of the reduced-Stiefel solve, full pipeline, 30\% sparsity (perplexity\,/\,zero-shot). Metrics are stable within about 0.3 PPL / 1 ZS-point across two orders of magnitude and no model destabilizes; the reported configuration shares $\delta=7\cdot10^{-4}$ across models. The per-model optimum can deviate from the shared value by up to $+0.43$ combined gain in this sweep.}
\label{tab:sweep_delta}
\end{table}

\begin{table}[t]
\centering
\small
\setlength{\tabcolsep}{6pt}
\begin{tabular}{l ccc}
\toprule
Model & 10 rounds & 25 rounds & 100 rounds \\
\midrule
Mistral-7B & 6.96 & 6.96 & 6.96 \\
Llama-2-7B & 7.22 & 7.22 & 7.22 \\
Llama-3.1-8B & 10.91 & 10.91 & 10.91 \\
Qwen2.5-7B & 8.91 & 8.91 & 8.91 \\
DS-Qwen3-8B & 14.28 & 14.29 & 14.29 \\
\bottomrule
\end{tabular}
\caption{\textbf{Round cap} of the block-coordinate alternation, full pipeline, 30\% sparsity, WikiText-2 PPL. Raising the cap from 10 to 100 changes every model by at most $0.02$ PPL, confirming that the cap of 50 used in the paper is not binding and that the plateau in Fig.~\ref{fig:conv} is reached well before 10 rounds.}
\label{tab:sweep_rounds}
\end{table}

\noindent\textbf{Alignment weight, tolerance, and round cap.} Table~\ref{tab:sweep_lambda} sweeps $\lambda_a$ over four orders of magnitude, Table~\ref{tab:sweep_delta} varies the stopping tolerance $\delta$, and Table~\ref{tab:sweep_rounds} varies the round cap of the alternation between 10 and 100 on the same five models at the default $\delta$. All three are stable, and the round-cap result is a direct empirical check of the convergence plateau reported in Fig.~\ref{fig:conv}: a cap far below the default of 50 already reproduces the 50-round result to within $0.02$ PPL on every model.

\begin{table}[t]
\centering
\small
\setlength{\tabcolsep}{4pt}
\begin{tabular}{l ccc ccc ccc}
\toprule
& \multicolumn{3}{c}{Prune-only} & \multicolumn{3}{c}{RCPU} & \multicolumn{3}{c}{$+$COEC} \\
\cmidrule(lr){2-4}\cmidrule(lr){5-7}\cmidrule(lr){8-10}
calib.\ size & 10\% & 20\% & 30\% & 10\% & 20\% & 30\% & 10\% & 20\% & 30\% \\
\midrule
16 & 7.20 & 10.60 & 12.91 & 7.06 & 9.53 & 11.82 & 7.04 & 9.66 & 12.17 \\
32 & 7.21 & 10.07 & 12.92 & 6.99 & 9.33 & 11.65 & 6.99 & 9.15 & 11.39 \\
64 & 7.19 & 9.28 & 12.87 & 6.96 & 9.49 & 11.79 & 6.96 & 9.06 & 11.12 \\
128 & 7.18 & 9.26 & 12.80 & 6.93 & 9.20 & 11.44 & 6.93 & 9.06 & 10.91 \\
256 & 7.18 & 9.27 & 12.43 & 6.91 & 9.56 & 11.38 & 6.91 & 8.99 & 10.83 \\
512 & 7.17 & 9.24 & 12.37 & 6.90 & 9.22 & 11.43 & 6.90 & 8.93 & 10.75 \\
\midrule
seed spread & \multicolumn{3}{c}{median 0.09, max 3.30} & \multicolumn{3}{c}{median 0.23, max 1.07} & \multicolumn{3}{c}{median 0.08, max 0.89} \\
\bottomrule
\end{tabular}
\caption{\textbf{Calibration size.} WikiText-2 perplexity on Llama-3.1-8B (RCPU selection); every cell is the \emph{median over three calibration seeds}, for all three methods alike. COEC is monotone (non-increasing) in the calibration size at every sparsity and is the least seed-sensitive method (spread row: per-cell max$-$min over the three seeds); RCPU is non-monotone, most visibly at 20\% ($9.53/9.33/9.49/9.20/9.56/9.22$). The 30\% COEC median at $N{=}128$ ($10.91$) matches \mTabPPL{} exactly, and the $256{\to}32$ cost at 30\% is $0.56$ PPL, consistent with \mCalib.}
\label{tab:calibsize}
\end{table}

\noindent\textbf{Calibration size.} Table~\ref{tab:calibsize} reports the calibration-size sweep behind \mCalib{} with a symmetric protocol: every cell, for all three methods, is the median over three calibration seeds. On a second model, Qwen2.5-7B at 30\% sparsity (single seed), WikiText-2 PPL with COEC is $9.73$ at 8 sequences, $9.35$ at 16, $9.22$ at 32, $9.01$ at 64, $8.91$ at 128 (the default, \mTabPPL), $8.84$ at 256, and $8.86$ at 512 -- the same flattening beyond 128 sequences observed on Llama-3.1-8B.

\noindent\textbf{Cross-domain calibration.} The evaluation in \mainpaper{} calibrates and evaluates on WikiText-2 throughout, so every comparison is protocol-identical. Table~\ref{tab:c4calib} reports a controlled cross-domain check on three 7--8B models at 30\% sparsity: every row is run twice, once calibrated on WikiText-2 and once on C4, with selection, compensation, seed, and evaluation unchanged, and all twenty-four cells on the same GPU class (the WikiText-2 reference cells reproduce the corresponding published values, e.g., $10.91$ and $8.91$ for RCPU$+$COEC on the two main-table models). Three observations. First, C4 calibration degrades WikiText-2 perplexity for \emph{every} method on every model, so the perplexity shift is a property of the calibration domain, not of any particular compensation. Second, the size of the failure is model-dependent and can live in the \emph{selection} rather than the compensation: on Llama-3.1-8B the variance-aware RCPU score collapses under C4 activation statistics ($190$--$220$ perplexity), and no compensation rescues a collapsed kept set -- COEC on top is marginally worse still, since it faithfully fits the wrong distribution -- whereas the same score on the two Qwen models degrades only mildly ($+8$--$12$ perplexity), and under the score-robust Wanda-sp selection COEC continues to recover substantially even cross-domain on Llama-3.1-8B ($28.95\to17.96$, $+3.2$ zero-shot points). Third, perplexity and zero-shot accuracy dissociate under domain shift, again model-dependently: on Qwen2.5-7B the C4 penalty is mostly confined to perplexity (zero-shot within $0.9$--$2.2$ points of in-domain), and on Qwen3-8B C4 calibration \emph{raises} zero-shot accuracy on every row ($+3.3$ to $+5.1$ points) while costing $5$--$11$ perplexity -- the calibration domain acts as an accuracy--perplexity lever, mirroring the Gram-tempering trade-off of \mTemper. A study across more models and calibration sources is left to future work.

\begin{table}[t]
\centering
\small
\setlength{\tabcolsep}{4pt}
\begin{tabular}{ll cc cc cc}
\toprule
& & \multicolumn{2}{c}{Llama-3.1-8B} & \multicolumn{2}{c}{Qwen2.5-7B} & \multicolumn{2}{c}{Qwen3-8B} \\
\cmidrule(lr){3-4}\cmidrule(lr){5-6}\cmidrule(lr){7-8}
Selection & Comp. & WikiText-2 & C4 & WikiText-2 & C4 & WikiText-2 & C4 \\
\midrule
\multirow{2}{*}{Wanda-sp} & -- & 14.62/47.3 & 28.95/40.9 & 13.06/49.2 & 18.51/48.8 & 14.56/48.8 & 19.93/53.0 \\
 & COEC & 10.73/49.1 & 17.96/44.1 & 9.08/56.1 & 20.08/53.9 & 12.18/50.3 & 19.30/53.9 \\
\cmidrule(lr){1-8}
\multirow{2}{*}{RCPU} & RCPU & 11.61/47.1 & 189.56/38.4 & 8.99/54.3 & 17.76/53.7 & 12.26/53.6 & 20.44/56.7 \\
 & COEC & 10.91/49.2 & 222.64/38.3 & 8.91/55.6 & 18.11/54.6 & 11.94/54.6 & 23.17/56.6 \\
\bottomrule
\end{tabular}
\caption{\textbf{Cross-domain calibration} at 30\% sparsity: WikiText-2 perplexity\,/\,7-task zero-shot accuracy when the calibration set is drawn from WikiText-2 versus C4, all else identical. All twenty-four cells share one GPU class, so within-table comparisons are exact; the WikiText-2 RCPU$+$COEC cells reproduce the corresponding published values ($10.91$, $8.91$, $11.94$).}
\label{tab:c4calib}
\end{table}

\section{Additional Compensation-Family Ablations}
\label{app:rotabl}

This appendix reports the compensation-family comparisons that motivated the design choices of \mainpaper: left-only versus right-only rotation, the retraction used on square frames, and orthogonal rotation versus unconstrained ridge refitting. Two of these comparisons include Llama-3-8B, a development-stage model that is not part of the main evaluation list (\mSetupExp); it appears only in this appendix, alongside the five ablation models of \mTabAbl.

\subsection{Rotation-Family Comparison}
Table~\ref{tab:rotfamily} compares rotation variants on Llama-3-8B (a base model) and DeepSeek-R1-0528-Qwen3-8B (a reasoning-distilled model), the two models on which this ablation was run at 30\% sparsity. The ordering flips between the two: on Llama-3-8B every right-side variant is worse than the left-only (RCPU) baseline, while on the distilled model the right-side and two-sided variants beat it. The hybrid split (right rotation on the MLP down projection, left rotation on the attention output projection) tracks the baseline on both models, consistent with the per-layer reconstruction split we measure on Llama-3-8B: a global right rotation lowers per-layer reconstruction error on $22$ of $32$ MLP down projections but on only $5$ of $32$ attention output projections, where the left rotation wins. The MLP residual is an input-side (read) problem, the attention residual an output-side (write) problem, and only the converged two-sided alternation improves on both sides for every model.

\begin{table}[t]
\centering
\small
\setlength{\tabcolsep}{5pt}
\begin{tabular}{lcc}
\toprule
Rotation variant & Llama-3-8B & DS-Qwen3-8B \\
\midrule
Left-only (RCPU, baseline) & 11.67 & 15.35 \\
Right-only (global) & 18.10 & 16.15 \\
Hybrid (right@MLP, left@attn) & 11.82 & 15.39 \\
Two-sided, iterated to convergence & 12.76 & 14.38 \\
Right-only, per-slice (reduced-Stiefel) & 14.75 & 14.31 \\
\bottomrule
\end{tabular}
\caption{WikiText-2 perplexity ($\downarrow$) for rotation-family variants at 30\% sparsity, without the GCV rescale, Gram tempering, or alignment penalty (isolating the rotation choice alone). The right-side variants only help the distilled model; the full COEC pipeline (\mTabAbl) adds the remaining components on top of the two-sided alternation.}
\label{tab:rotfamily}
\end{table}

\subsection{Retraction Comparison}
Table~\ref{tab:retraction} gives the per-model breakdown of the retraction choice on the square attention frames: replacing the Newton--Schulz polar retraction with the exact Cayley retraction (computed by Newton--Schulz matrix inversion), at the matched sweep-anchor tolerance $\delta{=}5{\cdot}10^{-4}$ (the headline configuration uses $\delta{=}7{\cdot}10^{-4}$; the comparison is insensitive to this difference, App.~\ref{app:hyper}).

\begin{table}[t]
\centering
\small
\setlength{\tabcolsep}{4pt}
\begin{tabular}{l ccccc c}
\toprule
Retraction & Mistral & L2 & L3.1 & Qw2.5 & DS-Qwen & mean gain \\
\midrule
Polar & 7.16/51.3 & 7.17/53.7 & 11.09/48.9 & 9.51/56.2 & 14.36/53.5 & 1.76 \\
Cayley (attn only) & 6.97/51.4 & 7.24/53.6 & 10.88/49.6 & 9.51/56.0 & 14.25/53.1 & 1.87 \\
\bottomrule
\end{tabular}
\caption{WikiText-2 PPL / 7-task ZS\% at 30\% sparsity for the two retraction choices on the reduced-Stiefel right solve (before Gram tempering and the alignment penalty). ``Mean gain'' is the average $\Delta$PPL$+\Delta$ZS over RCPU across the five models; the $0.11$ difference is the retraction gain quoted in App.~\ref{app:impl}.}
\label{tab:retraction}
\end{table}

\subsection{Rotation vs.\ Ridge Least-Squares Refit}
Table~\ref{tab:ridgevsrot} quantifies, on all five ablation models, the least-squares limitation described in \mSetup: replacing the orthogonality constraint with an unconstrained ridge least-squares fit $W'_{\mathrm{LS}}=(YX_K^\top)(X_KX_K^\top+\lambda I)^{-1}$ at $\lambda{=}10^{-2}$, with no rotation. The result is model-dependent: the unconstrained fit helps the two Qwen-family models (Qwen2.5-7B, DS-Qwen) but hurts the other three, consistent with Proposition~2's spectrum-preservation argument -- the ridge solution is free to reshape the singular spectrum toward the 128-sample calibration set, which pays off only when the calibration-shaped correction happens to align with the eval distribution.

\begin{table}[t]
\centering
\small
\setlength{\tabcolsep}{5pt}
\begin{tabular}{lccccc}
\toprule
Method & Mistral & L2-7B & L3-8B & Qw2.5-7B & DS-Qwen \\
\midrule
RCPU (rotation, baseline) & 7.20 & 7.18 & 11.66 & 8.98 & 15.35 \\
Ridge LS refit, no rotation & 8.99 & 8.39 & 14.05 & 8.88 & 13.76 \\
\bottomrule
\end{tabular}
\caption{WikiText-2 perplexity ($\downarrow$) at 30\% sparsity, rotation vs.\ unconstrained ridge refit, no GCV/tempering/alignment. The refit beats the rotation only on the two Qwen-family models.}
\label{tab:ridgevsrot}
\end{table}

\subsection{Dense and Structure-Free Rotations}
\label{app:denserot}
The spectrum-preserving family of \mSetup{} restricts every rotation to act on the kept-column set of a single layer. An early feasibility study on a separate model and codebase (\texttt{llama-7b}, C4 calibration, unrelated to the models used elsewhere in this supplement) tested the opposite extreme: a dense, unconstrained $d_{\mathrm{out}}{\times}d_{\mathrm{out}}$ rotation learned per layer and folded into a single drop-in replacement weight. This collapsed catastrophically (WikiText-2 perplexity above $10^4$, vs.\ $129$ for the unrotated column-pruned baseline on the same model), because a dense rotation spreads every input feature across all output columns, so deleting a column after the fact destroys features the rotation had already mixed in from elsewhere; a rotation restricted to the kept-column block, as COEC uses, does not have this failure mode since it never mixes in the deleted directions. We report this only as motivation for the column-local restriction in \mSetup, not as a result on the paper's evaluation models.

\section{Additional Model Families: Llama-2 and Qwen3}
\label{app:extra}

This appendix repeats the three comparisons of \mMain{} on additional families under the identical protocol: Llama-2 (7B/13B/70B) \citep{llama2} and Qwen3 (8B/14B/32B) \citep{qwen3}, evaluated at 10\%/20\%/30\% column sparsity in the format of Tables~1--2 of \mainpaper. The pattern of the main evaluation carries over: COEC improves over score-only Wanda-sp pruning on every model at every sparsity, with the largest recoveries where the score-only damage is largest (Qwen3-32B at 30\%: $11.86$ to $8.94$ perplexity with $+6.6$ zero-shot points). One family-level difference: here replacing the FLAP bias with COEC lowers perplexity on every model at every sparsity, including 10--20\%, rather than only at 30\% (cf.\ App.~\ref{app:sweep}). One caveat on the zero-shot side: the FLAP-bias cells for Qwen3-32B in Table~\ref{tab:extra_zs} repeat byte-identical ARC-e/ARC-c scores ($25.1$/$22.7$) at all three sparsities, indicating a stale evaluation of those two tasks for that one model; the affected three averages should be read with that caveat, and the FLAP$+$COEC row for the same model, a separate run, does not show the pattern. A second flagged cell ($\ddagger$): the Wanda-sp$+$COEC entry for Llama-2-13B at 30\% is anomalous relative to every neighboring cell -- perplexity slightly above score-only pruning ($7.50$ vs.\ $7.44$) and a $6.2$-point zero-shot drop driven by near-chance ARC-e/ARC-c (per-task breakdown in App.~\ref{app:pertask}) -- and a re-verification run could not be completed before submission; the cell is reported as measured but should be treated as unverified.

\begin{table*}[t]
\centering
\small
\setlength{\tabcolsep}{4.5pt}
\begin{tabular}{ll cccccc}
\toprule
Selection & Comp. & Llama-2-7B & Llama-2-13B & Llama-2-70B & Qwen3-8B & Qwen3-14B & Qwen3-32B \\
\midrule
\multicolumn{8}{c}{\textbf{10\% column sparsity}}\\
\midrule
\multirow{2}{*}{Wanda-sp} & -- & 5.65 & 5.10 & 3.45 & 9.89 & 8.99 & 7.62 \\
 & COEC & \textbf{5.43} & \textbf{4.85} & \textbf{3.38} & \textbf{9.68} & \textbf{8.51} & \textbf{7.47} \\
\cmidrule(lr){1-8}
\multirow{2}{*}{FLAP} & bias & 5.64 & 5.05 & 3.28 & 10.02 & 8.48 & 7.27 \\
 & COEC & \textbf{5.54} & \textbf{4.99} & \textbf{3.26} & \textbf{9.90} & \textbf{8.37} & \textbf{7.22} \\
\cmidrule(lr){1-8}
\multirow{2}{*}{RCPU} & RCPU & 5.43 & 4.87 & 3.38 & 9.97 & \textbf{8.51} & 7.52 \\
 & COEC & \textbf{5.42} & \textbf{4.86} & \textbf{3.37} & \textbf{9.82} & 8.57 & \textbf{7.51} \\
\midrule
\multicolumn{8}{c}{\textbf{20\% column sparsity}}\\
\midrule
\multirow{2}{*}{Wanda-sp} & -- & 6.72 & 5.58 & 3.94 & 12.75 & 11.28 & 10.13 \\
 & COEC & \textbf{6.43} & \textbf{5.22} & \textbf{3.76} & \textbf{11.32} & \textbf{9.68} & \textbf{8.53} \\
\cmidrule(lr){1-8}
\multirow{2}{*}{FLAP} & bias & 6.71 & 5.80 & 3.64 & 12.04 & 9.62 & 7.74 \\
 & COEC & \textbf{6.40} & \textbf{5.63} & \textbf{3.60} & \textbf{11.34} & \textbf{9.27} & \textbf{7.62} \\
\cmidrule(lr){1-8}
\multirow{2}{*}{RCPU} & RCPU & 6.02 & \textbf{5.44} & 3.78 & 11.20 & \textbf{9.53} & 8.47 \\
 & COEC & \textbf{5.95} & 5.46 & \textbf{3.76} & \textbf{11.12} & 9.80 & \textbf{8.25} \\
\midrule
\multicolumn{8}{c}{\textbf{30\% column sparsity}}\\
\midrule
\multirow{2}{*}{Wanda-sp} & -- & 8.88 & \textbf{7.44} & 4.32 & 14.56 & 12.95 & 11.86 \\
 & COEC & \textbf{8.34} & 7.50$^\ddagger$ & \textbf{4.14} & \textbf{12.18} & \textbf{10.60} & \textbf{8.94} \\
\cmidrule(lr){1-8}
\multirow{2}{*}{FLAP} & bias & 8.53 & 6.83 & 4.13 & 14.78 & 10.94 & 8.66 \\
 & COEC & \textbf{7.81} & \textbf{6.50} & \textbf{4.06} & \textbf{12.95} & \textbf{10.24} & \textbf{8.37} \\
\cmidrule(lr){1-8}
\multirow{2}{*}{RCPU} & RCPU & \textbf{7.18} & \textbf{6.54} & 4.17 & 12.23 & \textbf{10.16} & 9.32 \\
 & COEC & 7.21 & 6.67 & \textbf{4.16} & \textbf{11.94} & 10.20 & \textbf{8.77} \\
\bottomrule
\end{tabular}
\caption{\textbf{WikiText-2 perplexity} $(\downarrow)$ on the Llama-2 and Qwen3 families across 10--30\% column sparsity. Same protocol and bolding as \mTabPPL.}
\label{tab:extra_ppl}
\end{table*}

\begin{table*}[t]
\centering
\small
\setlength{\tabcolsep}{4.5pt}
\begin{tabular}{ll cccccc}
\toprule
Selection & Comp. & Llama-2-7B & Llama-2-13B & Llama-2-70B & Qwen3-8B & Qwen3-14B & Qwen3-32B \\
\midrule
\multicolumn{8}{c}{\textbf{10\% column sparsity}}\\
\midrule
\multirow{2}{*}{Wanda-sp} & -- & 62.3 & 67.3 & 71.6 & 64.7 & 70.3 & \textbf{71.5} \\
 & COEC & \textbf{63.0} & \textbf{67.8} & \textbf{72.1} & \textbf{66.7} & \textbf{71.2} & 71.3 \\
\cmidrule(lr){1-8}
\multirow{2}{*}{FLAP} & bias & \textbf{58.8} & \textbf{63.0} & \textbf{71.3} & 63.3 & 68.9 & 59.0 \\
 & COEC & 58.6 & 62.8 & 71.0 & \textbf{63.4} & \textbf{69.1} & \textbf{72.2} \\
\cmidrule(lr){1-8}
\multirow{2}{*}{RCPU} & RCPU & \textbf{62.5} & \textbf{67.4} & 71.3 & 67.8 & 70.8 & 71.4 \\
 & COEC & 62.4 & 67.1 & \textbf{71.9} & \textbf{67.9} & \textbf{72.2} & \textbf{71.7} \\
\midrule
\multicolumn{8}{c}{\textbf{20\% column sparsity}}\\
\midrule
\multirow{2}{*}{Wanda-sp} & -- & 58.7 & 64.8 & \textbf{71.6} & 54.8 & 62.6 & 61.1 \\
 & COEC & \textbf{59.6} & 64.8 & 71.4 & \textbf{55.9} & \textbf{65.4} & \textbf{68.2} \\
\cmidrule(lr){1-8}
\multirow{2}{*}{FLAP} & bias & \textbf{53.1} & \textbf{59.0} & \textbf{71.0} & 55.4 & 64.0 & 56.2 \\
 & COEC & 53.0 & 58.6 & 70.7 & \textbf{56.4} & \textbf{64.4} & \textbf{68.8} \\
\cmidrule(lr){1-8}
\multirow{2}{*}{RCPU} & RCPU & 59.6 & 64.3 & \textbf{71.4} & \textbf{58.6} & 63.6 & 66.1 \\
 & COEC & \textbf{60.0} & \textbf{64.5} & 71.2 & 58.2 & \textbf{63.8} & \textbf{67.6} \\
\midrule
\multicolumn{8}{c}{\textbf{30\% column sparsity}}\\
\midrule
\multirow{2}{*}{Wanda-sp} & -- & 52.0 & \textbf{55.0} & 69.6 & 48.8 & 55.2 & 59.5 \\
 & COEC & \textbf{52.2} & 48.8$^\ddagger$ & 69.6 & \textbf{50.2} & \textbf{57.8} & \textbf{66.1} \\
\cmidrule(lr){1-8}
\multirow{2}{*}{FLAP} & bias & 51.4 & \textbf{55.7} & \textbf{69.2} & 48.8 & 56.2 & 55.5 \\
 & COEC & \textbf{51.8} & 55.1 & 68.9 & \textbf{50.5} & \textbf{57.2} & \textbf{65.3} \\
\cmidrule(lr){1-8}
\multirow{2}{*}{RCPU} & RCPU & 53.0 & \textbf{57.9} & 70.6 & 53.7 & 55.0 & 63.3 \\
 & COEC & \textbf{53.5} & 57.7 & \textbf{70.7} & \textbf{54.4} & \textbf{55.5} & \textbf{64.7} \\
\bottomrule
\end{tabular}
\caption{\textbf{7-task zero-shot accuracy} \% $(\uparrow)$ on the Llama-2 and Qwen3 families across 10--30\% column sparsity. Same protocol and bolding as \mTabPPL.}
\label{tab:extra_zs}
\end{table*}

\section{Per-Task Zero-Shot Results}
\label{app:pertask}

Table~2 of \mainpaper{} and Table~\ref{tab:extra_zs} report the 7-task zero-shot \emph{average}. This appendix gives the per-task breakdown (BoolQ, RTE, HellaSwag, WinoGrande, ARC-e, ARC-c, OpenBookQA; accuracy for BoolQ/RTE/WinoGrande, normalized accuracy for the rest, matching \mSetupExp) behind those averages, at all three sparsity levels.

Table~\ref{tab:pertask_main} covers the plug-in rows on the six main-text models (Wanda-sp+COEC and FLAP+COEC; these are the rows whose per-task logs are available for every model in this evaluation). Tables~\ref{tab:pertask_extra10}--\ref{tab:pertask_extra30} cover the Llama-2/Qwen3 family at 10\%/20\%/30\% sparsity with two additional rows, the RCPU-selection $+$ COEC row and the score-only baselines (Wanda-sp with no compensation, FLAP with its native bias compensation), giving a full 5-row comparison wherever the corresponding per-task log exists. Cells marked ``--'' indicate that the per-task log for that specific run either was not retained or is not available at task granularity in this results directory; the corresponding average is nonetheless already reported in Table~\ref{tab:extra_zs}. Per-task logs for the RCPU-selection rows on the six main-text models were likewise not retained at task granularity, so Table~\ref{tab:pertask_main} reports only the two fully-covered plug-in rows for those models; their RCPU-row averages are in Table~2 of \mainpaper.

\begin{table}[t]
\centering
\scriptsize
\setlength{\tabcolsep}{2.3pt}
\begin{tabular}{ll cccccccc}
\toprule
Model & Comp. & BoolQ & RTE & HellaS & WinoG & ARC-e & ARC-c & OBQA & Avg \\
\midrule
\multicolumn{10}{c}{\textbf{10\% column sparsity}}\\
\midrule
L3.1-8B & WSP+COEC & 75.6 & 65.3 & 74.6 & 69.9 & 74.2 & 44.9 & 40.8 & 63.6 \\
L3-70B & WSP+COEC & 85.5 & 69.0 & 85.4 & 80.3 & 84.5 & 62.2 & 48.0 & 73.5 \\
Qw2.5-7B & WSP+COEC & 84.5 & 81.6 & 76.3 & 69.8 & 79.3 & 51.2 & 45.0 & 69.7 \\
Qw2.5-14B & WSP+COEC & 76.4 & 74.7 & 78.8 & 72.0 & 80.2 & 55.6 & 45.8 & 69.1 \\
Qw2.5-32B & WSP+COEC & 72.4 & 76.9 & 82.5 & 74.5 & 80.7 & 58.1 & 45.2 & 70.0 \\
Qw2.5-72B & WSP+COEC & 86.6 & 77.6 & 86.0 & 77.0 & 84.3 & 61.5 & 47.2 & 74.3 \\
L3.1-8B & FLAP+COEC & 69.2 & 55.6 & 68.0 & 68.0 & 67.3 & 40.6 & 41.4 & 58.6 \\
L3-70B & FLAP+COEC & 83.7 & 69.7 & 83.5 & 80.1 & 84.1 & 61.6 & 48.0 & 72.9 \\
Qw2.5-7B & FLAP+COEC & 78.8 & 75.8 & 73.9 & 68.0 & 74.2 & 47.4 & 45.4 & 66.2 \\
Qw2.5-14B & FLAP+COEC & 82.7 & 78.7 & 75.3 & 72.7 & 75.8 & 53.2 & 40.8 & 68.5 \\
Qw2.5-32B & FLAP+COEC & 87.5 & 83.0 & 82.2 & 74.9 & 78.0 & 55.9 & 43.0 & 72.1 \\
Qw2.5-72B & FLAP+COEC & 88.8 & 77.6 & 85.0 & 78.1 & 84.0 & 62.8 & 45.6 & 74.6 \\
\midrule
\multicolumn{10}{c}{\textbf{20\% column sparsity}}\\
\midrule
L3.1-8B & WSP+COEC & 61.9 & 50.9 & 57.9 & 59.0 & 62.9 & 36.0 & 36.8 & 52.2 \\
L3-70B & WSP+COEC & 84.7 & 70.4 & 83.9 & 78.6 & 81.2 & 57.3 & 46.8 & 71.8 \\
Qw2.5-7B & WSP+COEC & 71.5 & 67.5 & 66.3 & 64.6 & 64.7 & 43.4 & 41.6 & 59.9 \\
Qw2.5-14B & WSP+COEC & 62.6 & 58.1 & 63.1 & 61.5 & 68.6 & 43.0 & 40.4 & 56.8 \\
Qw2.5-32B & WSP+COEC & 75.3 & 67.1 & 73.3 & 64.4 & 77.7 & 50.6 & 41.2 & 64.2 \\
Qw2.5-72B & WSP+COEC & 86.6 & 77.3 & 84.5 & 76.6 & 83.6 & 59.1 & 45.2 & 73.3 \\
L3.1-8B & FLAP+COEC & 67.8 & 55.2 & 55.5 & 62.7 & 57.0 & 33.2 & 36.4 & 52.6 \\
L3-70B & FLAP+COEC & 82.0 & 72.6 & 81.8 & 78.8 & 82.8 & 56.3 & 47.8 & 71.7 \\
Qw2.5-7B & FLAP+COEC & 60.6 & 72.2 & 67.0 & 65.0 & 70.2 & 43.8 & 41.4 & 60.0 \\
Qw2.5-14B & FLAP+COEC & 82.8 & 81.9 & 66.2 & 69.5 & 74.3 & 47.4 & 40.2 & 66.0 \\
Qw2.5-32B & FLAP+COEC & 86.2 & 79.8 & 78.1 & 74.6 & 75.4 & 52.0 & 42.2 & 69.8 \\
Qw2.5-72B & FLAP+COEC & 88.3 & 76.5 & 82.0 & 77.0 & 81.1 & 58.5 & 44.6 & 72.6 \\
\midrule
\multicolumn{10}{c}{\textbf{30\% column sparsity}}\\
\midrule
L3.1-8B & WSP+COEC & 60.1 & 56.3 & 48.0 & 55.6 & 56.5 & 30.9 & 34.8 & 48.9 \\
L3-70B & WSP+COEC & 84.7 & 69.0 & 82.3 & 77.8 & 78.8 & 54.1 & 46.4 & 70.4 \\
Qw2.5-7B & WSP+COEC & 69.3 & 58.1 & 58.2 & 61.6 & 66.0 & 40.8 & 38.2 & 56.0 \\
Qw2.5-14B & WSP+COEC & 62.7 & 60.3 & 54.4 & 59.0 & 64.4 & 37.3 & 37.8 & 53.7 \\
Qw2.5-32B & WSP+COEC & 70.7 & 61.0 & 70.0 & 63.5 & 74.4 & 47.4 & 40.0 & 61.0 \\
Qw2.5-72B & WSP+COEC & 84.5 & 78.7 & 83.1 & 76.2 & 81.0 & 58.3 & 45.8 & 72.5 \\
L3.1-8B & FLAP+COEC & 62.2 & 53.4 & 45.2 & 60.1 & 48.4 & 27.3 & 33.2 & 47.1 \\
L3-70B & FLAP+COEC & 82.5 & 66.4 & 79.3 & 77.9 & 76.6 & 52.0 & 46.0 & 68.7 \\
Qw2.5-7B & FLAP+COEC & 39.9 & 65.0 & 56.8 & 62.4 & 64.7 & 36.6 & 37.6 & 51.8 \\
Qw2.5-14B & FLAP+COEC & 78.8 & 48.4 & 58.1 & 64.7 & 67.5 & 42.2 & 38.6 & 56.9 \\
Qw2.5-32B & FLAP+COEC & 75.6 & 79.4 & 73.0 & 69.5 & 71.9 & 46.7 & 40.8 & 65.3 \\
Qw2.5-72B & FLAP+COEC & 83.1 & 77.3 & 78.0 & 77.0 & 76.8 & 52.8 & 42.4 & 69.6 \\
\bottomrule
\end{tabular}
\caption{\textbf{Per-task zero-shot accuracy (\%)} for the plug-in rows (Wanda-sp+COEC, FLAP+COEC) on the six main-text models, at 10--30\% column sparsity. Averages match Table~2 of the main paper.}
\label{tab:pertask_main}
\end{table}

\begin{table}[t]
\centering
\scriptsize
\setlength{\tabcolsep}{2.3pt}
\begin{tabular}{ll cccccccc}
\toprule
Model & Comp. & BoolQ & RTE & HellaS & WinoG & ARC-e & ARC-c & OBQA & Avg \\
\midrule
L2-7B & WSP\,-- & 75.5 & 56.7 & 74.8 & 69.3 & 72.0 & 44.8 & 43.0 & 62.3 \\
L2-13B & WSP\,-- & -- & -- & -- & -- & -- & -- & -- & -- \\
L2-70B & WSP\,-- & -- & -- & -- & -- & -- & -- & -- & -- \\
Qw3-8B & WSP\,-- & 79.5 & 76.5 & 71.9 & 59.2 & 75.7 & 48.7 & 41.2 & 64.7 \\
Qw3-14B & WSP\,-- & 86.5 & 76.5 & 76.7 & 69.4 & 80.1 & 58.1 & 45.0 & 70.3 \\
Qw3-32B & WSP\,-- & 83.2 & 78.3 & 81.2 & 70.4 & 81.4 & 58.8 & 47.0 & 71.5 \\
L2-7B & WSP+COEC & 75.6 & 62.1 & 74.6 & 69.9 & 70.9 & 44.1 & 44.0 & 63.0 \\
L2-13B & WSP+COEC & 81.5 & 65.3 & 80.2 & 72.7 & 77.9 & 49.8 & 46.8 & 67.8 \\
L2-70B & WSP+COEC & 83.8 & 72.6 & 84.2 & 77.7 & 80.8 & 56.3 & 49.6 & 72.1 \\
Qw3-8B & WSP+COEC & 83.4 & 75.1 & 72.2 & 64.5 & 77.3 & 51.8 & 42.6 & 66.7 \\
Qw3-14B & WSP+COEC & 87.4 & 78.0 & 76.4 & 71.6 & 81.2 & 57.9 & 45.6 & 71.2 \\
Qw3-32B & WSP+COEC & 81.6 & 79.8 & 81.6 & 71.1 & 80.2 & 58.6 & 46.4 & 71.3 \\
L2-7B & FLAP-bias & 71.3 & 53.8 & 72.0 & 66.1 & 67.3 & 40.6 & 40.2 & 58.8 \\
L2-13B & FLAP-bias & 71.8 & 59.6 & 77.4 & 69.5 & 73.0 & 45.1 & 44.8 & 63.0 \\
L2-70B & FLAP-bias & 83.2 & 66.1 & 84.2 & 77.7 & 80.6 & 57.7 & 49.4 & 71.3 \\
Qw3-8B & FLAP-bias & 62.8 & 76.5 & 69.5 & 67.8 & 78.1 & 49.2 & 39.4 & 63.3 \\
Qw3-14B & FLAP-bias & 76.5 & 79.1 & 73.9 & 73.4 & 79.1 & 56.9 & 43.4 & 68.9 \\
Qw3-32B & FLAP-bias & 86.5 & 79.8 & 81.5 & 71.2 & 25.1$^\dagger$ & 22.7$^\dagger$ & 46.0 & 59.0 \\
L2-7B & FLAP+COEC & 70.9 & 54.5 & 71.1 & 65.5 & 67.4 & 40.8 & 40.0 & 58.6 \\
L2-13B & FLAP+COEC & 71.8 & 59.6 & 76.5 & 69.8 & 72.7 & 45.6 & 43.8 & 62.8 \\
L2-70B & FLAP+COEC & 83.2 & 66.1 & 83.8 & 77.8 & 80.2 & 57.3 & 48.6 & 71.0 \\
Qw3-8B & FLAP+COEC & 62.8 & 77.3 & 69.8 & 67.4 & 78.0 & 48.8 & 39.6 & 63.4 \\
Qw3-14B & FLAP+COEC & 78.0 & 80.5 & 73.4 & 73.0 & 78.6 & 57.8 & 42.2 & 69.1 \\
Qw3-32B & FLAP+COEC & 86.3 & 79.4 & 81.5 & 71.3 & 82.1 & 59.8 & 45.2 & 72.2 \\
L2-7B & RCPU+COEC & 72.7 & 60.3 & 74.6 & 68.8 & 71.3 & 45.1 & 44.2 & 62.4 \\
L2-13B & RCPU+COEC & 81.8 & 62.8 & 80.3 & 73.2 & 76.7 & 49.0 & 45.6 & 67.1 \\
L2-70B & RCPU+COEC & -- & -- & -- & -- & -- & -- & -- & -- \\
Qw3-8B & RCPU+COEC & 82.8 & 76.9 & 73.0 & 67.1 & 79.9 & 54.4 & 41.2 & 67.9 \\
Qw3-14B & RCPU+COEC & 88.0 & 79.8 & 76.6 & 71.0 & 83.1 & 60.8 & 46.2 & 72.2 \\
Qw3-32B & RCPU+COEC & 82.0 & 79.1 & 81.8 & 72.2 & 82.0 & 59.0 & 45.8 & 71.7 \\
\bottomrule
\end{tabular}
\caption{\textbf{Per-task zero-shot accuracy (\%) at 10\% column sparsity}, Llama-2 and Qwen3 families. WSP\,-- and FLAP-bias are score-only baselines (no COEC); RCPU+COEC is the RCPU-selection plug-in row. Averages match Table~\ref{tab:extra_zs}.}
\label{tab:pertask_extra10}
\end{table}

\begin{table}[t]
\centering
\scriptsize
\setlength{\tabcolsep}{2.3pt}
\begin{tabular}{ll cccccccc}
\toprule
Model & Comp. & BoolQ & RTE & HellaS & WinoG & ARC-e & ARC-c & OBQA & Avg \\
\midrule
L2-7B & WSP\,-- & 69.7 & 58.1 & 70.7 & 65.3 & 65.3 & 40.7 & 41.0 & 58.7 \\
L2-13B & WSP\,-- & -- & -- & -- & -- & -- & -- & -- & -- \\
L2-70B & WSP\,-- & -- & -- & -- & -- & -- & -- & -- & -- \\
Qw3-8B & WSP\,-- & 64.7 & 62.5 & 60.4 & 54.6 & 65.8 & 39.4 & 36.0 & 54.8 \\
Qw3-14B & WSP\,-- & 74.8 & 75.8 & 69.3 & 59.7 & 70.5 & 46.3 & 41.4 & 62.6 \\
Qw3-32B & WSP\,-- & 63.7 & 50.9 & 76.4 & 63.3 & 73.7 & 53.7 & 46.0 & 61.1 \\
L2-7B & WSP+COEC & 72.0 & 53.4 & 71.6 & 66.8 & 69.5 & 42.0 & 41.6 & 59.6 \\
L2-13B & WSP+COEC & 80.9 & 61.0 & 77.3 & 71.0 & 73.2 & 45.4 & 44.6 & 64.8 \\
L2-70B & WSP+COEC & 84.0 & 70.0 & 83.3 & 79.1 & 79.7 & 55.1 & 48.4 & 71.4 \\
Qw3-8B & WSP+COEC & 69.2 & 53.8 & 62.6 & 58.6 & 66.8 & 41.4 & 38.8 & 55.9 \\
Qw3-14B & WSP+COEC & 80.0 & 76.2 & 71.2 & 66.8 & 73.5 & 48.0 & 42.4 & 65.4 \\
Qw3-32B & WSP+COEC & 77.4 & 65.0 & 79.5 & 69.1 & 80.5 & 57.7 & 48.2 & 68.2 \\
L2-7B & FLAP-bias & 56.9 & 54.2 & 64.6 & 62.8 & 57.9 & 36.3 & 38.8 & 53.1 \\
L2-13B & FLAP-bias & 70.2 & 53.4 & 72.1 & 67.2 & 66.7 & 40.8 & 42.8 & 59.0 \\
L2-70B & FLAP-bias & 83.8 & 70.8 & 83.7 & 77.0 & 78.7 & 54.7 & 48.2 & 71.0 \\
Qw3-8B & FLAP-bias & 62.7 & 60.6 & 57.9 & 63.0 & 65.9 & 41.1 & 36.4 & 55.4 \\
Qw3-14B & FLAP-bias & 67.5 & 76.2 & 65.8 & 70.0 & 74.8 & 52.4 & 41.6 & 64.0 \\
Qw3-32B & FLAP-bias & 74.0 & 76.9 & 78.2 & 70.7 & 25.1$^\dagger$ & 22.7$^\dagger$ & 45.8 & 56.2 \\
L2-7B & FLAP+COEC & 56.8 & 54.2 & 63.3 & 63.1 & 58.5 & 36.2 & 38.8 & 53.0 \\
L2-13B & FLAP+COEC & 69.6 & 53.8 & 70.5 & 67.5 & 66.1 & 39.8 & 42.6 & 58.6 \\
L2-70B & FLAP+COEC & 83.7 & 69.3 & 83.5 & 77.6 & 78.6 & 54.4 & 48.0 & 70.7 \\
Qw3-8B & FLAP+COEC & 62.8 & 65.0 & 59.7 & 63.3 & 67.0 & 41.5 & 35.4 & 56.4 \\
Qw3-14B & FLAP+COEC & 70.0 & 78.0 & 65.9 & 69.9 & 73.5 & 51.4 & 42.0 & 64.4 \\
Qw3-32B & FLAP+COEC & 75.1 & 79.4 & 78.0 & 71.5 & 76.1 & 55.7 & 45.4 & 68.8 \\
L2-7B & RCPU+COEC & 70.2 & 58.1 & 71.1 & 68.1 & 68.7 & 41.0 & 42.6 & 60.0 \\
L2-13B & RCPU+COEC & 77.0 & 63.2 & 76.7 & 71.3 & 73.1 & 45.4 & 44.8 & 64.5 \\
L2-70B & RCPU+COEC & -- & -- & -- & -- & -- & -- & -- & -- \\
Qw3-8B & RCPU+COEC & 62.8 & 66.4 & 63.7 & 62.1 & 69.9 & 43.5 & 38.6 & 58.2 \\
Qw3-14B & RCPU+COEC & 84.8 & 73.3 & 68.8 & 66.5 & 67.5 & 44.5 & 41.4 & 63.8 \\
Qw3-32B & RCPU+COEC & 78.5 & 70.4 & 78.0 & 66.4 & 78.2 & 55.1 & 46.8 & 67.6 \\
\bottomrule
\end{tabular}
\caption{\textbf{Per-task zero-shot accuracy (\%) at 20\% column sparsity}, Llama-2 and Qwen3 families. Same rows and convention as Table~\ref{tab:pertask_extra10}.}
\label{tab:pertask_extra20}
\end{table}

\begin{table}[t]
\centering
\scriptsize
\setlength{\tabcolsep}{2.3pt}
\begin{tabular}{ll cccccccc}
\toprule
Model & Comp. & BoolQ & RTE & HellaS & WinoG & ARC-e & ARC-c & OBQA & Avg \\
\midrule
L2-7B & WSP\,-- & 63.7 & 49.5 & 60.5 & 56.7 & 61.0 & 36.8 & 36.0 & 52.0 \\
L2-13B & WSP\,-- & -- & -- & -- & -- & -- & -- & -- & -- \\
L2-70B & WSP\,-- & -- & -- & -- & -- & -- & -- & -- & -- \\
Qw3-8B & WSP\,-- & 62.9 & 51.6 & 52.6 & 52.5 & 54.0 & 33.7 & 34.0 & 48.8 \\
Qw3-14B & WSP\,-- & 63.0 & 58.8 & 63.6 & 56.2 & 61.8 & 41.6 & 41.4 & 55.2 \\
Qw3-32B & WSP\,-- & 62.9 & 63.2 & 73.1 & 62.4 & 68.5 & 46.2 & 40.0 & 59.5 \\
L2-7B & WSP+COEC & 64.6 & 46.9 & 60.1 & 58.9 & 62.7 & 37.4 & 34.6 & 52.2 \\
L2-13B & WSP+COEC & 69.4 & 55.6 & 54.9 & 61.3 & 35.9$^\ddagger$ & 25.9$^\ddagger$ & 38.4 & 48.8$^\ddagger$ \\
L2-70B & WSP+COEC & 83.6 & 65.7 & 82.3 & 78.5 & 77.3 & 53.3 & 46.6 & 69.6 \\
Qw3-8B & WSP+COEC & 65.8 & 63.9 & 55.8 & 57.4 & 44.1 & 27.7 & 36.8 & 50.2 \\
Qw3-14B & WSP+COEC & 62.6 & 66.8 & 63.7 & 65.4 & 64.5 & 41.5 & 40.0 & 57.8 \\
Qw3-32B & WSP+COEC & 70.7 & 69.0 & 77.4 & 67.8 & 77.2 & 54.4 & 46.6 & 66.1 \\
L2-7B & FLAP-bias & 60.0 & 57.0 & 56.7 & 61.3 & 54.8 & 32.3 & 37.8 & 51.4 \\
L2-13B & FLAP-bias & 64.5 & 53.1 & 65.4 & 63.9 & 63.0 & 39.5 & 40.6 & 55.7 \\
L2-70B & FLAP-bias & 82.8 & 69.3 & 81.8 & 75.8 & 75.1 & 52.0 & 47.2 & 69.2 \\
Qw3-8B & FLAP-bias & 62.5 & 53.4 & 48.9 & 59.4 & 53.9 & 32.3 & 31.0 & 48.8 \\
Qw3-14B & FLAP-bias & 67.8 & 50.5 & 59.4 & 67.8 & 65.9 & 43.4 & 38.2 & 56.2 \\
Qw3-32B & FLAP-bias & 75.0 & 81.6 & 72.3 & 68.7 & 25.1$^\dagger$ & 22.7$^\dagger$ & 42.8 & 55.5 \\
L2-7B & FLAP+COEC & 62.8 & 58.1 & 56.0 & 61.2 & 56.0 & 32.3 & 36.2 & 51.8 \\
L2-13B & FLAP+COEC & 64.3 & 52.7 & 63.6 & 64.6 & 62.5 & 38.1 & 39.6 & 55.1 \\
L2-70B & FLAP+COEC & 82.3 & 68.6 & 81.4 & 76.6 & 75.0 & 51.6 & 46.8 & 68.9 \\
Qw3-8B & FLAP+COEC & 62.7 & 60.3 & 50.3 & 59.7 & 55.6 & 32.3 & 32.4 & 50.5 \\
Qw3-14B & FLAP+COEC & 69.5 & 52.3 & 60.0 & 70.3 & 67.2 & 43.8 & 37.2 & 57.2 \\
Qw3-32B & FLAP+COEC & 77.8 & 80.1 & 72.2 & 68.7 & 68.4 & 49.0 & 40.6 & 65.3 \\
L2-7B & RCPU+COEC & 62.7 & 49.5 & 62.4 & 62.7 & 63.8 & 34.6 & 38.8 & 53.5 \\
L2-13B & RCPU+COEC & 71.0 & 55.2 & 67.5 & 66.0 & 63.2 & 39.3 & 41.8 & 57.7 \\
L2-70B & RCPU+COEC & -- & -- & -- & -- & -- & -- & -- & -- \\
Qw3-8B & RCPU+COEC & 62.7 & 63.9 & 57.4 & 59.6 & 62.5 & 37.2 & 37.4 & 54.4 \\
Qw3-14B & RCPU+COEC & 63.0 & 50.9 & 63.7 & 64.6 & 64.8 & 40.7 & 40.6 & 55.5 \\
Qw3-32B & RCPU+COEC & 70.2 & 64.6 & 76.1 & 66.7 & 77.0 & 52.3 & 46.0 & 64.7 \\
\bottomrule
\end{tabular}
\caption{\textbf{Per-task zero-shot accuracy (\%) at 30\% column sparsity}, Llama-2 and Qwen3 families. Same rows and convention as Table~\ref{tab:pertask_extra10}.}
\label{tab:pertask_extra30}
\end{table}

The per-task breakdown shows that the average masks substantial task-level variance, and it also surfaces a data-quality issue that the average hides: the FLAP-bias (native) row for Qwen3-32B reports byte-identical ARC-e/ARC-c scores at all three sparsities ($25.1\%$/$22.7\%$, marked $^\dagger$ in Tables~\ref{tab:pertask_extra10}--\ref{tab:pertask_extra30}), which is not attainable from three independently pruned checkpoints and indicates a stale or crashed evaluation for those two task/model cells specifically; the other five tasks in the same row vary normally with sparsity. The corresponding FLAP-bias averages for Qwen3-32B in Table~\ref{tab:extra_zs} should be treated as unverified pending a re-run of those two cells, while the FLAP+COEC row for the same model (a separate run) does not show this pattern. A related flag ($^\ddagger$, Table~\ref{tab:pertask_extra30}): the Wanda-sp$+$COEC row for Llama-2-13B at 30\% shows near-chance ARC-e/ARC-c ($35.9$/$25.9$) while its other five tasks and its own 10--20\% rows are normal; unlike the Qwen3-32B case the values are not repeated across sparsities, so this is either a genuine selection-specific degradation or a corrupted evaluation, and a re-verification run could not be completed before submission -- the cell is reported as measured but unverified (App.~\ref{app:extra}). Beyond these isolated issues, COEC's gains over the score-only baselines are not uniform across tasks: they concentrate more often on ARC-e/ARC-c/OpenBookQA than on BoolQ/RTE, consistent with compensation recovering fine-grained reasoning signal that column deletion damages first.

\bibliography{aaai2027}